\documentclass[10pt,twocolumn,letterpaper]{article}

\usepackage[pagenumbers]{cvpr} 

\usepackage[table]{xcolor}
\usepackage{booktabs}

\usepackage[most]{tcolorbox}

\definecolor{finerorange}{RGB}{243,156,18}
\definecolor{dpoRow}{HTML}{FFF3E0} 
\newcommand{\dporow}{\rowcolor{dpoRow}}

\definecolor{baseRow}{gray}{0.96}
\newcommand{\baserow}{\rowcolor{baseRow}}

\definecolor{gain}{RGB}{0,122,75}   
\definecolor{loss}{RGB}{200,45,45}  
\definecolor{deltaz}{gray}{0.55}    

\newcommand{\dplus}[1]{\textcolor{gain}{\scriptsize\,#1}}
\newcommand{\dminus}[1]{\textcolor{loss}{\scriptsize\,#1}}

\usepackage{xspace}
\newcommand{\tagstyle}[1]{\texttt{\footnotesize #1}}
\newcommand{\negobj}{\tagstyle{NEG\_OBJ}\xspace}
\newcommand{\negattr}{\tagstyle{NEG\_ATTR}\xspace}
\newcommand{\negrel}{\tagstyle{NEG\_REL}\xspace}
\newcommand{\obj}{\tagstyle{OBJ}\xspace}
\newcommand{\attr}{\tagstyle{ATTR}\xspace}
\newcommand{\rel}{\tagstyle{REL}\xspace}

\usepackage{adjustbox}
\usepackage{multirow}
\newcommand{\myparagraph}[1]{\noindent\textbf{#1}}

\newcommand{\ours}{FINER-Tuning\xspace}

\definecolor{cvprblue}{rgb}{0.21,0.49,0.74}
\usepackage[pagebackref,breaklinks,colorlinks,allcolors=cvprblue]{hyperref}

\def\paperID{}
\def\confName{CVPR}
\def\confYear{2026}

\title{FINER: MLLMs Hallucinate under Fine-grained Negative Queries}

\author{Rui Xiao\textsuperscript{1,2}, Sanghwan Kim\textsuperscript{1,2,3}, Yongqin Xian\textsuperscript{4}, Zeynep Akata\textsuperscript{1,2,3}, Stephan Alaniz\textsuperscript{5}\\[5pt]
\textsuperscript{1}Technical University of Munich  \quad \textsuperscript{2}Munich Center for Machine Learning \\
\textsuperscript{3}Helmholtz Munich \quad \textsuperscript{4}Google \quad \textsuperscript{5}LTCI, T\'el\'ecom Paris, Institut Polytechnique de Paris, France \\
}

\begin{document}
\maketitle
\begin{abstract}
Multimodal large language models (MLLMs) struggle with hallucinations, particularly with fine-grained queries, a challenge underrepresented by existing benchmarks that focus on coarse image-related questions. We introduce \textbf{FI}ne-grained \textbf{NE}gative que\textbf{R}ies (\textbf{FINER}), alongside two benchmarks: \textbf{FINER-CompreCap} and \textbf{FINER-DOCCI}. Using FINER, we analyze hallucinations across four settings: multi-object, multi-attribute, multi-relation, and ``what'' questions. Our benchmarks reveal that MLLMs hallucinate when fine-grained mismatches co-occur with genuinely present elements in the image. To address this, we propose \textbf{\ours}, leveraging Direct Preference Optimization (DPO) on FINER-inspired data. Finetuning four frontier MLLMs with \ours yields up to 24.2\% gains (InternVL3.5-14B) on hallucinations from our benchmarks, 
while simultaneously improving performance on eight existing hallucination suites, and enhancing general multimodal capabilities across six benchmarks. Code, benchmark, and models are available at \href{https://explainableml.github.io/finer-project/}{https://explainableml.github.io/finer-project/}.
\end{abstract}    
\begin{figure}[t]
  \centering
  \includegraphics[width=\linewidth]{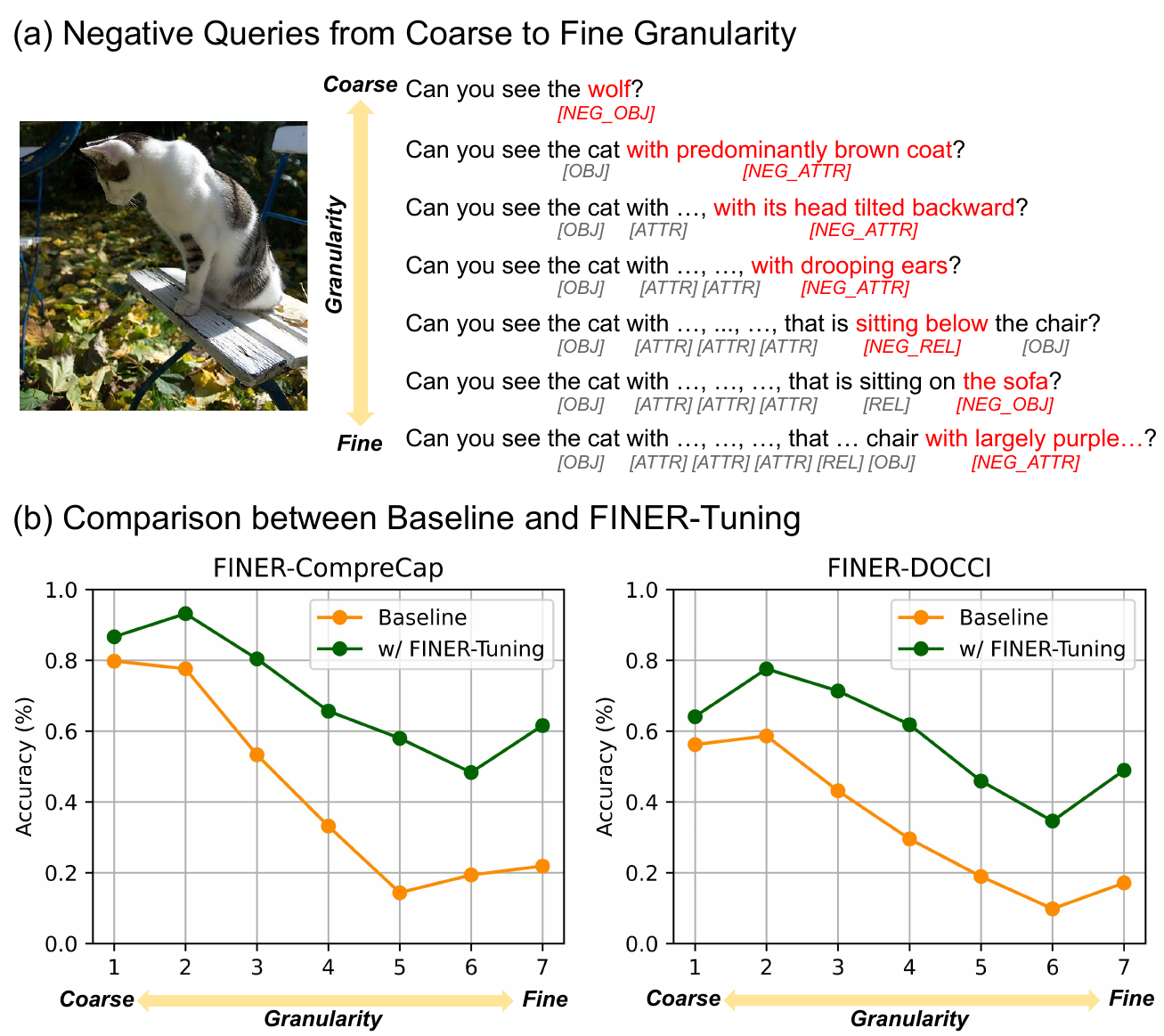}
  \caption{
    We compare the performance InternVL3.5-14B~\cite{wang2025internvl35} (Baseline) with the one fine-tuned by \ours under negative queries of seven different granularity levels.}
  \label{fig:teaser}
  \vspace{-2mm}
\end{figure}

\section{Introduction}
\label{sec:intro}
Multimodal large language models (MLLMs) have demonstrated significant progress in visual perception~\cite{achiam2023gpt} and instruction following~\cite{liu2023llava}, enabling increasingly sophisticated image question answering. Real-world users, however, often ask fine-grained questions requiring precise understanding of image content. While current models~\cite{liu2024llavanext, bai2025qwen25vl, wang2025internvl35} handle coarse questions reasonably well, it remains unclear whether they can detect nuanced errors in detailed user queries when describing image content. This is critical in domains like medical visual question answering, where trustworthiness requires spotting and correcting errors in complex queries. In the context of natural images, we focus on hallucination~\cite{rohrbach2018chair, bai2024hallucination}, the generation of answers unsupported by the image, and define ``negative queries'' as those asking about non-existent image content.
Prior studies show MLLMs often exhibit false-positive hallucination, failing to answer ``No'' to negative queries~\cite{li2023pope,augustin2025dash,wang2023amber,zhang2025mmmc}. Yet, these probes are largely coarse; POPE and DASH focus on \emph{single} object presence~\cite{li2023pope,augustin2025dash},
and AMBER includes only \emph{single} objects, attributes, and relations~\cite{wang2023amber}.
This raises an important question: \emph{Can MLLMs reject fine-grained mistakes involving multiple objects, attributes, and relations, rather than only coarse mismatches?} To investigate, we first conduct a motivation study, increasing the granularity of negative queries to probe for false positives.

\myparagraph{Question granularity affects hallucination.}
We examine how MLLMs behave as negative queries become progressively \emph{more fine-grained}. Mimicking how human constructs a sentence: starting with a single object, adding attributes, and then relations, we construct queries of increasing granularity from coarse to fine, as shown in Fig.~\ref{fig:teaser}. This yields seven levels, each injecting a single, fine-grained contradiction (\negobj, \negattr, or \negrel) while keeping the rest of the description visually consistent.  For each sample, we feed the model with the image and each of the seven queries separately, limiting the answer to ``Yes'' or ``No'', while the correct answer is always ``No''. We sample from two sources: 320 from \textsc{FINER-CompreCap} and 1{,}687 from \textsc{FINER-DOCCI}. We report averaged accuracy per level for \textsc{InternVL3.5-14B}~\cite{wang2025internvl35} and the model finetuned with \ours.

As shown in Fig.~\ref{fig:teaser}, the accuracy of \textsc{InternVL3.5-14B} steadily decreases with increased query granularity, dropping from $\sim80\%$ at level 1 to $\sim20\%$ by levels 5-7 on \textsc{FINER-CompreCap}, and from $\sim58\%$ at level 1 to $\sim15\%$ by levels 6-7 on \textsc{FINER-DOCCI}. This demonstrates the model’s brittleness to fine-grained negations: as granularity increases, it more often answers ``Yes'' to queries that should be ``No'', resulting in more false positives. The model finetuned with \ours, however, consistently demonstrates performance gains, particularly at finer granularity. This highlights MLLMs’ susceptibility to hallucination at finer granularity and the potential for improvement.

Hence, we ask: \emph{Can we systematically study hallucinations under fine-grained negative queries?} Our initial analysis mixes objects, attributes, and relations, hindering isolation of causal factors. To disentangle these, we introduce \textsc{FINER-CompreCap} and \textsc{FINER-DOCCI}, which group queries into four settings: multiple objects (Multi-obj), multiple attributes (Multi-attr), multiple relations (Multi-rel), and ``what''-questions (Wh). The first three target existence and binding, assessing whether the model can detect errors hidden in multiple objects, attributes, and relations. The Wh-setting probes factual answering with ill-posed queries, asking ``what''-questions about a target object with one incorrect attribute. Together, these four settings reveal whether a model can say ``No'' to precise but wrong claims, beyond handling coarse mismatches.
\section{FINER Benchmarks}

\begin{figure*}[t]
  \centering
  \includegraphics[width=\textwidth]{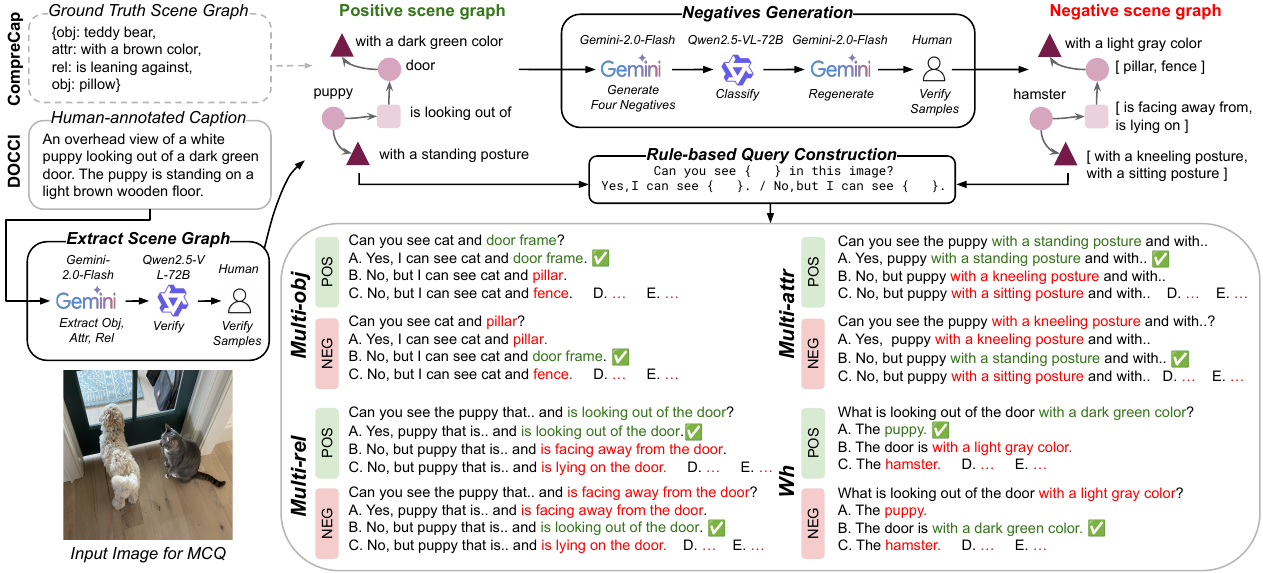}
  \caption{
    Data construction pipeline for FINER benchmarks. For \textsc{FINER-DOCCI}, we extract the positive scene graph (SG) from DOCCI~\cite{onoe2024docci} captions, while for \textsc{FINER-CompreCap}, the SG is provided by CompreCap~\cite{lu2025comprecap}. From the positive SG, we generate the negative SG using Qwen3-14B~\cite{yang2025qwen3} as negatives generator for \textsc{FINER-CompreCap} and Gemini-2.0-Flash~\cite{team2023gemini} for \textsc{FINER-DOCCI}. Finally, a rule-based query construction pipeline builds multiple choice questions. In practice, choices are shuffled in both benchmarks.
  }
  \label{fig:benchmark_construction}
  \vspace{-2mm}
\end{figure*}

Our FINER benchmarks aim to compose negative questions involving multiple semantic elements, i.e., objects, attributes, and relations, to evaluate an MLLM's ability to detect and reason about missing or incorrect components in a scene, even with subtle perturbations. We begin by explaining our benchmark construction as illustrated in Fig.~\ref{fig:benchmark_construction}.

\subsection{Question Construction Pipeline}
\label{sec:question_construction_pipeline}
We base our FINER benchmarks on the scene graph (SG) of an image, encoding objects (\obj), their attributes (\attr), and spatial or semantic relations (\rel). For each component, we generate negative counterparts (\negobj, \negattr, \negrel), semantically plausible but incorrect substitutions (e.g., replacing ``door frame'' with ``pillar''). Unlike prior work~\cite{li2023pope, augustin2025dash}, which rely on a single negative, we generate four distinct negative variants per entity (as described in Sec. \ref{sec:negatives}). The initial processing steps are visualized at the top of Fig.~\ref{fig:benchmark_construction}.

We then use a template-based approach to compose positive questions ($q^+$) mentioning multiple elements of the same category sampled from the positive SG. For example, a multiple-object question ($q^+_{\text{multi-obj}}$) might be ``Can you see cat and door frame?''. Corresponding negative questions ($q^-$) are constructed by replacing one randomly chosen element with a randomly sampled, negative counterpart (e.g., ``Can you see cat and pillar?''). The correct answers are ``Yes'' and ``No'' respectively. 
To move beyond binary responses, we construct Multiple Choice Questions (MCQs) requiring the model to specify the correct entities in the image. For example, the correct answer to $q^-_{\text{multi-obj}}$ would be ``No, but I can see cat and door frame''. 
We use the other negative options of the same component as distractors for the other answer options (see ``Multi-obj'' in Fig 2.).
Equivalently, we construct $q^\pm_{\text{multi-attr}}$ and $q^\pm_{\text{multi-rel}}$ from the SGs' attributes and relations. Finally, we create ``what''-questions (Wh) asking about an object in relation to another, using either its positive or negative attribute.
The complete question template is described in Sec.~B in the supplementary.

\myparagraph{Benchmarks.} Based on this pipeline, we constructed \textsc{FINER-CompreCap} (based on CompreCap~\cite{lu2025comprecap}) and \textsc{FINER-DOCCI} (based on DOCCI~\cite{onoe2024docci}). CompreCap provides human-annotated scene graphs, but is limited to COCO images. DOCCI consists of 5K images with long human-annotated captions which allow us to create a more large-scale question set. The detailed statistics of both benchmarks are in Sec.~B in the supplementary. \textsc{FINER-CompreCap} consists of 6,300 Multi-obj, 3,338 Multi-attr, 4,280 Multi-rel, and 3,166 Wh MCQs with a maximum of 6,3,3 objects, attributes, or relations per question. \textsc{FINER-DOCCI} comprises 10,000 Multi-obj, 28,630 Multi-attr, 11,542 Multi-rel, and 20,944 Wh MCQs with a maximum of 6,5,3 objects, attributes, or relations per question. In the following, we detail how we extract the SG from DOCCI, and how we generate the negative components.

\subsection{Scene Graph Extraction}
\label{sec:sg_extraction}

For DOCCI, where ground-truth SGs are unavailable, we build a non-panoptic SG by extracting objects, attributes, and relations directly from the human-written long captions. We use a multi-stage pipeline powered by Gemini-2.0-Flash~\cite{team2023gemini}, with filtering by a strong MLLM (Qwen2.5VL-72B~\cite{bai2025qwen25vl}) and human verification on sampled data, to convert captions into SG-like annotations. The validation steps reduce the risk of introducing incorrect features into the SG which is particularly important for \rel. We provide more details regarding the pipeline in Sec.~\ref{sec:supp_pos_sg_docci} in supplementary.

\subsection{Negatives Generation}
\label{sec:negatives}

Starting from the positive SGs, we generate four corresponding negatives for each object, attribute, and relation, using an LLM with carefully designed prompts. We use Qwen3-14B~\cite{yang2025qwen3} for FINER-CompreCap and Gemini-2.0-Flash~\cite{team2023gemini} for FINER-DOCCI to ensure consistency with the SG creation. To decrease the risk of generated negatives being present in the image, we use a strong MLLM (Qwen2.5-VL-72B) as a discriminator. If it fails to identify the positive item mixed into the negatives, we conclude that at least one negative is ambiguous or present in the image. Based on the MLLM's classification entropy, we identify which negatives require to be regenerated and repeat this process iteratively. Human verifies samples to specify regeneration thresholds.
For more details on the negatives generation, please refer to Sec.~\ref{sec:supp_negatives_generation_pipeline} in the supplementary.

\subsection{Evaluation Setting} 

As binary “Yes/No” responses are vulnerable to model biases, we use MCQs to move models beyond simple negation and enforce visual understanding, with each MCQ including one correct answer and four distractors. To prevent bias toward positive or negative answers, we pair each negative MCQ ($q^-$) with its corresponding positive MCQ ($q^+$), requiring both to be answered correctly. 
This pairing ensures models cannot succeed by simply memorizing “No” patterns or exploiting label imbalances. As a result, let $M(\cdot)$ be the model, we define paired accuracy as the primary evaluation metric for N paired questions of $q^+$ and $q^-$:
\begin{equation}
\mathrm{Acc}_{\text{paired}}
=\frac{1}{N}\!\sum_{i=1}^{N}\Gamma(M(x_i,q_i^+))\,\Gamma(M(x_i,q_i^-))
\end{equation}
where $\Gamma(\cdot)$ evaluates to 1 for correct responses and 0 otherwise. 
This metric requires success on both positive and negative variants, ensuring robustness against false positives and false negatives.

\begin{figure*}[t] 
  \centering
  \includegraphics[width=\textwidth]{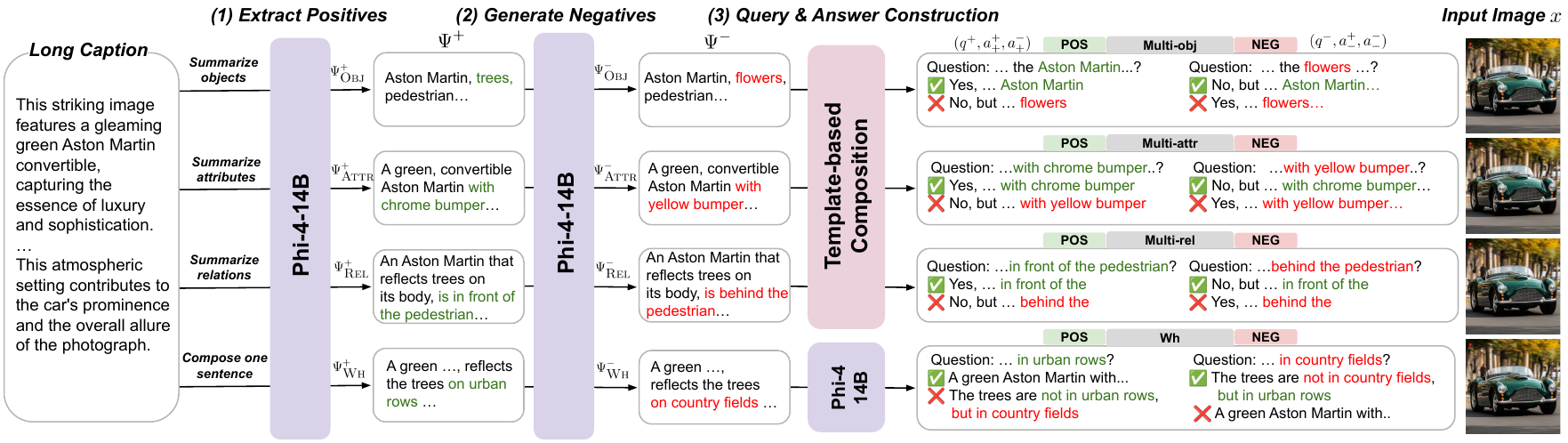}
  \caption{
    Training data generation pipeline for \ours. (1) We adopt long captions from Pixmo~\cite{deitke2025molmo} and extract diverse phrases with \textsc{PHI-4-14B}~\cite{abdin2024phi}. (2) We then prompt the same LLM to modify and generate negative phrases. (3) We construct both positive and negative query-answer tuples via template-based composition or LLM generation.
  }
  \label{fig:finer_dpo}
  \vspace{-2mm}
\end{figure*}

\section{Training with FINER (\ours)}
\label{sec:training_with_finer}
Observing MLLM vulnerabilities under FINER, we address them with a data-driven training approach via direct preference optimization (DPO)~\cite{rafailov2023direct} using \emph{fine-grained negative queries}, denoted as \ours. Unlike approaches optimizing for simple queries~\cite{zhao2023hadpo,yang2025opadpo,yu2024rlaifv}, \ours employs minimally edited, semantically precise contradictions over objects, attributes, and relations (e.g., ``car with yellow bumper'' vs. ``car with chrome bumper''), including both fine-grained positive and negative queries. Fig.~\ref{fig:finer_dpo} illustrates our training data generation pipeline. It is inspired by the four settings in our benchmarks with both accept and reject answers for every query. This focuses learning on detecting fine-grained hallucinations in the queries, rather than solely avoiding them in the model's responses.

\myparagraph{Setup.}
We select data \emph{avoiding in-distribution leakage}, excluding COCO data~\cite{lin2014microsoft}, and the DOCCI training split~\cite{onoe2024docci}. To leverage the availability of dense image annotations, we adopt Pixmo-caption~\cite{deitke2025molmo} as our base corpus. We further avoid using the LLMs used for benchmark construction, employing \texttt{Phi-4-14B}~\cite{abdin2024phi} for our training data pipeline.

\myparagraph{(1) Extract Positives.}
As illustrated in Fig.~\ref{fig:finer_dpo}, given a long caption, we prompt Phi-4-14B to extract fine-grained positive phrases, mirroring our four evaluation scenarios: Multi-obj, Multi-attr, Multi-rel, and Wh. We define the following four positive phrase types:
\begin{equation}
\label{eq:pos_phrases}
\Psi^+ \in \big\{\Psi_{\textsc{Obj}}^+,\ \Psi_{\textsc{Attr}}^+,\ \Psi_{\textsc{Rel}}^+,\ \Psi_{\textsc{Wh}}^+\big\}
\end{equation}

The LLM produces:
$\Psi_{\textsc{Obj}}^+$: a phrase summarizing the objects;
$\Psi_{\textsc{Attr}}^+$: a phrase summarizing attributes for a random object;
$\Psi_{\textsc{Rel}}^+$: a phrase summarizing relations between a random object and others;
$\Psi_{\textsc{Wh}}^+$: a composed sentence describing two objects with a relation and summarized attributes, subsequently forming a positive question-answer pair. Our prompt templates are detailed in Sec.~\ref{sec:supp_templates}.

\myparagraph{(2) Generate Negatives.}
Transforming the positive phrases $\Psi^+$, we generate negative phrases $\Psi^-$ with the same LLM:
\begin{equation}
    \label{eq:neg_phrases}
    \Psi^-\in\big\{\Psi_{\textsc{Obj}}^-,\ \Psi_{\textsc{Attr}}^-,\ \Psi_{\textsc{Rel}}^-,\ \Psi_{\textsc{Wh}}^-\big\}
\end{equation}

For each phrase type $ \Psi_{\textsc{T}}^+ $ (where $ \textsc{T} \in \{\textsc{Obj}, \textsc{Attr}, \textsc{Rel}, \textsc{Wh}\} $), we randomly select one instance of $ \textsc{T} $, and prompt the LLM to replace that instance with a negative, forming $ \Psi_{\textsc{T}}^- $.
Please refer to Sec.~E for the complete prompt details.

\myparagraph{(3) Query \& Answer Construction.}
With $\Psi^+$ and $\Psi^-$, we construct query-answer pairs for DPO training, including both positive ($q^+$) and negative ($q^-$) questions paired with accepted ($a^+$) and rejected ($a^-$) responses. $a^+$ begins with the correct response ("Yes" for $q^+$, "No" for $q^-$) and mentions the correct image features, while $a^-$ is the opposite.

For \textsc{Obj}/\textsc{Attr}/\textsc{Rel}, we directly use question-answer templates on $\Psi^+$ and $\Psi^-$ to construct $(q^+, a^+_+, a^-_+)$ and $(q^-, a^+_-, a^-_-)$ pairs. We use five templates to avoid overfitting to the benchmark’s prompt pattern, as detailed in Sec.~G. For \textsc{Wh}, data pairs are already constructed by the LLM due to the free-form nature of these questions and answers. Fig.~\ref{fig:finer_dpo} provides example data for all data types and more examples are provided in Sec.~\ref{sec:supp_training_details} in the supplementary.

\myparagraph{DPO Training.} This creates a dataset of preference tuples
\begin{equation}
    \label{eq:pref_tuple}
    \mathcal{D}=\{(x,q^s,a^+_s,a^-_s)\}, s \in \{ +, -\}
\end{equation}
where $x$ is the image. Let $\pi_\theta(\cdot\mid x,q)$ be the policy and $\pi_{\mathrm{ref}}$ be a frozen reference model. We train with DPO, maximizing the probability that the policy ranks $a^+$ above $a^-$:
\begingroup
\small
\setlength{\jot}{1pt}
\begin{equation}\label{eq:dpo-core}
\begin{aligned}
\Delta_\theta(x,q) &:= \log \pi_\theta(a^+\!\mid\!x,q) - \log \pi_\theta(a^-\!\mid\!x,q),\\
\Delta_{\mathrm{ref}}(x,q) &:= \log \pi_{\mathrm{ref}}(a^+\!\mid\!x,q) - \log \pi_{\mathrm{ref}}(a^-\!\mid\!x,q),\\
\mathcal{L}_{\mathrm{DPO}}(\theta) &= -\,\mathbb{E}_{(x,q,a^+,a^-)\sim\mathcal{D}}
\Big[\log\sigma\!\big(\beta(\Delta_\theta-\Delta_{\mathrm{ref}})\big)\Big].
\end{aligned}
\end{equation}
\endgroup
where $\sigma(\cdot)$ is the logistic function and
$\beta=0.1$.

\section{Experiments}
\label{sec:exp}
We present experiments of \ours on
three tasks, i.e., evaluation on FINER benchmarks (Sec.~\ref{sec:exp_finer_benchmarks}), other hallucination benchmarks (Sec.~\ref{sec:exp_other_hallu}), and general MLLM capabilities (Sec.~\ref{sec:exp_general_cap}). In addition, we show qualitative examples on FINER benchmarks (Sec.~\ref{sec:exp_qualitative_results}), and ablate important training strategies and subset selections (Sec.~\ref{sec:exp_ablation_studies}).

\begin{table*}[t]
  \caption{Paired accuracy ($\text{Acc}_{\text{paired}}$) results on FINER-CompreCap and FINER-DOCCI. $^*$For Gemini-2.5-Flash, we evaluate on the whole \textsc{FINER-CompreCap} and on 3K MCQs per setting in \textsc{FINER-DOCCI} due to the scale of the benchmark.}
  \label{tab:main}
  \centering
  \resizebox{0.85\textwidth}{!}{
  \setlength{\tabcolsep}{5pt}
  \begin{tabular}{@{}lccccccccc@{}}
    \toprule
    & & \multicolumn{4}{c}{FINER-CompreCap} & \multicolumn{4}{c}{FINER-DOCCI} \\
    \cmidrule(lr){3-6} \cmidrule(lr){7-10}
    Models & Size &Multi-obj & Multi-attr & Multi-rel & Wh & Multi-obj & Multi-attr & Multi-rel & Wh \\
    \midrule
    \baserow \textit{Random Guess} &  - & 4.0 & 4.0 & 4.0 & 4.0 & 4.0 & 4.0 & 4.0 & 4.0 \\
    LRV-V2~\cite{liu2023lrv} &  13B & 6.1 & 6.8 & 5.6 & 4.0 & 6.3 & 5.4 & 6.1 &  5.2 \\
     LLaVA-RLHF~\cite{sun2023llava-rlhf}  &  13B & 11.4 & 2.0 & 1.1 & 6.9 & 7.3 & 3.0 & 5.1 & 5.3 \\
     RLHF-V~\cite{yu2023rlhfv}  &  13B & 13.4 & 6.1 & 1.6 & 10.8 & 13.2 & 7.2 & 8.1 & 7.0 \\
      OPA-DPO~\cite{yang2025opadpo}  &  13B & 10.9 & 3.0 & 2.2 & 6.9 & 8.1 & 5.5 & 8.3 & 8.0 \\
     RLAIF-V~\cite{yu2024rlaifv}  &  12B & 62.2 & 39.6 & 19.2 & 20.5 & 46.5 & 31.7 & 32.4 & 19.4 \\
    \hline
    LLaVA-1.6~\cite{liu2024llavanext} &  7B  & 25.3 & 13.0 & 7.6 & 15.3 & 10.1 & 12.3 & 8.2 & 13.3 \\
    \dporow \quad +\ours  & 7B
      & 48.4\dplus{23.1} & 38.4\dplus{25.4} & 24.2\dplus{16.6} & 22.1\dplus{6.8}
      & 26.4\dplus{16.3} & 29.4\dplus{17.1} & 24.7\dplus{16.5} & 18.5\dplus{5.2} \\
    Qwen2.5-VL~\cite{bai2025qwen25vl} & 7B  & 69.2 & 62.5 & 30.1 & 28.9 & 48.7 & 47.5 & 36.7 & 23.4 \\
    \dporow \quad +\ours  & 7B
      & 71.4\dplus{2.2} & 67.0\dplus{4.5} & 38.3\dplus{8.2} & 34.8\dplus{5.9}
      & 49.8\dplus{1.1} & 52.2\dplus{4.7} & 43.4\dplus{6.7} & 28.0\dplus{4.6} \\
    InternVL-3.5~\cite{wang2025internvl35} & 8B & 75.0 & 72.5 & 49.8 & 23.5 & 58.1 & 54.3 & 41.8 & 16.8 \\
    \dporow \quad +\ours   &  8B
      & 77.1\dplus{2.1} & 78.9\dplus{6.4} & 64.1\dplus{14.3} & 34.2\dplus{10.7}
      & 62.6\dplus{4.5} & 60.1\dplus{5.8} & 52.7\dplus{10.9} & 23.7\dplus{6.9} \\
    InternVL-3.5~\cite{wang2025internvl35} & 14B & 74.5 & 68.1 & 47.0 & 21.8 & 58.6 & 55.9 & 41.4 & 15.6 \\
    \dporow \quad +\ours   &  14B
      & 80.0\dplus{5.5} & 78.9\dplus{10.8} & 71.2\dplus{24.2} & 30.1\dplus{8.3}
      & 65.9\dplus{7.3} & 65.0\dplus{9.1} & 57.0\dplus{15.6} & 23.0\dplus{7.4} \\
    \hline
    \baserow InternVL-3.5~\cite{wang2025internvl35}  &  38B
      &  77.8 & 78.1 & 66.8 & 50.9 & 62.3 & 64.8 & 54.2 &  36.6  \\
    \baserow Gemini-2.5-Flash~\cite{comanici2025gemini25}$^*$  &  -
      &  75.7 & 77.3 & 77.8 & 58.2 & 64.4 & 64.5 & 56.7 & 49.6   \\
    \bottomrule
  \end{tabular}
  }
  \vspace{-1mm}
\end{table*}

\subsection{Experimental Setup}
\label{sec:exp_exp_setup}
\myparagraph{Fine-tuning Setup.}
We are interested in applying \ours to frontier-MLLMs: LLaVA-NeXT-7B (LLaVA-1.6-7B)~\cite{liu2024llavanext}, Qwen2.5-VL-7B-Instruct~\cite{bai2025qwen25vl}, and InternVL-3.5-8B~\cite{wang2025internvl35}. To test scalability within our compute limits, we also include InternVL-3.5-14B~\cite{wang2025internvl35}. We fine-tune each model on our constructed data with maximally 160k preference tuples. All models are trained for one epoch using LLaMA-Factory~\cite{zheng2024llamafactory} with LoRA~\cite{hu2022lora}. Full training details are in Sec.~\ref{sec:supp_training_details} in the supplementary.

\myparagraph{Evaluation Setup.}
We evaluate all models on three tasks across 16 benchmarks. We primarily use VLMEvalKit~\cite{duan2024vlmevalkit} for standardized evaluations. For benchmarks not integrated in VLMEvalKit, we follow each benchmark’s official evaluation protocol. Refer to Sec.~\ref{sec:supp_evaluation_details} in supplementary for details.

\subsection{Results on FINER benchmarks}
\label{sec:exp_finer_benchmarks}
\myparagraph{Baselines.}
We primarily compare the performance of the four frontier MLLMs before and after \ours, and also show the performance of stronger models such as InternVL-3.5-38B and Gemini-2.5-Flash~\cite{team2023gemini}. Additionally, we benchmark hallucination-aware fine-tuning methods such as RLAIF-V~\cite{yu2024rlaifv}, OPA-DPO~\cite{yang2025opadpo}, RLHF-V~\cite{yu2023rlhfv}, Llava-RLHF~\cite{sun2023llava-rlhf}, and LRV-Instruct-V2~\cite{liu2023lrv}. Note that different methods are typically based on different MLLMs and fine-tuned on different data. 
Given their effectiveness on general hallucination reduction, we aim to find out how well they fare on our FINER benchmarks. Furthermore, we estimate human performance with a human study on a subset of 20 MCQs for each setting. The results and details of our human study can be found in Sec.~\ref{sec:supp_human_study} in the supplementary.

\myparagraph{Main results.} The results are presented in Tab.~\ref{tab:main}.
Base model capability strongly influences overall performance. Hallucination-aware fine-tuning methods like RLHF-V~\cite{yu2023rlhfv} and LLaVA-RLHF~\cite{sun2023llava-rlhf} only achieve 1.6\% and 1.1\% paired accuracy on the Multi-rel subset of \textsc{FINER-CompreCap}. RLAIF-V-12B, while remaining the best among these methods, scores substantially below advanced MLLMs, including Qwen2.5-VL and InternVL-3.5. This shows that mitigating hallucination on previous datasets do not directly translate to our FINER benchmarks, highlighting the importance to start from and improve upon frontier MLLMs.

Meanwhile, \ours consistently improves all baselines. Specifically, on \textsc{FINER-CompreCap}, LLaVA-1.6 shows remarkable 23.1\% and 25.4\%, and 16.6\% on Multi-obj, Multi-Attr and Multi-Rel subsets, and InternVL-3.5-14B shows improvements of up to 24.2\% (Multi-rel), outperforming its 38B version by 4.4\%. On \textsc{FINER-DOCCI}, \ours on InternVL-3.5-14B scores on-paar with Gemini-2.5-Flash in 3 out of 4 settings.
Moreover, Wh-questions challenge all models. Even InternVL-3.5-38B and Gemini-2.5-Flash achieve only 36.6\% and 49.6\% $\text{Acc}_{\text{paired}}$ on \textsc{FINER-DOCCI}, leaving room for future research on reducing hallucinations in FINER.

\begin{figure*}[t]
  \centering
  \resizebox{0.9\linewidth}{!}{%
    \includegraphics{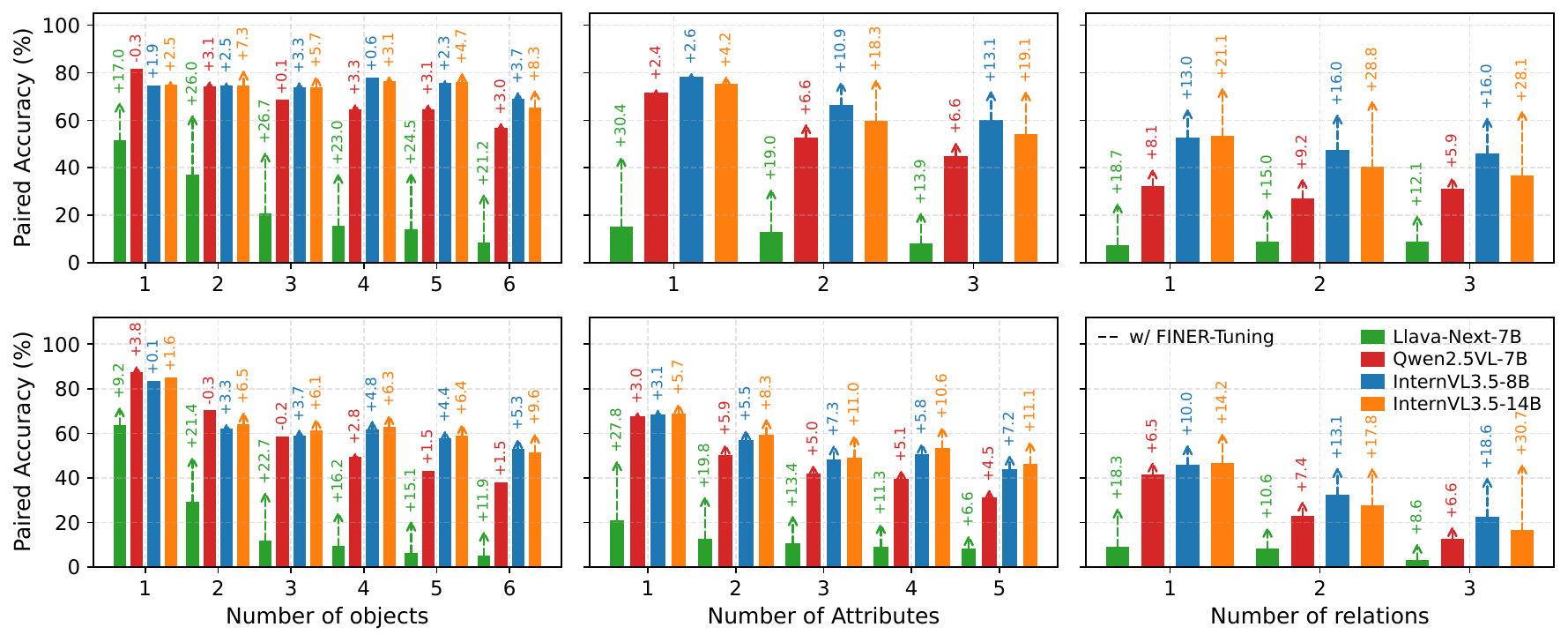}%
  }
  \caption{$\text{Acc}_\text{paired}$ versus the number of objects, attributes, and relations.
  Top: \textsc{FINER-CompreCap}; Bottom: \textsc{FINER-DOCCI}. Dashed arrows show the gain from \ours.}
  \label{fig:length_bias}
  \vspace{-1mm}
\end{figure*}

\myparagraph{Different number of objects, attributes and relations.}  
Both FINER benchmarks cover Multi-obj, Multi-attr, and Multi-rel settings. We study how $\text{Acc}_{\text{paired}}$ changes as the number of entities increases (Fig.~\ref{fig:length_bias}). Models show similar trends in all three settings: performance drops as the entity counts increases, with much smaller drops in Multi-obj. \ours consistently improves performance, with larger gains in Multi-attr and Multi-rel, and the gains grow with higher counts. For example, \ours improves InternVL3.5-14B by 8.3\%, 19.1\% and 28.1\% in 6-obj, 3-attr and 3-rel setting on \textsc{FINER-CompreCap}.

\begin{table*}[t]
  \caption{Results on hallucination benchmarks including discriminative (DASH~\cite{augustin2025dash}, POPE~\cite{li2023pope}, RePOPE~\cite{neuhaus2025repope}, HallusionBench~\cite{guan2024hallusionbench}, AMBER~\cite{wang2023amber}, CRPE\_R~\cite{wang2024allseeing_v2}) and generative ones (MMHal-Bench~\cite{sun2023llava-rlhf}, HaloQuest~\cite{wang2024haloquest}). Sc.:Score (max. 6); HR.: Hallucination Rate.}
  \label{tab:other_hallu_bench}
  \centering
  \small
  \setlength{\tabcolsep}{2.2pt} 
  \renewcommand{\arraystretch}{1.02} 
  \begin{adjustbox}{max width=\linewidth} 
  \begin{tabular}{@{}l c c c c c c c | cc c@{}}
    \toprule
    & & \multicolumn{1}{c}{DASH} & \multicolumn{1}{c}{POPE} & \multicolumn{1}{c}{RePOPE} & \multicolumn{1}{c}{HallBench} & \multicolumn{1}{c}{AMBER} & \multicolumn{1}{c}{CRPE\_R} &\multicolumn{2}{c}{MMHal-Bench}  & \multicolumn{1}{c}{HaloQuest} \\
    \cmidrule(lr){3-3} \cmidrule(lr){4-4} \cmidrule(lr){5-5}
    \cmidrule(lr){6-6} \cmidrule(lr){7-7} \cmidrule(lr){8-8} \cmidrule(lr){9-10} \cmidrule(lr){11-11}
    Models & Size & Acc.\ {\small$\uparrow$} & Acc. \ {\small$\uparrow$} & Acc.\ {\small$\uparrow$} & aAcc.\ {\small$\uparrow$} & Acc.\ {\small$\uparrow$} & Acc.\ {\small$\uparrow$} & Sc.\ {\small$\uparrow$} & HR.\ {\small$\downarrow$} & Sc.\ {\small$\uparrow$} \\
    \midrule
    OmniLMM~\cite{omnilmm}  & 12B & 79.0 & 88.0 & 93.8 & 54.9 & 86.9 & 51.7 & 3.5 & 34.0 & 39.9\\
\baserow \quad +RLAIF-V~\cite{yu2024rlaifv}  & 12B
& 76.3\dminus{2.7} & 87.7\dminus{0.3} & 93.4\dminus{0.4}
& 53.7\dminus{1.2} & 87.4\dplus{0.5} & 52.2\dplus{0.5} & 4.0\dplus{0.5} & 29.0\dplus{5.0} & 62.4\dplus{22.5} \\

    \hline
    LLaVA-1.6~\cite{liu2024llavanext}  & 7B
      & 58.0  & 88.2 & 92.3 & 33.0  & 78.1 & 56.5 & 3.3 & 43.0  & 44.2 \\
    \dporow \quad +\ours   &  7B
      & 57.4\dminus{0.6}   & 88.8\dplus{0.6} & 93.2\dplus{0.9} 
      & 36.3\dplus{3.3} & 85.0\dplus{6.9} & 56.0\dminus{0.5} & 3.5\dplus{0.2} & 40.0\dplus{3.0}  & 63.5\dplus{19.3} \\
    Qwen2.5-VL~\cite{bai2025qwen25vl}  &  7B
      & 74.6  & 86.4 & 92.4
      & 65.4  & 85.2  & 69.9 & 4.6 & 18.0  & 74.8\\
    \dporow \quad +\ours  &  7B
      & 76.6\dplus{2.0} & 87.2\dplus{0.8} & 92.8\dplus{0.4}
      & 68.5\dplus{3.1} & 85.8\dplus{0.6} & 70.7\dplus{0.8} & 4.7\dplus{0.1} & 15.0\dplus{3.0} & 80.8\dplus{6.0} \\
    InternVL-3.5~\cite{wang2025internvl35}  &  8B
      & 68.3  & 88.6 & 91.5
      & 71.0  & 88.2 & 67.7 & 4.5 & 19.0  & 62.4 \\
    \dporow \quad +\ours     &   8B
      & 74.5\dplus{6.2}  & 89.4\dplus{0.8} & 93.1\dplus{1.6}
      & 73.0\dplus{2.0} & 88.6\dplus{0.4} & 68.0\dplus{0.3} & 4.6\dplus{0.1}  & 14.0\dplus{5.0}  & 73.5\dplus{11.1} \\
     InternVL-3.5~\cite{wang2025internvl35}  &  14B
      & 55.8  & 89.5 & 91.8
      & 69.5  & 88.0 & 67.2 & 4.7 & 11.0  & 65.0   \\
    \dporow \quad +\ours     &   14B
      & 61.3\dplus{5.5}  & 90.2\dplus{0.7} & 93.6\dplus{1.8}
      & 71.2\dplus{1.7} & 89.4\dplus{1.4} & 69.0\dplus{1.8} & 4.7 & 10.0\dplus{1.0}  & 71.0\dplus{6.0} \\
    \bottomrule
  \end{tabular}
  \end{adjustbox}
  \vspace{-1mm}
\end{table*}

\subsection{Results on other hallucination benchmarks}
\label{sec:exp_other_hallu}
\ours achieves consistent improvements on FINER benchmarks. Hence, we are interested how well models fine-tuned with \ours generalize to other hallucination benchmarks.
Additionally, we show the performance of RLAIF-V-12B against its baseline model OmniLMM-12B~\cite{omnilmm}, to see whether other hallucination reduction methods achieve balanced improvements across various hallucination benchmarks.
We evaluate models on both discriminative benchmarks like DASH~\cite{augustin2025dash}, POPE~\cite{li2023pope}, RePOPE~\cite{neuhaus2025repope}, HallusionBench~\cite{guan2024hallusionbench}, AMBER~\cite{wang2023amber}, CRPE relation split (CRPE\_R)~\cite{wang2024allseeing_v2}, as well as generative benchmarks like MMHalBench~\cite{sun2023llava-rlhf} and HaloQuest~\cite{wang2024haloquest}. The summarized results are shown in Tab.~\ref{tab:other_hallu_bench}. In supplementary, We further include detailed breakdowns (Tabs.~\ref{tab:supp_other_hallu_bench} and \ref{tab:supp_other_hallu2}), results for AMBER \textit{generative} (Tab.~\ref{tab:supp_ext_results_amberg}) and comparisons with more methods (Tab.~\ref{tab:extended_comparison}). 
Intuitively, \ours strengthens discrimination through FINER training; our results on discriminative benchmarks confirm this. \ours consistently improves Qwen2.5-VL and InternVL-3.5 across all benchmarks. On DASH, it boosts the two InternVL-3.5 variants by 6.2\% and 5.5\%. LLaVA-1.6 also gains 6.9\% on AMBER with \ours. \ours further reduces hallucination on generative benchmarks. On MMHal-Bench, it lowers hallucination rate for all base models, reaching 10\% with InternVL-3.5-14B. On HaloQuest, it improves LLaVA-1.6 by 19.3\%. Even for Qwen2.5-VL and InternVL-3.5, we observe at least 6\% gains. In contrast, while RLAIF-V delivers strong gains on generative benchmarks, its improvements on discriminative tasks are less consistent, where \ours benefits both. RLAIF-V degrades performance compared to the base OmniLMM on benchmarks like DASH, POPE, RePOPE, and HallusionBench. By comparing these ``deltas" between fine-tuned models and baselines, we show that \ours is a \textit{balanced} approach that leads to a comprehensive reduction in hallucination. These results also validate the effectiveness of FINER benchmarks, showing that improvements on FINER benchmarks align with broader improvements in other benchmarks as well.

\subsection{Results on general capabilities}
\label{sec:exp_general_cap}
Since \ours adds fine-grained negative queries to DPO, a natural concern is \textit{over-rejection}: the model becoming overly cautious, refusing answerable questions, or regressing on existing skills. To test this, we compare each base model and its \ours-tuned counterpart on six additional benchmarks: MMStar~\cite{chen2024mmstar} (general abilities), TextVQA~\cite{singh2019textvqa}, ChartQA~\cite{masry2022chartqa}, MMVP~\cite{tong2024eyes} (vision-centric abilities), NaturalBench~\cite{li2024naturalbench} (compositionality), and V$^*$ (visual search). The results are shown in Tab.~\ref{tab:other_benchmarks}. Unlike prior work reporting an ``alignment tax'', with gains on target benchmarks at the cost of general ability~\cite{zhang2025mmmc}, \ours avoids this trade-off and even improves strong baselines on general benchmarks (improving InternVL3.5-14B by 1.4\%). This shows that FINER provides a useful training signal that complements the model’s internal capabilities.

\begin{table}[t]
   \caption{Results on six general purpose MLLM benchmarks. M.S.: MMStar~\cite{chen2024mmstar}; Text: TextVQA~\cite{singh2019textvqa}; Chart: ChartQA~\cite{masry2022chartqa}; M.P.: MMVP~\cite{tong2024eyes}; N.B.: NaturalBench~\cite{li2024naturalbench}; V$^{*}$: V$^{*}$ Bench~\cite{wu2024vstar}}
   \label{tab:other_benchmarks}
  \centering
  \resizebox{0.95\linewidth}{!}{
  \setlength{\tabcolsep}{3pt}
  \renewcommand{\arraystretch}{1.05}
  \begin{tabular}{@{}l c c c c c c c@{}}
    \toprule
    Models & M.S.  & Text & Chart & M.P. & N.B.  & V$^{*}$ & Avg.\\
    \midrule
    OmniLMM\text{-}12B  & 39.7  & \textbf{64.5} & 24.2 & 69.7 & \textbf{26.9} & 52.9 & \textbf{46.3} \\
    \baserow \quad +RLAIF-V    & \textbf{40.9 } & \textbf{64.5} & \textbf{25.1} & \textbf{70.0} & 19.4 & \textbf{54.4} & 45.7 \\
    LLaVA-1.6-7B  & 37.6  & 63.7 & 54.4 & 65.0 & 15.7 & 53.9 & 48.4 \\
    \dporow \quad +\ours    & \textbf{39.2}  & \textbf{63.9} & \textbf{54.9} & \textbf{68.7} &  \textbf{19.8} & \textbf{55.0} & \textbf{50.3} \\
    Qwen2.5\text{-}VL-7B  & 63.7  & 84.9 & \textbf{87.0 }& 76.7 & \textbf{34.1}  &  72.7 & 69.8 \\
    \dporow \quad +\ours    & \textbf{64.7}  & \textbf{85.1} & 86.4 &\textbf{77.3} & \textbf{34.1}  &  \textbf{72.8} & \textbf{70.1} \\
     InternVL3.5-8B  & 68.0  &  77.8 & \textbf{86.7} & 76.7 &  30.4 & 69.1 & 68.1 \\
    \dporow \quad +\ours & \textbf{68.3}  & \textbf{77.9} & \textbf{86.7 }& \textbf{77.0} & \textbf{31.1}  &  \textbf{71.2} & \textbf{68.7} \\
    InternVL3.5-14B  & 67.2  & \textbf{77.2} & 86.4 & 78.3 & 30.7  &  68.0 & 68.0 \\
    \dporow \quad +\ours & \textbf{67.7}  & \textbf{77.2} & \textbf{86.8} & \textbf{78.7} &  \textbf{35.5}  & \textbf{70.2} & \textbf{69.4} \\
    \bottomrule
  \end{tabular}
  }
  \vspace{-1mm}
\end{table}

\begin{figure*}[t] 
  \centering
  \includegraphics[width=.95\textwidth]{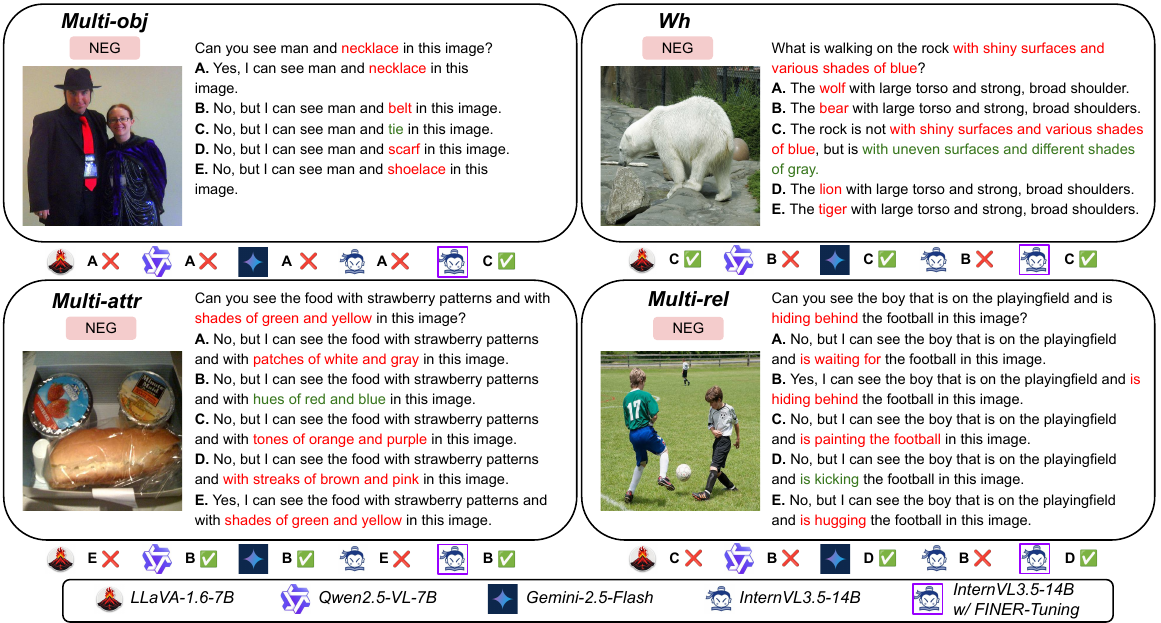}
  \caption{
   Qualiative examples of FINER-CompreCap MCQs for each category together with MLLM answers.}
  \label{fig:qualitative_results}
  \vspace{-2mm}
\end{figure*}


\subsection{Qualitative Results}
\label{sec:exp_qualitative_results}
Figure~\ref{fig:qualitative_results} shows four \textsc{FINER-CompreCap} examples; more qualitative results, including \textsc{FINER-DOCCI}, are in Sec.~E in the supplementary. \ours avoids the spurious “necklace” in the Multi-obj case and correctly identifies the fine color details of the strawberry-patterned food in the Multi-attr case. In the Multi-rel example, both Qwen2.5-VL and InternVL3.5 hallucinate the second relation as ``hiding behind the football". In the Wh example, \ours shifts InternVL-3.5-14B from answering “bear” to flagging the incorrect attribute of the rock. These examples indicate that \ours helps the model detect fine-grained errors and locate correct the information in complex queries.

\subsection{Ablation Studies}
\label{sec:exp_ablation_studies}
\myparagraph{Training strategies.}
\ours trains on both positive and negative queries $\{(x, q^+, a^+_+, a^-_+),\ (x, q^-, a^+_-, a^-_-)\}$. To ablate this setting, we investigate the training with and without positive questions, and compare the performance of DPO against supervised fine-tuning (SFT).





We train four InternVL-3.5-8B variants accordingly and compare with the baseline in Tab.~\ref{tab:ablation_study}. Results show mixed outcomes for SFT: with both queries, SFT reduces Multi-obj performance by 36.7\% relative to the baseline. DPO with only negative queries exceeds the base model but still lags behind DPO with both query types (\ours), underscoring the value of training with both.

\begin{table}[t]
   \caption{Ablation study on different training strategies. SFT methods only use $a^+$. The base model is InternVL-3.5-8B~\cite{wang2025internvl35}. Q.Type: Query Type;
   M.S.: MMStar~\cite{chen2024mmstar}}
   \label{tab:ablation_study}
  \centering
  \resizebox{0.9\linewidth}{!}{
  \setlength{\tabcolsep}{3pt}
  \renewcommand{\arraystretch}{1.05}
  \begin{tabular}{@{}l c c c c c c c c@{}}
    \toprule
    \multicolumn{1}{@{}l}{Method} & \multicolumn{2}{c}{Q.Type} & \multicolumn{4}{c}{FINER-CompreCap} & \multicolumn{2}{c}{Other}  \\
    \cmidrule(lr){2-3} \cmidrule(lr){4-7} \cmidrule(lr){8-9}
    & Neg & Both & Obj & Attr & Rel & Wh & RePOPE & M.S. \\
    \midrule
    \baserow Base  & - & - &  74.2 & 71.9 &  49.8 &  25.5  & 91.5 & 68.0   \\
      +SFT  & \checkmark &  &  47.4  & 59.7 & 53.8 & \textbf{38.7}  & 69.1 & 61.7  \\
       +SFT  & & \checkmark &  37.5  & 49.5 & 55.2 & 18.9  & 92.2 & 63.3  \\
       +DPO  & \checkmark &  &  75.8 & 75.2 & 52.4 & 29.8  & \textbf{93.1} & \textbf{68.3}  \\
    \dporow +DPO  &  & \checkmark & \textbf{76.5} & \textbf{78.3} & \textbf{64.1} & 36.1  & \textbf{93.1} & \textbf{68.3}  \\
    \bottomrule
  \end{tabular}
  }
  \vspace{-1mm}
\end{table}

\myparagraph{Training on subsets.}
Our training data matches the benchmark query types: Multi-Obj, Multi-Attr, Multi-Rel, and Wh. We train InternVL-3.5-8B on each subset separately and compare to \ours trained on all subsets, keeping the total number of training samples fixed at 160k. As shown in Tab.~\ref{tab:subset_training}, models trained only on Multi-Obj, Multi-Rel, or Wh achieve the best scores on their corresponding tests. Notably, they also improve on other settings, suggesting the model is not merely echoing supervision from data: FINER fosters a more general rejection pattern that transfers beyond the seen subset. Overall, training on all subsets yields the most balanced results.

\begin{table}[t]
 \caption{Training-on-subset ablation for \ours with InternVL-3.5-8B~\cite{wang2025internvl35}. Obj/Attr/Rel denote Multi-obj/Multi-attr/Multi-rel for both training and evaluation.}
  \label{tab:subset_training}
  \centering
  \resizebox{0.8\linewidth}{!}{
  \setlength{\tabcolsep}{3pt}
  \renewcommand{\arraystretch}{1.05}
  \begin{tabular}{@{}l c c c c c c@{}}
    \toprule
    \multicolumn{1}{@{}l}{Train Subset} & \multicolumn{4}{c}{FINER-CompreCap} & \multicolumn{2}{c}{Other} \\
    \cmidrule(lr){2-5} \cmidrule(lr){6-7}
    & Obj & Attr & Rel & Wh & RePOPE & M.S. \\
    \midrule
     \baserow Base  &  74.2 & 71.9 &  49.8 &  25.5 & 91.5 & 68.0   \\
     Obj    &  \textbf{78.8}  & 76.4 & 54.2 & 28.7   & \textbf{93.5} & 67.9 \\
     Attr   &  71.3  & 76.7 & 56.8 & 26.5  & 91.5 & 68.2 \\
     Rel   &  69.2 & 73.0 & \textbf{66.7} & 24.1  & 91.4 & 67.7 \\
     Wh  & 75.9  & 75.3 & 55.0 & \textbf{46.5} & 92.9 & \textbf{68.3}  \\
     \dporow All  &76.5  & \textbf{78.3} & 64.1 & 36.1  & 93.1 & \textbf{68.3}  \\
    \bottomrule
  \end{tabular}
  }
  \vspace{-1mm}
\end{table}

\section{Related Works}
\label{sec:related_works}
\myparagraph{Hallucination Benchmarks.}
POPE~\cite{li2023pope} probes object hallucination by asking yes-or-no questions. RePOPE~\cite{neuhaus2025repope} identifies and corrects annotation errors in POPE. Amber~\cite{wang2023amber} categorizes hallucinations into “object,” “relation,” and “attribute” types in its discriminative subset. A common limitation of these benchmarks is their reliance on the MSCOCO dataset~\cite{lin2014microsoft}. Therefore, DASH~\cite{augustin2025dash} applies retrieval to select challenging images from LAION-5B~\cite{relaion}. CRPE~\cite{wang2024allseeing_v2} focuses on relation hallucinations but is limited to single-relation cases. 
NOPE~\cite{lovenia2024nope} targets non-existent objects, not attribute or relation hallucinations. ROPE~\cite{chen2024multiobject} probes object classes with visual prompts (bounding boxes). Unlike ROPE, our Multi-obj setting randomly replaces a positive object with a negative one and does not rely on MSCOCO/ADE20K box annotations~\cite{lin2014microsoft, zhou2017ade20k}.
MMHal-Bench~\cite{sun2023llava-rlhf} evaluates hallucination via eight types of questions with limited scale. HaloQuest~\cite{wang2024haloquest} includes a “false premise” subset with a similar motivation to our Wh setting. However, our setting differs: we target false premises in fine-grained attributes of existing objects, whereas HaloQuest primarily targets non-existent objects.

\myparagraph{Hallucination-aware Fine-tuning.}
Prior work reduces hallucinations via supervised or contrastive tuning and instruction-based data augmentation: LRV-Instruct~\cite{liu2023lrv} adds negative instructions to MiniGPT-4~\cite{zhu2023minigpt} and mPLUG-Owl~\cite{ye2023mplug}; HALVA~\cite{sarkar2024halva} builds paired correct vs.~hallucinated responses for contrastive learning; PerturboLLaVA~\cite{chen2025perturbollava} trains under misleading contexts; REVERSE~\cite{wu2025reverse} adds uncertainty tokens and retrospective reasoning. Other studies use preference learning: OPA-DPO~\cite{yang2025opadpo} constructs on-policy corrections with GPT-4V; CHiP~\cite{fu2025chip} decomposes the DPO loss into three hierarchies; HA-DPO~\cite{zhao2023hadpo} detects and corrects hallucinations with GPT-4; LLaVA-RLHF~\cite{sun2023llava-rlhf} and RLHF-V~\cite{yu2023rlhfv} rely on human preferences; RLAIF-V~\cite{yu2024rlaifv} iterates with model feedback. \ours differs in three ways: (1) we target fine-grained negative \emph{input} queries, not only response-side errors~\cite{yang2025opadpo, zhao2023hadpo, yu2024rlaifv, yu2023rlhfv, sun2023llava-rlhf, sarkar2024halva}; (2) we post-train frontier MLLMs beyond the LLaVA family~\cite{yang2025opadpo, sarkar2024halva} and show strong performance against FINER; (3) we use standard DPO with a scalable data pipeline and a small LLM~\cite{abdin2024phi} for annotation, avoiding costly closed-source models and multi-iteration training~\cite{zhao2023hadpo, yang2025opadpo, liu2023lrv, sarkar2024halva, chen2025perturbollava, yu2024rlaifv}.

\section{Conclusion and Limitation}
\label{sec:conclusion}

\myparagraph{Conclusion.}
We introduced \textsc{FINER}, a suite of fine-grained negative queries that reveals how current MLLMs fail under precise negations. Systematic evaluation across all four settings of \textsc{FINER-CompreCap} and \textsc{FINER-DOCCI} shows that even frontier MLLMs remain vulnerable to FINER-induced hallucinations. To address this, we proposed \ours, a simple, model-agnostic recipe that aligns models to react correctly to fine-grained negative queries. Across diverse backbones and training regimes, \ours consistently reduces hallucinations and improves paired accuracy on FINER benchmarks, as well as a wide range of hallucination and general purpose benchmarks.
Despite these gains, high-granularity cases and Wh questions remain challenging. Future work will focus on stronger negation-aware reasoning, that comprehensively enhances MLLMs' capabilities. We envision FINER as a start to incentivize better benchmarks and methods.

\myparagraph{Limitations.}
Despite careful filtering, the large-scale benchmark is not fully curated by human; constructing a noise-free, fully human-validated FINER benchmark is left for future research. Our rule-based MCQ construction enables flexible entity combinations but may reduce question naturalness. Future work could refine phrasing with LLMs or human rewrites while ensuring correctness. In addition, our Multi-rel subsets contain at most three relations, which, with a suitable data source, could be extended to improve model capabilities and further challenge FINER.

\newpage
\myparagraph{Acknowledgments.}
This work was supported by the German Research Foundation (DFG): SFB 1233, Robust Vision: Inference Principles and Neural Mechanisms, TP A2, project number: 276693517. This work was partially
funded by the ERC (853489 - DEXIM), the German Federal Ministry of Education and Research (BMBF, grant number: 01IS18039A), and the Alfried Krupp von Bohlen und Halbach Foundation, which we thank for their generous support. This work is also supported by Hi! PARIS and ANR/France 2030 program (ANR-23-IACL-0005). This project was also supported by Google.org with a Google Cloud Platform (GCP) credit award. The authors gratefully acknowledge the scientific support and resources of the AI service infrastructure \textit{LRZ AI Systems} provided by the Leibniz Supercomputing Centre (LRZ) of the Bavarian Academy of Sciences and Humanities (BAdW), funded by Bayerisches Staatsministerium für Wissenschaft und Kunst (StMWK). In addition, the authors gratefully acknowledge the Gauss Centre for Supercomputing e.V. (\url{www.gauss-centre.eu}) for funding this project by providing computing time on the GCS Supercomputer JUWELS~\cite{JUWELS} at Jülich Supercomputing Centre (JSC).
{
    \small
    \bibliographystyle{ieeenat_fullname}
    \bibliography{main}

@String(CVPR= {IEEE Conf. Comput. Vis. Pattern Recog.})

@String(ICCV= {Int. Conf. Comput. Vis.})

@String(ECCV= {Eur. Conf. Comput. Vis.})

@String(ICLR = {Int. Conf. Learn. Represent.})

@String(CVPR  = {CVPR})

@String(ICCV  = {ICCV})

@String(ECCV  = {ECCV})

@String(ICLR  = {ICLR})

@inproceedings{li2023pope,
  title={Evaluating Object Hallucination in Large Vision-Language Models},
  author={Li, Yifan and Du, Yifan and Zhou, Kun and Wang, Jinpeng and Zhao, Xin and Wen, Ji-Rong},
  booktitle={EMNLP},
  year={2023}
}

@article{wang2023amber,
  title={Amber: An llm-free multi-dimensional benchmark for mllms hallucination evaluation},
  author={Wang, Junyang and Wang, Yuhang and Xu, Guohai and Zhang, Jing and Gu, Yukai and Jia, Haitao and Wang, Jiaqi and Xu, Haiyang and Yan, Ming and Zhang, Ji and others},
  journal={arXiv},
  year={2023}
}

@article{neuhaus2025repope,
    author = {RePOPE: Impact of Annotation Errors on the POPE Benchmark},
    title = {Neuhaus, Yannic and Hein, Matthias},
    journal = {arXiv},
    year = {2025}
}

@inproceedings{augustin2025dash,
    title={DASH: Detection and Assessment of Systematic Hallucinations of VLMs},
    author={Augustin, Maximilian and Neuhaus, Yannic and Hein, Matthias},
    booktitle={ICCV},
    year={2025}
}

@inproceedings{
    zhang2025mmmc,
    title={Robust Multimodal Large Language Models Against Modality Conflict},
    author={Zongmeng Zhang and Wengang Zhou and Jie Zhao and Houqiang Li},
    booktitle={ICML},
    year={2025},
}

@article{sun2023llava-rlhf,
  title={Aligning large multimodal models with factually augmented rlhf},
  author={Sun, Zhiqing and Shen, Sheng and Cao, Shengcao and Liu, Haotian and Li, Chunyuan and Shen, Yikang and Gan, Chuang and Gui, Liang-Yan and Wang, Yu-Xiong and Yang, Yiming and others},
  journal={arXiv},
  year={2023}
}

@inproceedings{yu2023rlhfv,
    author = {Yu, Tianyu and Yao, Yuan and Zhang, Haoye and He, Taiwen and Han, Yifeng and Cui, Ganqu and Hu, Jinyi and Liu, Zhiyuan and Zheng, Hai-Tao and Sun, Maosong and others},
    title = {Rlhf-v: Towards trustworthy mllms via behavior alignment from fine-grained correctional human feedback},
    booktitle = {CVPR},
    year = {2024}
}

@inproceedings{yu2024rlaifv,
  title={RLAIF-V: Aligning MLLMs through Open-Source AI Feedback for Super GPT-4V Trustworthiness}, 
  author={Yu, Tianyu and Zhang, Haoye and Yao, Yuan and Dang, Yunkai and Chen, Da and Lu, Xiaoman and Cui, Ganqu and He, Taiwen and Liu, Zhiyuan and Chua, Tat-Seng and Sun, Maosong},
  booktitle={CVPR},
  year={2025},
}

@inproceedings{liu2023lrv,
  title={Aligning Large Multi-Modal Model with Robust Instruction Tuning},
  author={Liu, Fuxiao and Lin, Kevin and Li, Linjie and Wang, Jianfeng and Yacoob, Yaser and Wang, Lijuan},
  booktitle={ICLR},
  year={2024}
}

@misc{liu2024llavanext,
    title={LLaVA-NeXT: Improved reasoning, OCR, and world knowledge},
    url={https://llava-vl.github.io/blog/2024-01-30-llava-next/},
    author={Liu, Haotian and Li, Chunyuan and Li, Yuheng and Li, Bo and Zhang, Yuanhan and Shen, Sheng and Lee, Yong Jae},
    month={January},
    year={2024}
}

@article{bai2025qwen25vl,
  title={Qwen2. 5-vl technical report},
  author={Bai, Shuai and Chen, Keqin and Liu, Xuejing and Wang, Jialin and Ge, Wenbin and Song, Sibo and Dang, Kai and Wang, Peng and Wang, Shijie and Tang, Jun and others},
  journal={arXiv},
  year={2025}
}

@article{wang2025internvl35,
  title={Internvl3. 5: Advancing open-source multimodal models in versatility, reasoning, and efficiency},
  author={Wang, Weiyun and Gao, Zhangwei and Gu, Lixin and Pu, Hengjun and Cui, Long and Wei, Xingguang and Liu, Zhaoyang and Jing, Linglin and Ye, Shenglong and Shao, Jie and others},
  journal={arXiv},
  year={2025}
}

@inproceedings{lu2025comprecap,
  title={Benchmarking large vision-language models via directed scene graph for comprehensive image captioning},
  author={Lu, Fan and Wu, Wei and Zheng, Kecheng and Ma, Shuailei and Gong, Biao and Liu, Jiawei and Zhai, Wei and Cao, Yang and Shen, Yujun and Zha, Zheng-Jun},
  booktitle={CVPR},
  year={2025}
}

@article{chen2024mmstar,
  title={Are we on the right way for evaluating large vision-language models?},
  author={Chen, Lin and Li, Jinsong and Dong, Xiaoyi and Zhang, Pan and Zang, Yuhang and Chen, Zehui and Duan, Haodong and Wang, Jiaqi and Qiao, Yu and Lin, Dahua and others},
  journal={NeurIPS},
  year={2024}
}

@inproceedings{singh2019textvqa,
    title={Towards VQA Models That Can Read},
    author={Singh, Amanpreet and Natarjan, Vivek and Shah, Meet and Jiang, Yu and Chen, Xinlei and Parikh, Devi and Rohrbach, Marcus},
    booktitle={CVPR},
    year={2019}
}

@inproceedings{masry2022chartqa,
  title={ChartQA: A Benchmark for Question Answering about Charts with Visual and Logical Reasoning},
  author={Masry, Ahmed and Do, Xuan Long and Tan, Jia Qing and Joty, Shafiq and Hoque, Enamul},
  booktitle={Findings of ACL},
  year={2022}
}

@inproceedings{tong2024eyes,
  title={Eyes wide shut? exploring the visual shortcomings of multimodal llms},
  author={Tong, Shengbang and Liu, Zhuang and Zhai, Yuexiang and Ma, Yi and LeCun, Yann and Xie, Saining},
  booktitle={CVPR},
  year={2024}
}

@inproceedings{li2024naturalbench,
  title={Naturalbench: Evaluating vision-language models on natural adversarial samples},
  author={Li, Baiqi and Lin, Zhiqiu and Peng, Wenxuan and Nyandwi, Jean de Dieu and Jiang, Daniel and Ma, Zixian and Khanuja, Simran and Krishna, Ranjay and Neubig, Graham and Ramanan, Deva},
  booktitle={NeurIPS},
  year={2024}
}

@inproceedings{wu2024vstar,
  title={V?: Guided visual search as a core mechanism in multimodal llms},
  author={Wu, Penghao and Xie, Saining},
  booktitle={CVPR},
  year={2024}
}

@inproceedings{guan2024hallusionbench,
  title={Hallusionbench: an advanced diagnostic suite for entangled language hallucination and visual illusion in large vision-language models},
  author={Guan, Tianrui and Liu, Fuxiao and Wu, Xiyang and Xian, Ruiqi and Li, Zongxia and Liu, Xiaoyu and Wang, Xijun and Chen, Lichang and Huang, Furong and Yacoob, Yaser and others},
  booktitle={CVPR},
  year={2024}
}

@inproceedings{wang2024allseeing_v2,
  title={The All-Seeing Project V2: Towards General Relation Comprehension of the Open World},
  author={Wang, Weiyun and Ren, Yiming and Luo, Haowen and Li, Tiantong and Yan, Chenxiang and Chen, Zhe and Wang, Wenhai and Li, Qingyun and Lu, Lewei and Zhu, Xizhou and others},
  booktitle={ECCV},
  year={2024}
}

@article{team2023gemini,
  title={Gemini: a family of highly capable multimodal models},
  author={Team, Gemini and Anil, Rohan and Borgeaud, Sebastian and Alayrac, Jean-Baptiste and Yu, Jiahui and Soricut, Radu and Schalkwyk, Johan and Dai, Andrew M and Hauth, Anja and Millican, Katie and others},
  journal={arXiv},
  year={2023}
}

@inproceedings{lin2014microsoft,
  title={Microsoft coco: Common objects in context},
  author={Lin, Tsung-Yi and Maire, Michael and Belongie, Serge and Hays, James and Perona, Pietro and Ramanan, Deva and Doll{\'a}r, Piotr and Zitnick, C Lawrence},
  booktitle={ECCV},
  year={2014},
}

@inproceedings{onoe2024docci,
  title={Docci: Descriptions of connected and contrasting images},
  author={Onoe, Yasumasa and Rane, Sunayana and Berger, Zachary and Bitton, Yonatan and Cho, Jaemin and Garg, Roopal and Ku, Alexander and Parekh, Zarana and Pont-Tuset, Jordi and Tanzer, Garrett and others},
  booktitle={ECCV},
  year={2024}
}

@inproceedings{deitke2025molmo,
  title={Molmo and pixmo: Open weights and open data for state-of-the-art vision-language models},
  author={Deitke, Matt and Clark, Christopher and Lee, Sangho and Tripathi, Rohun and Yang, Yue and Park, Jae Sung and Salehi, Mohammadreza and Muennighoff, Niklas and Lo, Kyle and Soldaini, Luca and others},
  booktitle={CVPR},
  year={2025}
}

@article{abdin2024phi,
  title={Phi-4 technical report},
  author={Abdin, Marah and Aneja, Jyoti and Behl, Harkirat and Bubeck, S{\'e}bastien and Eldan, Ronen and Gunasekar, Suriya and Harrison, Michael and Hewett, Russell J and Javaheripi, Mojan and Kauffmann, Piero and others},
  journal={arXiv},
  year={2024}
}

@inproceedings{zheng2024llamafactory,
  title={LlamaFactory: Unified Efficient Fine-Tuning of 100+ Language Models},
  author={Yaowei Zheng and Richong Zhang and Junhao Zhang and Yanhan Ye and Zheyan Luo and Zhangchi Feng and Yongqiang Ma},
  booktitle={ACL},
  year={2024}
}

@article{hu2022lora,
  title={Lora: Low-rank adaptation of large language models.},
  author={Hu, Edward J and Shen, Yelong and Wallis, Phillip and Allen-Zhu, Zeyuan and Li, Yuanzhi and Wang, Shean and Wang, Lu and Chen, Weizhu and others},
  journal={ICLR},
  year={2022}
}

@inproceedings{deng2025words,
  title={Words or Vision: Do Vision-Language Models Have Blind Faith in Text?},
  author={Deng, Ailin and Cao, Tri and Chen, Zhirui and Hooi, Bryan},
  booktitle={CVPR},
  year={2025}
}

@article{chen2025perturbollava,
  title={PerturboLLaVA: Reducing multimodal hallucinations with perturbative visual training},
  author={Chen, Cong and Liu, Mingyu and Jing, Chenchen and Zhou, Yizhou and Rao, Fengyun and Chen, Hao and Zhang, Bo and Shen, Chunhua},
  journal={ICLR},
  year={2025}
}

@InProceedings{yang2025opadpo,
    author    = {Yang, Zhihe and Luo, Xufang and Han, Dongqi and Xu, Yunjian and Li, Dongsheng},
    title     = {Mitigating Hallucinations in Large Vision-Language Models via DPO: On-Policy Data Hold the Key},
    booktitle = {CVPR},
    year      = {2025},
}

@inproceedings{fu2025chip,
  title={Chip: Cross-modal hierarchical direct preference optimization for multimodal llms},
  author={Fu, Jinlan and Huangfu, Shenzhen and Fei, Hao and Shen, Xiaoyu and Hooi, Bryan and Qiu, Xipeng and Ng, See-Kiong},
  booktitle={ICLR},
  year={2025}
}

@misc{omnilmm,
  title = {Large multi-modal models for strong performance and efficient deployment},
author={OpenBMB},
  year         = {2024},
}

@inproceedings{sarkar2024halva,
  title={Data-augmented phrase-level alignment for mitigating object hallucination},
  author={Sarkar, Pritam and Ebrahimi, Sayna and Etemad, Ali and Beirami, Ahmad and Ar{\i}k, Sercan {\"O} and Pfister, Tomas},
  booktitle={ICLR},
  year={2025}
}

@misc{zhao2023hadpo,
      title={Beyond Hallucinations: Enhancing LVLMs through Hallucination-Aware Direct Preference Optimization}, 
      author={Zhiyuan Zhao and Bin Wang and Linke Ouyang and Xiaoyi Dong and Jiaqi Wang and Conghui He},
      year={2023},
      eprint={2311.16839},
      archivePrefix={arXiv},
      primaryClass={cs.CV}
}

@article{rohrbach2018chair,
  title={Object hallucination in image captioning},
  author={Rohrbach, Anna and Hendricks, Lisa Anne and Burns, Kaylee and Darrell, Trevor and Saenko, Kate},
  journal={arXiv},
  year={2018}
}

@misc{relaion,
  author = {LAION},
  title = {Releasing Re-LAION-5B: transparent iteration on LAION-5B with additional safety fixes},
  note = {Accessed: 30 aug, 2024},
  year = {2024}
}

@inproceedings{zhu2023minigpt,
  title={MiniGPT-4: Enhancing Vision-Language Understanding with Advanced Large Language Models},
  author={Zhu, Deyao and Chen, Jun and Shen, Xiaoqian and Li, Xiang and Elhoseiny, Mohamed},
  booktitle={ICLR},
  year = {2023}}

@article{ye2023mplug,
  title={mplug-owl: Modularization empowers large language models with multimodality},
  author={Ye, Qinghao and Xu, Haiyang and Xu, Guohai and Ye, Jiabo and Yan, Ming and Zhou, Yiyang and Wang, Junyang and Hu, Anwen and Shi, Pengcheng and Shi, Yaya and others},
  journal={arXiv},
  year={2023}
}

@article{achiam2023gpt,
  title={Gpt-4 technical report},
  author={Achiam, Josh and Adler, Steven and Agarwal, Sandhini and Ahmad, Lama and Akkaya, Ilge and Aleman, Florencia Leoni and Almeida, Diogo and Altenschmidt, Janko and Altman, Sam and Anadkat, Shyamal and others},
  journal={arXiv},
  year={2023}
}

@inproceedings{wang2024haloquest,
  title={Haloquest: A visual hallucination dataset for advancing multimodal reasoning},
  author={Wang, Zhecan and Bingham, Garrett and Yu, Adams Wei and Le, Quoc V and Luong, Thang and Ghiasi, Golnaz},
  booktitle={ECCV},
  year={2024}
}

@inproceedings{lovenia2024nope,
  title={Negative Object Presence Evaluation (NOPE) to Measure Object Hallucination in Vision-Language Models},
  author={Lovenia, Holy and Dai, Wenliang and Cahyawijaya, Samuel and Ji, Ziwei and Fung, Pascale},
  booktitle={Proceedings of the 3rd Workshop on ALVR},
  year={2024}
}

@inproceedings{wu2025reverse,
  title={Generate, but Verify: Reducing Hallucination in Vision-Language Models with Retrospective Resampling},
  author={Wu, Tsung-Han and Lee, Heekyung and Ge, Jiaxin and Gonzalez, Joseph E and Darrell, Trevor and Chan, David M},
  booktitle={NeurIPS},
  year={2025}
}

@misc{liu2023llava,
      title={Visual Instruction Tuning}, 
      author={Liu, Haotian and Li, Chunyuan and Wu, Qingyang and Lee, Yong Jae},
      publisher={NeurIPS},
      year={2023},
}

@inproceedings{chuang2023dola,
  title={DoLa: Decoding by Contrasting Layers Improves Factuality in Large Language Models},
  author={Chuang, Yung-Sung and Xie, Yujia and Luo, Hongyin and Kim, Yoon and Glass, James R and He, Pengcheng},
  booktitle={The Twelfth International Conference on Learning Representations},
  year={2023}
}

@article{yang2025qwen3,
  title={Qwen3 technical report},
  author={Yang, An and Li, Anfeng and Yang, Baosong and Zhang, Beichen and Hui, Binyuan and Zheng, Bo and Yu, Bowen and Gao, Chang and Huang, Chengen and Lv, Chenxu and others},
  journal={arXiv},
  year={2025}
}

@article{comanici2025gemini25,
  title={Gemini 2.5: Pushing the frontier with advanced reasoning, multimodality, long context, and next generation agentic capabilities},
  author={Comanici, Gheorghe and Bieber, Eric and Schaekermann, Mike and Pasupat, Ice and Sachdeva, Noveen and Dhillon, Inderjit and Blistein, Marcel and Ram, Ori and Zhang, Dan and Rosen, Evan and others},
  journal={arXiv},
  year={2025}
}

@article{rafailov2023direct,
  title={Direct preference optimization: Your language model is secretly a reward model},
  author={Rafailov, Rafael and Sharma, Archit and Mitchell, Eric and Manning, Christopher D and Ermon, Stefano and Finn, Chelsea},
  journal={NeurIPS},
  year={2023}
}

@inproceedings{duan2024vlmevalkit,
  title={Vlmevalkit: An open-source toolkit for evaluating large multi-modality models},
  author={Duan, Haodong and Yang, Junming and Qiao, Yuxuan and Fang, Xinyu and Chen, Lin and Liu, Yuan and Dong, Xiaoyi and Zang, Yuhang and Zhang, Pan and Wang, Jiaqi and others},
  booktitle={ACM MM},
  year={2024}
}

@article{bai2024hallucination,
  title={Hallucination of multimodal large language models: A survey},
  author={Bai, Zechen and Wang, Pichao and Xiao, Tianjun and He, Tong and Han, Zongbo and Zhang, Zheng and Shou, Mike Zheng},
  journal={arXiv},
  year={2024}
}

@inproceedings{chen2024multiobject,
  title={Multi-Object Hallucination in Vision Language Models},
  author={Chen, Xuweiyi and Ma, Ziqiao and Zhang, Xuejun and Xu, Sihan and Qian, Shengyi and Yang, Jianing and Fouhey, David and Chai, Joyce},
  booktitle={NeurIPS},
  year={2024}
}

@inproceedings{zhou2017ade20k,
        title={Scene Parsing through ADE20K Dataset},
        author={Zhou, Bolei and Zhao, Hang and Puig, Xavier and Fidler, Sanja and Barriuso, Adela and Torralba, Antonio},
        booktitle={CVPR},
        year={2017}
}

@inproceedings{loshchilovadamw,
  title={Decoupled Weight Decay Regularization},
  author={Loshchilov, Ilya and Hutter, Frank},
  booktitle={ICLR},
  year={2019}
}

@inproceedings{tu2025ode,
  title={ODE: Open-Set Evaluation of Hallucinations in Multimodal Large Language Models},
  author={Tu, Yahan and Hu, Rui and Sang, Jitao},
  booktitle={CVPR},
  year={2025}
}

@inproceedings{ding2024hallupi,
  title={Hallu-pi: Evaluating hallucination in multi-modal large language models within perturbed inputs},
  author={Ding, Peng and Wu, Jingyu and Kuang, Jun and Ma, Dan and Cao, Xuezhi and Cai, Xunliang and Chen, Shi and Chen, Jiajun and Huang, Shujian},
  booktitle={ACM MM},
  year={2024}
}

@inproceedings{liu2024improved,
  title={Improved baselines with visual instruction tuning},
  author={Liu, Haotian and Li, Chunyuan and Li, Yuheng and Lee, Yong Jae},
  booktitle={CVPR},
  year={2024}
}

@article{zhou2024povid,
  title={Aligning modalities in vision large language models via preference fine-tuning},
  author={Zhou, Yiyang and Cui, Chenhang and Rafailov, Rafael and Finn, Chelsea and Yao, Huaxiu},
  journal={arXiv},
  year={2024}
}

@article{JUWELS,
author = {{J\"{u}lich Supercomputing Centre}},
title = {{JUWELS Cluster and Booster: Exascale Pathfinder with Modular Supercomputing Architecture at Juelich Supercomputing Centre}},
journal = {Journal of large-scale research facilities},
year = {2021}
}

@inproceedings{xiao2025flair,
  title={FLAIR: VLM with Fine-grained Language-informed Image Representations},
  author={Xiao, Rui and Kim, Sanghwan and Georgescu, Mariana-Iuliana and Akata, Zeynep and Alaniz, Stephan},
  booktitle={CVPR},
  year={2025}
}

@inproceedings{liu2025unveiling,
  title={Unveiling the ignorance of mllms: Seeing clearly, answering incorrectly},
  author={Liu, Yexin and Liang, Zhengyang and Wang, Yueze and Wu, Xianfeng and Tang, Feilong and He, Muyang and Li, Jian and Liu, Zheng and Yang, Harry and Lim, Sernam and others},
  booktitle={Proceedings of the Computer Vision and Pattern Recognition Conference},
  year={2025}
}

@inproceedings{kim2025cosmos,
  title={Cosmos: Cross-modality self-distillation for vision language pre-training},
  author={Kim, Sanghwan and Xiao, Rui and Georgescu, Mariana-Iuliana and Alaniz, Stephan and Akata, Zeynep},
  booktitle={Proceedings of the IEEE/CVF Conference on Computer Vision and Pattern Recognition},
  pages={14690--14700},
  year={2025}
}
}

\clearpage
\setcounter{page}{1}
\setcounter{section}{0}
\renewcommand\thesection{\Alph{section}}
\maketitlesupplementary

\section{Extended Related Works}
\label{sec:supp_extended_related_works}
\subsection{Hallucination benchmarks}
CHAIR~\cite{rohrbach2018chair} benchmarks object hallucination in image captioning by measuring how many generated words actually appear in the image, based on ground-truth captions and object segmentations. However, the CHAIR metric suffers from instability issues~\cite{li2023pope}. POPE~\cite{li2023pope} simplifies hallucination detection by asking models yes-or-no questions. RePOPE~\cite{neuhaus2025repope} identifies annotation errors in POPE and provides a revised version. Amber~\cite{wang2023amber} evaluates hallucinations in both generative and discriminative settings. In the discriminative setting, it categorizes hallucinations into “object,” “relation,” and “attribute” types. A common limitation of these benchmarks is their reliance on the MSCOCO dataset~\cite{lin2014microsoft}. To better detect object hallucinations at scale, DASH~\cite{augustin2025dash} adopts a retrieval-based approach to select images from LAION-5B~\cite{relaion}. CRPE~\cite{wang2024allseeing_v2} focuses on relation-based hallucinations but limits its evaluation to single-relation cases. 

Beyond hallucination detection, MMMC~\cite{zhang2025mmmc} introduces the concept of “modality conflicts,” referring to mismatches between the image and the text query, an approach we consider coarse-grained negative querying. FLAIR~\cite{xiao2025flair} constructs DOCCI-FG that also adopts DOCCI captions to test how well vision-language models understand images from a fine-grained perspective. COSMOS~\cite{kim2025cosmos} evaluates and further improves fine-grained vision-language alignment via a self-distillation approach. The “Blind-faith-in-Text” phenomenon~\cite{deng2025words} shows that when a conflicting textual context is prefixed to a query, models tend to trust the text more than the image. Similarly, Hallu-PI~\cite{ding2024hallupi} evaluates hallucinations by appending additional images or texts as a perturbation. In our work, we do not add extra textual context. Instead, we design user queries that contain subtle and nuanced conflicts with the image, allowing us to study hallucination behavior without altering the conversational setup. MMVU~\cite{liu2025unveiling} also proposes a benchmark that investigates “negative questions.” The key difference is that our work studies this problem at a finer level of granularity.

HaloQuest~\cite{wang2024haloquest} includes a “false premise” subset with a similar motivation to our Wh setting. However, our setting differs because our false premises lie in the fine-grained attributes of existing objects, while HaloQuest mainly focuses on non-existent objects. Likewise, NOPE~\cite{lovenia2024nope} mainly evaluates hallucinations involving non-existent objects but does not test hallucinations related to attributes or relations. ROPE~\cite{chen2024multiobject} evaluates object hallucinations by prompting MLLM to pick the correct objects corresponding multiple input visual prompts. While this approach shares similarity with our Multi-obj subset, we aim for more flexibility by directly inserting the negative object at random position in the prompt and we do not rely on bounding boxes annotation from MSCOCO-Panoptic~\cite{lin2014microsoft} or ADE20K~\cite{zhou2017ade20k}. ODE~\cite{tu2025ode} introduces an open-set dynamic hallucination evaluation to prevent data contamination. This also aligns with our intuition to adopt DOCCI~\cite{onoe2024docci} as an additional data source and create the less-saturated \textsc{FINER-DOCCI}.

\subsection{Hallucination-aware Fine-tuning}
To reduce hallucinations, various fine-tuning techniques have been developed for MLLMs. Closely related to our motivation, LRV-Instruct~\cite{liu2023lrv} applies supervised fine-tuning (SFT) to MiniGPT-4~\cite{zhu2023minigpt} and mPLUG-Owl~\cite{ye2023mplug}, and introduces negative instructions by manipulating objects and factual knowledge using GPT-4~\cite{achiam2023gpt}. HALVA~\cite{sarkar2024halva} leverages Gemini Vision Pro~\cite{team2023gemini} to construct both correct and hallucinated responses, and applies a contrastive loss between them, explicitly pushing the model away from hallucinated generations.

PerturboLLaVA~\cite{chen2025perturbollava} appends misleading textual context as perturbations generated by GPT-4o~\cite{achiam2023gpt} and trains the model via instruction tuning to remain robust under such distracting inputs. REVERSE~\cite{wu2025reverse} expands the model’s vocabulary with special uncertainty tokens and builds a large-scale instruction-following dataset; the model learns to perform retrospective reasoning whenever these tokens are triggered, allowing it to revise potentially hallucinated content. RLHF-V~\cite{yu2023rlhfv} and LLaVA-RLHF~\cite{sun2023llava-rlhf} apply reinforcement learning from human feedback (RLHF) to vision-language models, using human preference signals to improve response quality and reduce hallucinations. RLAIF-V~\cite{yu2024rlaifv} instead leverages AI feedback (RLAIF): a stronger teacher model provides automatic preference judgments, and the student model is updated in a self-evolving manner over multiple training rounds.

Several studies employ Direct Preference Optimization (DPO) to reduce hallucinations. OPA-DPO~\cite{yang2025opadpo} constructs on-policy data for hallucination mitigation and uses GPT-4V for fine-grained hallucination correction in the training set. CHiP~\cite{fu2025chip} decomposes the DPO objective into response-level, segment-level, and token-level components to better localize hallucinations. HA-DPO~\cite{zhao2023hadpo} also uses GPT-4~\cite{achiam2023gpt} to identify and correct hallucinations in model outputs. POVID~\cite{zhou2024povid} adopts GPT-4V to inject hallucinated objects, attributes, and relations directly into the dispreferred responses, encouraging the model to reject these patterns during training.

In light of these works, our approach differs in three main aspects. First, most prior studies~\cite{yang2025opadpo, zhao2023hadpo, yu2024rlaifv, yu2023rlhfv, sun2023llava-rlhf, sarkar2024halva, zhou2024povid} focus on detecting and correcting hallucinations in model responses, whereas we explicitly construct fine-grained negative \emph{input queries} at the object, attribute, and relation level. Second, previous efforts~\cite{yang2025opadpo, sarkar2024halva} primarily target the LLaVA family, while we directly post-train several state-of-the-art MLLMs and evaluate them on the FINER benchmarks, improving model's robustness against nuanced errors in queries. Third, \ours follows the standard DPO algorithm and does not require multi-iteration training as in RLAIF-V. Unlike prior works~\cite{zhao2023hadpo, yang2025opadpo, liu2023lrv, sarkar2024halva, chen2025perturbollava, zhou2024povid} that rely heavily on costly closed-source models to build training data, we propose a scalable pipeline that uses an open-source LLM~\cite{abdin2024phi} to generate high-quality preference pairs from existing long-caption datasets.

\begin{figure*}[t]
  \centering
  \includegraphics[width=\textwidth]{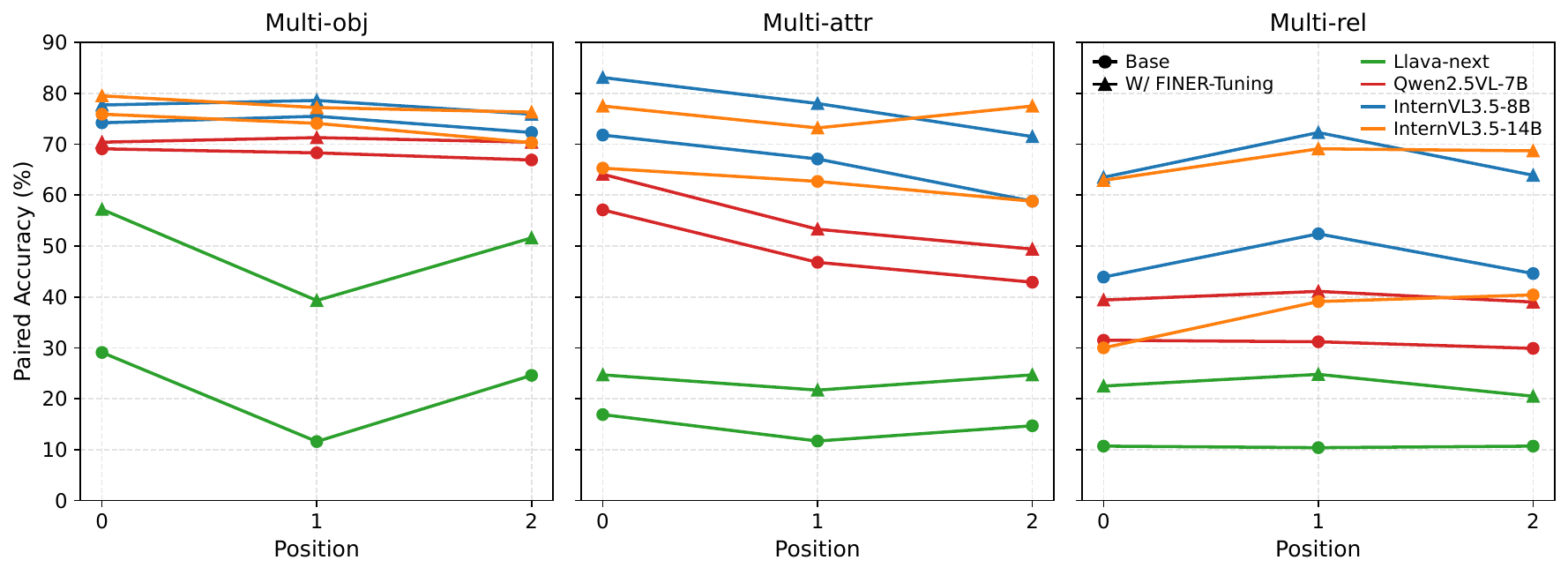}
  \caption{
  Positional bias analysis on \textsc{FINER-CompreCap}. We select all $q^\pm_{\text{multi-obj}}$, $q^\pm_{\text{multi-attr}}$, and $q^\pm_{\text{multi-rel}}$ that contain three entities. Since each $q^-$ always has exactly one negated entity, we cyclically move that negated entity to each of the three positions (and move the corresponding positive entity accordingly), and compute the averaged paired accuracy $\text{Acc}_{\text{paired}}$ for each position.
}

  \label{fig:supp_pos_bias}
\end{figure*}

\section{FINER Benchmark Details}
In this section, we describe the construction of \textsc{FINER-CompreCap} and \textsc{FINER-DOCCI}. \textsc{FINER-CompreCap} starts from human-annotated positive scene-graphs (SGs) with minor edits (Sec.~\ref{sec:supp_pos_sg_comprecap}). \textsc{FINER-DOCCI} derives positive SGs from dense captions (Sec.~\ref{sec:supp_pos_sg_docci}). We then apply the same negative-generation and filtering pipeline to obtain negative SGs (Sec.~\ref{sec:supp_negatives_generation_pipeline}). Finally, both positive and negative SGs are converted into benchmark questions via our rule-based MCQ pipeline (Sec.~\ref{sec:supp_mcq_design}).

The two benchmarks are motivated slightly differently. \textsc{FINER-CompreCap} builds on human-annotated SGs, supporting more precise evaluation. In contrast, \textsc{FINER-DOCCI} explores whether dense captions can be used to synthesize SGs beyond COCO object classes and images, enabling open-set evaluation~\cite{tu2025ode} at substantially larger scale. As a result, \textsc{FINER-DOCCI} is primarily designed to validate our findings at scale, rather than to maximize per-sample annotation fidelity.
\label{sec:supp_gemini_extract_example}
\label{sec:supp_cap2sg_pipeline}

\subsection{Positive SG for \textsc{FINER-CompreCap}}
\label{sec:supp_pos_sg_comprecap}

\begin{figure*}[t]
  \centering
  \includegraphics[width=0.85\textwidth]{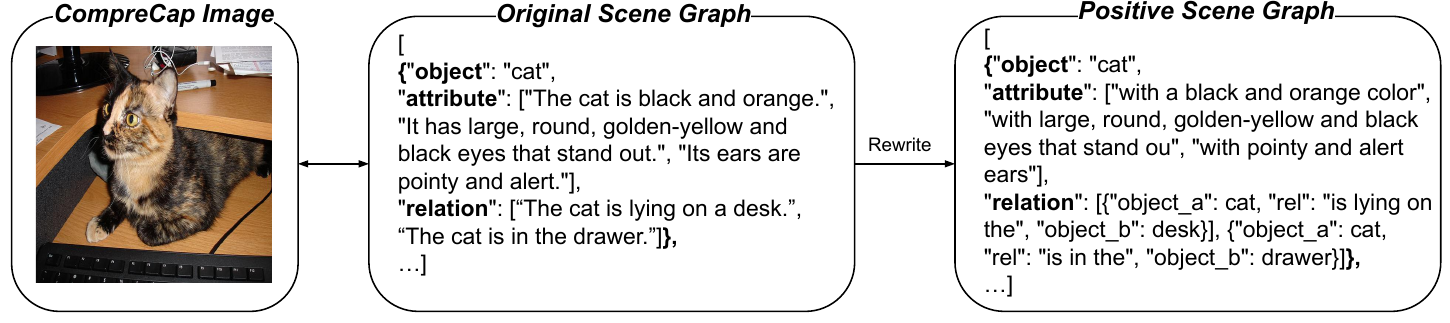}
  \caption{Example of positive scene graph (SG) in \textsc{FINER-CompreCap}. CompreCap~\cite{lu2025comprecap} already pairs each image with SG-like annotation. We further adopts Qwen3-14B~\cite{yang2025qwen3} to simply rewrite attribute sentences into phrases.}
  \label{fig:supp_pos_sg_comprecap}
\end{figure*}

CompreCap~\cite{lu2025comprecap} offers 560 human-annotated images, each with a scene-graph (SG) annotation. Each SG annotation already consists of objects, attributes, and relations. The attribute annotations in the original SG are lists of simple sentences, which we rewrite with Qwen3-14B~\cite{yang2025qwen3} into ``with \{attr\}'' phrases without changing their original meaning. The original relation annotations are also sentences describing a relation between a subject and an object. Therefore, we use a rule-based method to parse the relation sentences into dictionary-like annotations. These steps are necessary because we need to combine objects, relations, and attributes in our MCQ construction. We manually inspect the positive annotations to ensure their integrity. Since our preprocessing only changes sentence structure and does not introduce new annotations, it is robust. We provide an example SG in Fig.~\ref{fig:supp_pos_sg_comprecap}. As shown in Fig.~\ref{fig:supp_pos_sg_comprecap}, the original attribute ``The cat is black and orange'' is rewritten as ``with a black and orange color''. Meanwhile, the original relation ``The cat is lying on a desk'' is parsed into a dictionary-like structure.

\subsection{SG Extraction Pipeline for \textsc{FINER-DOCCI}}
\label{sec:supp_pos_sg_docci}

\begin{figure*}[t]
  \centering
  \includegraphics[width=\textwidth]{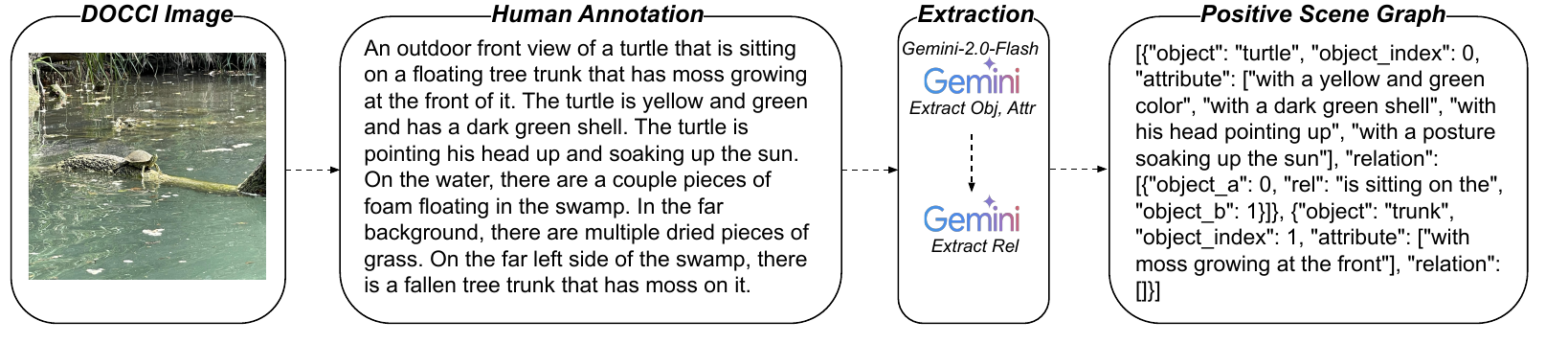}
  \caption{Example positive scene graph (SG) extracted by Gemini-2.0-Flash~\cite{team2023gemini}. Given a long human-annotated caption from DOCCI~\cite{onoe2024docci}, we apply a two-stage extraction pipeline to obtain the positive SG.}
  \label{fig:supp_gemini_extraction_example}
\end{figure*}

DOCCI~\cite{onoe2024docci} consists of 5{,}000 images, each paired with a detailed human-annotated caption. Such rich descriptions already contain the necessary information about objects, attributes, and relations. Fig.~\ref*{fig:supp_gemini_extraction_example} shows an example caption together with the positive scene graph extracted by Gemini-2.0-Flash~\cite{comanici2025gemini25}.

Directly prompting an LLM to ``summarize'' a full scene graph is known to be brittle and prone to errors. Instead, inspired by PerturboLLaVA~\cite{chen2025perturbollava}, which prompts an LLM to extract objects, attributes, and relations from long captions, we design a conservative two-stage extraction pipeline that decomposes the task into simpler subproblems and incorporates explicit cross-checks and human validation.

\myparagraph{Stage 1: object and attribute extraction.}
In the first stage, we only ask Gemini-2.0-Flash to extract objects and their attributes from the caption. The model is instructed to copy phrases \emph{verbatim} from the caption and to avoid inventing new entities or attributes. This turns the problem into a pure information extraction task rather than open-ended generation. The prompt is visualized in Fig.~\ref{fig:supp_prompt_gemini_obj_attr}. Human annotators inspect randomly sampled outputs to check the robustness of this stage, as the model only needs to detect and group textual mentions instead of inferring unseen content.

\myparagraph{Stage 2: relation extraction and validation.}
In the second stage, we consider pairs of extracted objects and ask Gemini-2.0-Flash whether the caption explicitly states a relation between them. Given the full caption and a candidate object pair, Gemini is instructed to either (i) return the exact relation phrase from the caption, or (ii) not return anything if no relation is explicitly mentioned. The model is explicitly told not to infer or imagine relations that are not written in the caption. This again restricts Gemini to acting as an information extractor, which increases reliability. The prompt is displayed in Fig.\ref{fig:supp_prompt_gemini_rel}.

Even with these restrictions, some errors in the extracted relations remain. To further filter noisy relations, we perform a joint visual-textual validation step. For each candidate relation, we:
\begin{itemize}
  \item run a binary classifier with Qwen2.5-VL-72B~\cite{bai2025qwen25vl} to decide whether the relation holds in the image; and
  \item query Gemini again, this time asking whether the relation is \emph{explicitly} supported by the caption.
\end{itemize}
If both models disagree with the proposed relation, we discard it. Among the misclassified relations, we further ask human annotators to verify a subset of 400 samples and, whenever they spot errors, remove incorrect extracted relation annotations. In total, this joint process of Qwen2.5-VL, Gemini, and humans filters out 1{,}771 relations.

Overall, this pipeline is deliberately conservative: we only keep relations that are supported by the caption (via extraction) and by the image (via a strong MLLM), with additional human checks on top. This design prioritizes precision over recall and makes our extracted SG for \textsc{FINER-DOCCI} more reliable despite the known challenges of using LLMs for scene-graph extraction.

\myparagraph{Quality Assessment.}
To assess the quality of the extracted objects, attributes, and relations in the positive SG of \textsc{FINER-DOCCI}, we run InternVL3.5-8B~\cite{wang2025internvl35} as a binary classifier. For each extracted object, attribute, or relation, the model is asked to answer ``Yes'' or ``No'' regarding its presence in the image. As a baseline, we apply the same procedure to the positive SG of \textsc{FINER-CompreCap}, whose scene graphs are human-annotated. The results are reported in Tab.~\ref{tab:positive_qualitative_assessments}. InternVL3.5-8B achieves comparable performance (96.4\% vs.\ 96.1\%) when classifying ground-truth objects in both benchmarks. For attributes, its accuracy on \textsc{FINER-DOCCI} is 3.2\% lower than on \textsc{FINER-CompreCap}. Given that the SG in \textsc{FINER-DOCCI} is much larger in scale than in \textsc{FINER-CompreCap} (see Tab.~\ref{tab:supp_neg_generation_stats}), this gap is acceptable. Notably, the accuracy on relations in the positive SG of \textsc{FINER-DOCCI} is slightly higher than that of \textsc{FINER-CompreCap} (85.1\% vs.\ 82.8\%). This likely reflects that the relation annotations in \textsc{FINER-DOCCI} are more detailed, providing the MLLM with more information to verify their correctness, rather than indicating that the human-annotated relations in \textsc{FINER-CompreCap} are of lower quality.

\begin{table}[t]
  \centering
  \caption{Quality assessment of the extracted positive objects, attributes, and relations for \textsc{FINER-DOCCI} using InternVL3.5-8B~\cite{wang2025internvl35} as a binary classifier. As a baseline, we also run InternVL3.5-8B as a binary classifier to classify the human annotations from \textsc{FINER-CompreCap}.}
  \label{tab:positive_qualitative_assessments}
  \setlength{\tabcolsep}{6pt}
  \renewcommand{\arraystretch}{1.05}
  \begin{tabular}{@{}lcccccc@{}}
    \toprule
    & \multicolumn{3}{c}{\textsc{FINER-CompreCap}} 
    & \multicolumn{3}{c}{\textsc{FINER-DOCCI}} \\
    \cmidrule(lr){2-4} \cmidrule(l){5-7}
    & Obj & Attr & Rel & Obj & Attr & Rel \\
    \midrule
    \dporow Acc. (\%) 
      & 96.4 & 91.5 &  82.8
      & 96.1 & 88.3 & 85.1 \\
    \bottomrule
  \end{tabular}
\end{table}

\subsection{Negatives Generation Pipeline.}
\label{sec:supp_negatives_generation_pipeline}

Having obtained the positive scene graphs (SGs) for both \textsc{FINER-CompreCap} and \textsc{FINER-DOCCI}, we construct a pipeline for generating negatives. For each object (\obj), attribute (\attr), and relation (\rel), we generate four negative counterparts, denoted as \negobj, \negattr, and \negrel.

\myparagraph{LLM-based negatives proposal.}
We first use an LLM as a ``negatives generator''. For \textsc{FINER-DOCCI} we use Gemini-2.0-Flash~\cite{team2023gemini}, and for \textsc{FINER-CompreCap} we use Qwen3-14B~\cite{yang2025qwen3}. Given a positive phrase (\obj, \attr, or \rel), the LLM is prompted to produce four negative phrases that have the opposite or a clearly different meaning from the positive. This step is efficient and does not directly inherit visual biases from any vision model, since it operates purely in text space.

A limitation of this step is that some generated negatives may in fact describe entities that are present in the image. Such ``false negatives'' are harmful for evaluation. Given the scale of the two positive SGs, pure human validation on the whole set is unfortunately not possible, so we need an automatic way to detect and filter these false negatives.

\myparagraph{MLLM-based discrimination and entropy.}
To filter these cases, we use Qwen2.5-VL-72B~\cite{bai2025qwen25vl} as a visual discriminator. For each positive phrase $x$ (where $x$ can be either \obj, \attr, or \rel) and its four candidate negatives $\{x_j^{-}\}_{j=1}^{4}$, we form a five-choice multiple-choice question with the candidate set
\[
\mathcal{C}(x) = \{x, x_1^{-}, x_2^{-}, x_3^{-}, x_4^{-}\}.
\]
We query Qwen2.5-VL-72B with the image and the set $\mathcal{C}(x)$, and obtain a probability distribution
\[
p = (p_1, \dots, p_5), \qquad \sum_{i=1}^{5} p_i = 1,
\]
over the five choices. We treat the original positive $x$ as the correct label. If the model selects $x$, the classification is correct; otherwise it is misclassified.

We compute the entropy of the model output
\begin{equation}
  H(p) = - \sum_{i=1}^{5} p_i \log p_i,
  \label{eq:supp_entropy}
\end{equation}
where the logarithm is natural. Low entropy means that the model is very confident in one of the options, while high entropy indicates uncertainty. If Qwen2.5-VL-72B makes a misclassification by choosing one negative while maintaining very low entropy, this indicates high confidence in its prediction. This likely reflects that the chosen entity somehow exists in the image (or, of course, the model can also be too confident about an actually wrong prediction).

\begin{figure*}[t]
  \centering
  \includegraphics[width=\textwidth]{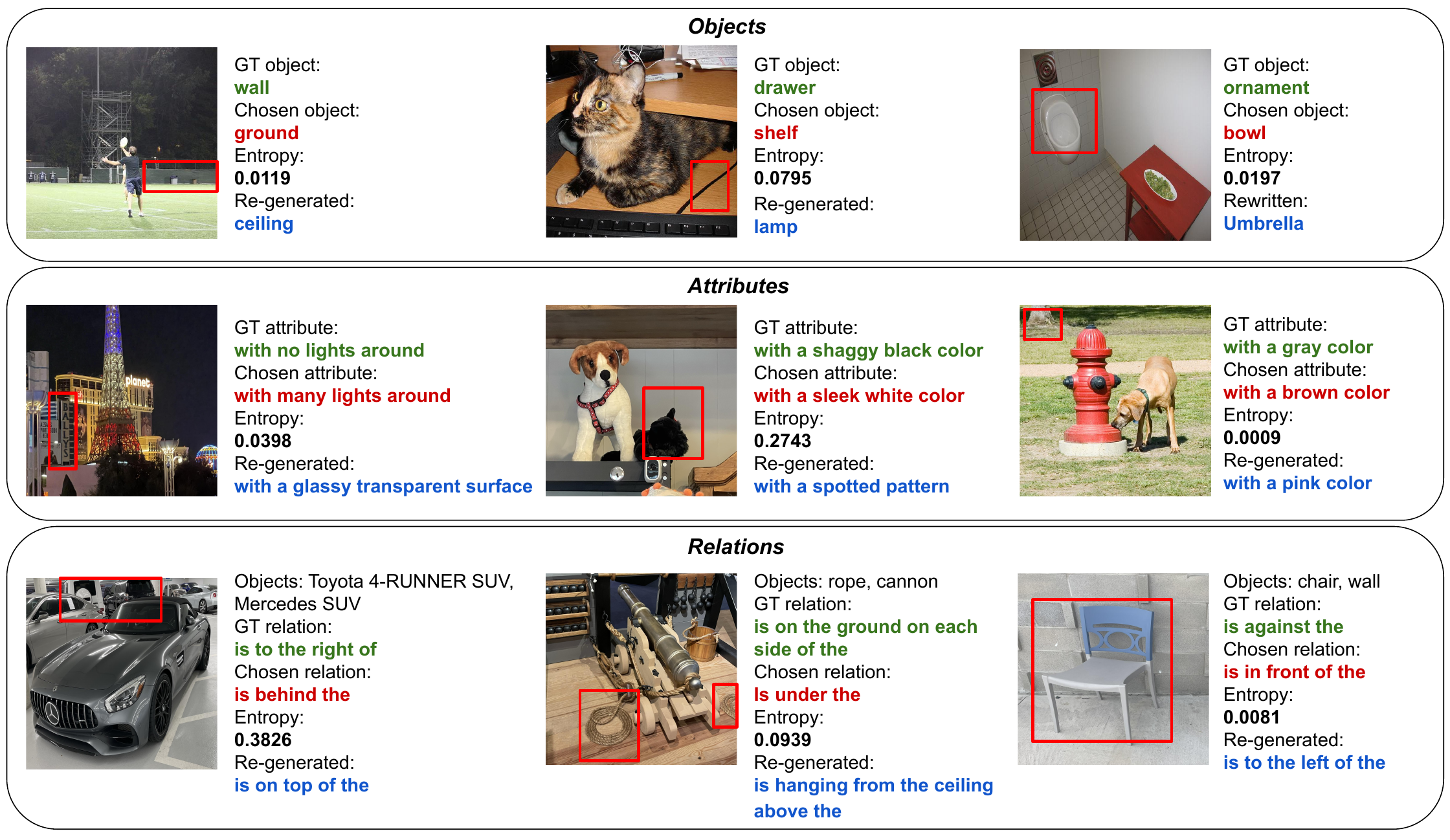}
  \caption{Examples of entropy-based filtering
for objects, attributes, and relations. The corresponding objects are shown with red bounding boxes. The ground-truth object/attribute/relation is highlighted in green. We prompt Qwen2.5-VL-72B~\cite{bai2025qwen25vl} to select the positive among four negatives. Green text indicates that the model makes an incorrect prediction and chooses a negative with low entropy scores. Blue text shows new negative candidates generated by the LLM. The examples are from both \textsc{FINER-CompreCap} and \textsc{FINER-DOCCI}.}

  \label{fig:supp_entropy_filtering}
\end{figure*}

We show several examples in Fig.~\ref{fig:supp_entropy_filtering}. Empirically, we observe that many bad negatives that actually appear in the image lead to misclassifications with very low entropy. For example, in one sample, ``ground'' is proposed as a negative for the object ``wall''. Since the ground region is clearly visible in the image, Qwen2.5-VL-72B strongly prefers the option ``ground'', with an entropy of $H(p) = 0.0119$. This indicates that the model is highly confident that ``ground'' is present in the image, and therefore this negative should be rejected. In such cases, we prompt the LLM again and rewrite the negative, for example from ``ground'' to ``ceiling'', which does not appear in the image.

However, low entropy does not always mean that the negative actually appears in the image; the MLLM can also be confidently wrong. For instance, in the car example in Fig.~\ref{fig:supp_entropy_filtering}, Qwen2.5-VL-72B misclassifies the relation phrase ``is behind the'' with low entropy $H(p) = 0.0119$, even though ``is behind the'' is a valid negative. In this case, we still replace it with a new negative proposal such as ``is on top of the'', which remains valid. Since our primary goal is to remove negatives that truly appear in the image, occasionally regenerating valid negatives is acceptable.

\myparagraph{Entropy-based filtering with human verification.}
We denote the entropy filtering threshold as $\theta$. For each benchmark and each level (object, attribute, relation), we choose a separate threshold $\theta$.

To set these thresholds, we first run Qwen2.5-VL-72B on the entire dataset and record, for each example, the model prediction and the corresponding entropy $H(p)$. We then collect all misclassified examples and sort them in ascending order of entropy. Starting from the lowest-entropy region, a human annotator verifies 10 misclassified examples and labels whether the proposed negative actually appears in the image. We then incrementally increase the candidate entropy threshold and, at each step, again sample 10 misclassified examples around the current threshold for human verification. We repeat this process until no ``bad negatives'' (negatives that truly appear in the image) are found among the 10 inspected samples; we then take the current entropy value as the threshold $\theta$ such that misclassified examples with $H(p) < \theta$ are likely to be true false negatives (the negative phrase is in the image), while those with higher entropy are retained as hard but valid negatives.

During the full pipeline, each negative candidate that leads to a misclassification with $H(p) < \theta$ is sent back to the LLM and regenerated. The new proposal is checked again by Qwen2.5-VL-72B with the same procedure. After each round of regeneration and classification, we subsample a small set of misclassified examples and ask a human annotator to inspect the remaining negatives. This human-in-the-loop process is to reduce the risk of systematic errors introduced by the automatic filtering pipeline.

We summarize the thresholds $\theta$, the total number of samples, and the number of regenerated negatives for each benchmark and each level (Obj, Attr, Rel) in Tab.~\ref{tab:supp_neg_generation_stats}.

\begin{table}[t]
  \centering
  \caption{Statistics for the generating negative scene graph for \textsc{FINER-CompreCap} (denoted as C-SG) and \textsc{FINER-DOCCI} (denoted as D-SG).Counts: number of objects, attributes and relations inside the SG annotation.$\theta$: entropy-based filtering threshold; \#Re-gen.: number of re-generated negatives.}
  \label{tab:supp_neg_generation_stats}
  \setlength{\tabcolsep}{4pt}
  \renewcommand{\arraystretch}{1.05}
  \begin{tabular}{@{}llrrrr@{}}
    \toprule
    Benchmark &  & $\theta$ & Counts & \# Re-gen.  \\
    \midrule
    \multirow{3}{*}{C-SG}
      & Obj  & 0.8 & 3505 &  320 \\
      & Attr &  0.8 & 4509 & 414  \\
      & Rel  & 0.4 & 3494 & 173 \\
    \midrule
    \multirow{3}{*}{D-SG}
      & Obj  & 0.8 & 24,528 & 3,242 \\
      & Attr & 0.4 & 52,911 & 2,827  \\
      & Rel  & 0.8 & 15,342 & 2,143  \\
    \bottomrule
  \end{tabular}
\end{table}

\myparagraph{Quality Assessment.}
Given the scale of our benchmarks, we adopt a model-based assessment approach. We assess the quality of the generated negatives by evaluating Qwen2.5-VL-72B on objects (Obj), attributes (Attr), and relations (Rel) in \textsc{FINER-CompreCap} and \textsc{FINER-DOCCI}. Tab.~\ref{tab:negative_qualitative_assessments} reports the corresponding classification accuracies. For example, Qwen2.5-VL-72B achieves 94.1\% accuracy when selecting the positive relation from its four negative counterparts in \textsc{FINER-CompreCap}, which supports the quality of the constructed negatives in this benchmark. On \textsc{FINER-DOCCI}, the model attains close to 90\% accuracy in objects and attributes. Note that \textsc{FINER-DOCCI} is designed to test whether rich, human-described semantics can enable large-scale hallucination evaluation, rather than building a small, noise-free benchmark fully curated by humans. Given its substantially larger scale and higher difficulty, we consider the achieved negatives classification accuracies to show a sufficient negatives quality that helps validating our findings \textit{at scale}.

\begin{table}[t]
  \centering
  \caption{Quality assessment of generated negatives. We show the classification accuracy of Qwen2.5-VL-72B~\cite{bai2025qwen25vl} after classifying the objects (obj), attributes (attr) and relations (rel) in \textsc{FINER-CompreCap} and \textsc{FINER-DOCCI}}
  \label{tab:negative_qualitative_assessments}
  \setlength{\tabcolsep}{6pt}
  \renewcommand{\arraystretch}{1.05}
  \begin{tabular}{@{}lcccccc@{}}
    \toprule
    & \multicolumn{3}{c}{\textsc{FINER-CompreCap}} 
    & \multicolumn{3}{c}{\textsc{FINER-DOCCI}} \\
    \cmidrule(lr){2-4} \cmidrule(l){5-7}
    & obj & attr & rel & obj & attr & rel \\
    \midrule
    \dporow Acc. (\%) 
      & 89.8 & 91.1 & 94.1 
      & 89.5 & 88.3 & 82.8 \\
    \bottomrule
  \end{tabular}
\end{table}

\subsection{MCQ Design}
\label{sec:supp_mcq_design}
Having obtained the positive SG and negative SG for \textsc{FINER-CompreCap} and \textsc{FINER-DOCCI}, we now construct MCQs. Sec.~\ref*{sec:question_construction_pipeline} already provides an explanation of our MCQ construction pipeline: we use a fixed template to compose both positive and negative MCQs ($q^\pm_{\text{multi-obj}}$, $q^\pm_{\text{multi-attr}}$, $q^\pm_{\text{multi-rel}}$). For $q^\pm_{\text{Wh}}$, we prompt Gemini-2.0-Flash to construct the question templates. We describe the two templates in detail.

\myparagraph{Fixed question template.}
We use a simple yes/no-style template for all $q^\pm_{\text{multi-obj}}$, $q^\pm_{\text{multi-attr}}$, and $q^\pm_{\text{multi-rel}}$. To make the format explicit, we display it as a small template box:
\begin{center}
\fbox{%
  \parbox{0.96\linewidth}{%
    \small
    \texttt{Can you see }$\{X\}$\texttt{ in this image?}\\[4pt]
    \texttt{A. Yes, I can see }$\{Y\}$\texttt{ in this image.}\\[4pt]
    \texttt{B. No, but I can see }$\{Z_1\}$\texttt{ in this image.}\\[2pt]
    \texttt{C. No, but I can see }$\{Z_2\}$\texttt{ in this image.}\\[2pt]
    \texttt{D. No, but I can see }$\{Z_3\}$\texttt{ in this image.}\\[2pt]
    \texttt{E. No, but I can see }$\{Z_4\}$\texttt{ in this image.}
  }%
}
\end{center}

Here, $\{X\}$, $\{Y\}$, and $\{Z_1\},\dots,\{Z_4\}$ are placeholders that will later be filled with phrases. In the benchmark, the choices are randomly shuffled.

\myparagraph{Construction of $q^\pm_{\text{multi-obj}}$, $q^\pm_{\text{multi-attr}}$, and $q^\pm_{\text{multi-rel}}$.} We only describe the construction process for $q^\pm_{\text{multi-obj}}$; the same procedure is applied to $q^\pm_{\text{multi-attr}}$ and $q^\pm_{\text{multi-rel}}$.

From the positive SG of an image, we first sample $k$ distinct objects and concatenate them into a positive multi-object phrase $P_{\text{obj}}^{+}$ (for example, ``dog, ball, and tree''). This phrase $P_{\text{obj}}^{+}$ contains only objects that truly appear in the image. We then randomly select one of these $k$ objects, denote the selected object by $o$, and retrieve its four negative counterparts $\{o_j^{-}\}_{j=1}^{4}$ from the negative SG. For each $j \in \{1,\dots,4\}$, we form a corrupted phrase $P_{\text{obj},j}^{-}$ by replacing $o$ in $P_{\text{obj}}^{+}$ with $o_j^{-}$ while keeping all other objects unchanged. Thus we obtain one positive phrase $P_{\text{obj}}^{+}$ and four negative phrases $P_{\text{obj},1}^{-},\dots,P_{\text{obj},4}^{-}$.

To build a \emph{positive} MCQ $q^{+}_{\text{multi-obj}}$, we instantiate the template by setting
\[
\begin{aligned}
\{X\} &= P_{\text{obj}}^{+}, \\
\{Y\} &= P_{\text{obj}}^{+}, \\
\{Z_j\} &= P_{\text{obj},j}^{-} \ \text{for } j=1,\dots,4.
\end{aligned}
\]

In this case, the question and the ``Yes'' option both describe the true configuration $P_{\text{obj}}^{+}$, while each ``No, but I can see $\{Z_j\}$'' option contains exactly one incorrect object. The option that contains $P_{\text{obj}}^{+}$ is treated as the correct answer.

To build a \emph{negative} MCQ $q^{-}_{\text{multi-obj}}$, we flip the roles of the positive and corrupted phrases in the template. We randomly choose one corrupted phrase, say $P_{\text{obj},1}^{-}$, and set
\[
\begin{aligned}
\{X\}   &= P_{\text{obj},1}^{-}, \\
\{Y\}   &= P_{\text{obj},1}^{-}, \\
\{Z_1\} &= P_{\text{obj}}^{+}, \\
\{Z_j\} &= P_{\text{obj},j}^{-} \ \text{for } j=2,3,4.
\end{aligned}
\]

Now the question asks about the corrupted phrase $P_{\text{obj},1}^{-}$, which does \emph{not} match the image. Consequently, the ``Yes'' choice becomes a false-positive option, because it incorrectly confirms the existence of $P_{\text{obj},1}^{-}$. The option that says ``No, but I can see $P_{\text{obj}}^{+}$ in this image'' is now the correct answer, since it both denies the existence of $P_{\text{obj},1}^{-}$ and affirms the true configuration $P_{\text{obj}}^{+}$. Note that we randomly pick which corrupted phrase is used as the query, so each of $P_{\text{obj},1}^{-},\dots,P_{\text{obj},4}^{-}$ has an equal chance to replace $\{X\}$.

This fixed pattern keeps the surface form of the questions consistent across all MCQs while allowing the underlying content to vary. The same construction is applied to $q^\pm_{\text{multi-attr}}$ and $q^\pm_{\text{multi-rel}}$ by treating attribute phrases and relation phrases as the basic units instead of objects.

\myparagraph{Wh question generation.}
Wh questions have more flexible surface forms than yes/no questions. To construct Wh-style questions, we start from a relation triplet in the scene graph,
\[
(\obj_1, \rel, \obj_2),
\]
where $\obj_1$ and $\obj_2$ are two objects and $\rel$ is the relation between them. Each object can have one or more attributes, e.g.\ $\mathcal{A}(\obj_1)$ for the first object.

Given a triplet $(\obj_1, \rel, \obj_2)$, we randomly choose one of the two objects as the \emph{answer target} and treat the other as \emph{context}. Concretely, we either ask about $\obj_1$ given $\obj_2$ or about $\obj_2$ given $\obj_1$. We then mask the answer target in the textual description and prompt Gemini-2.0-Flash to produce a natural Wh question. For example, for the relation $\text{(dog, is standing under, table)}$,
Gemini-2.0-Flash can generate questions such as
\begin{align*}
  &\text{``What is standing under the table?''} \quad \text{(ask about the dog)}\\
  &\text{``What is the dog standing under?''} \quad \text{(ask about the table).}
\end{align*}

\myparagraph{Wh MCQ template.}
Once we fix the Wh question pattern for a given triplet, we turn it into an MCQ by providing five answer options. We represent the question body and the five options using placeholders:
\begin{center}
\fbox{%
  \parbox{0.96\linewidth}{%
    \small
    \texttt{Q: }$\{Q\}$\\[4pt]
    \texttt{A. }$\{O_1\}$\\[2pt]
    \texttt{B. }$\{O_2\}$\\[2pt]
    \texttt{C. }$\{O_3\}$\\[2pt]
    \texttt{D. }$\{O_4\}$\\[2pt]
    \texttt{E. }$\{C\}$%
  }%
}
\end{center}
Here, $\{Q\}$ is the Wh question text, $\{O_1\},\dots,\{O_4\}$ are object-level answer candidates, and $\{C\}$ is a full-sentence \emph{correction} option that explicitly talks about the attribute of the target object. In the benchmark, the choices are randomly shuffled.

\myparagraph{Construction of $q^\pm_{\text{Wh}}$.}
We illustrate the construction using the running example with the context object ``dog'' and the answer target ``table''. The dog has a positive attribute $A^{+}$ (e.g.\ ``with brown fur'') and a sampled negative attribute $A^{-}$ (e.g.\ ``with yellow fur''), while the relation and context (e.g.\ ``standing under the table'') are fixed by the triplet $(\obj_1, \rel, \obj_2)$.

From the positive SG, we select ``table'' as the target object $o^\star$. We then randomly pick three negative objects $o^{-}_1,o^{-}_2,o^{-}_3$ for this slot from the negative SG (e.g.\ ``chair'', ``bench'', ``sofa''). Starting from the Wh question
\[
\text{``What is the dog standing under?''},
\]
We insert an attribute phrase for the dog and obtain an attribute-conditional question template
\[
q(A) \equiv \text{``What is the dog } A \text{ standing under?''}.
\]
Filling this template with $A^{+}$ or $A^{-}$ gives us a positive or negative Wh question with the same surface pattern. Note that in the FINER benchmarks, a single object can have multiple attributes. In that case, we include all of its attributes in the descriptive context, then randomly choose one of them as the target attribute $A^{+}$ and sample the corresponding negative attribute as $A^{-}$.

\emph{Positive Wh MCQ.}
For the \emph{positive} Wh question $q^{+}_{\text{Wh}}$, we fill the attribute slot with the true attribute $A^{+}$ and instantiate the MCQ template as
\[
\begin{aligned}
\{Q\}   &= q(A^{+}), \\
\{O_1\} &= o^\star, \\
\{O_j\} &= o^{-}_{j-1} \quad \text{for } j=2,3,4, \\
\{C\}   &= \text{``The dog is not } A^{+} \text{, but is } A^{-}\text{.''}
\end{aligned}
\]
The question $\{Q\}$ is now a valid Wh question about the image, and $\{O_1\}$ (the true object $o^\star$) is the correct answer. The three options $\{O_2\},\{O_3\},\{O_4\}$ are incorrect objects, and the correction sentence $\{C\}$ is also incorrect because it denies the true attribute $A^{+}$.

\emph{Negative Wh MCQ.}
For the \emph{negative} Wh question $q^{-}_{\text{Wh}}$, we instead fill the question template with the negative attribute $A^{-}$, which makes the premise of the question partially inconsistent with the image. We keep the same four object candidates but flip the correction sentence:
\[
\begin{aligned}
\{Q\}   &= q(A^{-}), \\
\{O_1\} &= o^\star, \\
\{O_j\} &= o^{-}_{j-1} \quad \text{for } j=2,3,4, \\
\{C\}   &= \text{``The dog is not } A^{-} \text{, but is } A^{+}\text{.''}
\end{aligned}
\]
Now the question $\{Q\}$ is \emph{incorrect} with respect to the image, because it attributes $A^{-}$ to the dog. The object-only options $\{O_1\},\dots,\{O_4\}$ all implicitly accept the wrong attribute in the question and are therefore treated as incorrect. The correction option $\{C\}$ is the unique correct answer: it denies the wrong attribute $A^{-}$ and restores the true attribute $A^{+}$.

In summary, $q^{+}_{\text{Wh}}$ asks a Wh question whose premise matches the image and is answered by the true object $o^\star$, while $q^{-}_{\text{Wh}}$ asks a Wh question whose premise uses a corrupted attribute and is correctly answered only by the explicit correction sentence. This construction mirrors the positive/negative symmetry used for the yes/no-style templates and keeps the Wh MCQs tightly grounded in the underlying scene graph.

\myparagraph{Benchmark statistics.}
As described in Sec.~\ref*{sec:question_construction_pipeline}, our MCQ design constructs both positive and negative questions for four settings: $q^\pm_{\text{multi-obj}}$, $q^\pm_{\text{multi-attr}}$, $q^\pm_{\text{multi-rel}}$, and $q^\pm_{\text{wh}}$. We present the detailed statistics of \textsc{FINER-CompreCap} and \textsc{FINER-DOCCI} in Tab.~\ref{tab:supp_detailed_stats}.

\myparagraph{Post-hoc correction of MCQs.}
After constructing the MCQs for \textsc{FINER-CompreCap} and \textsc{FINER-DOCCI}, humans further corrected a subset of them: 100 MCQs per setting for \textsc{FINER-CompreCap} and 200 MCQs per setting for \textsc{FINER-DOCCI}. In the 3-relation subset of \textsc{FINER-DOCCI}, we additionally observed cases where multiple relations referred to the same objects. We therefore performed further human cleaning, resulting in 199 improved paired MCQs in this setting.

\begin{table}[t]
  \centering
  \caption{Distribution of MCQ pairs over entity counts in \textsc{FINER-CompreCap} (\textsc{FINER-C}) and \textsc{FINER-DOCCI} (\textsc{FINER-D}). For each setting, we refer the entity counts for Obj/Attr/Rel as $k$ and the corresponding number of pairs $n_k$ in matching order. $(1, 6)$ represents that $k$ ranges from 1 to 6.}
  \label{tab:supp_detailed_stats}
  \setlength{\tabcolsep}{4pt}
  \renewcommand{\arraystretch}{1.05}
  \begin{tabular}{@{}llll@{}}
    \toprule
    Benchmark & Setting    & $k$ & \# pairs $n_k$ \\
    \midrule
    \multirow{4}{*}{\textsc{FINER-C}} 
      & $q^\pm_{\text{multi-obj}}$  & $(1, 6)$      & $560,560,560,558,535,377$ \\
      & $q^\pm_{\text{multi-attr}}$ & $(1, 3)$            & $966,472,231$             \\
      & $q^\pm_{\text{multi-rel}}$  & $(1, 3)$            & $1217,616,307$            \\
      & $q^\pm_{\text{wh}}$  &          - &      $1583$      \\
    \midrule
    \multirow{4}{*}{\textsc{FINER-D}}    
      & $q^\pm_{\text{multi-obj}}$  & $(1, 6)$      & $65,496,909,980,874,1676$ \\
      & $q^\pm_{\text{multi-attr}}$ & $(1, 5)$        & $2451,5363,3092,1575,1843$ \\
      & $q^\pm_{\text{multi-rel}}$  & $(1,3)$            & $4404,1168,199$           \\
      & $q^\pm_{\text{wh}}$  &         -  &      $10472$      \\
    \bottomrule
  \end{tabular}
\end{table}

\section{Training Details}
\label{sec:supp_training_details}
Sec.~\ref*{sec:training_with_finer} explains our training data generation pipeline, on which \ours is trained. We also briefly describe the fine-tuning setup in Sec.~\ref*{sec:exp_exp_setup}. In this section, we first present concrete examples of the training data, and then provide the detailed fine-tuning configuration.

\begin{figure*}[t]
  \centering
  \includegraphics[width=\textwidth]{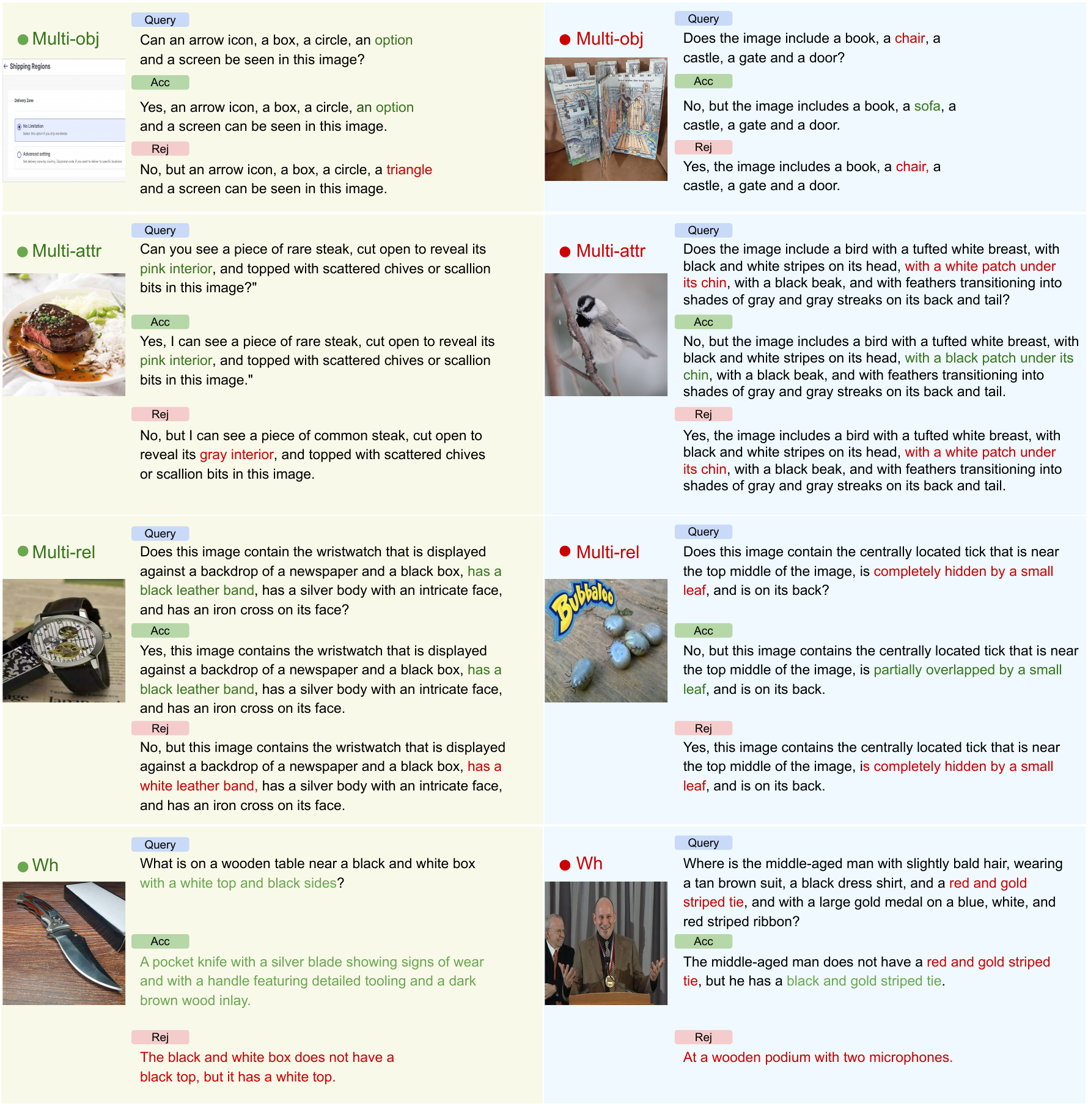}
  \caption{Examples from our constructed training set to train \ours. Positive queries are in green color, while negative queries are in red color. We show both positive ($(x, q_+, a^{+}_+, a^{-}_+)$) and negative ($(x, q_-, a^{+}_-, a^{-}_-)$) preference tuples across four subsets: Multi-obj, Multi-attr, Multi-rel, Wh.}
  \label{fig:supp_trainingset_examples}
\end{figure*}

\myparagraph{Training set examples.}
We apply the training data construction pipeline from Fig.~\ref*{fig:finer_dpo} to the first 24 shards of Pixmo-caption~\cite{deitke2025molmo}. As described in Sec.~\ref*{sec:training_with_finer}, each image $x$ can yield up to eight preference tuples $(x, q, a^{+}, a^{-})$ across the four subsets $\{\textsc{Obj}, \textsc{Attr}, \textsc{Rel}, \textsc{Wh}\}$. Applying the pipeline to 24 shards produces more than 1.6M preference tuples, which is more than we need for training. In practice, we only use the first 6 shards (about 440K tuples) and uniformly subsample at most 160K tuples for DPO training. We visualize representative training examples $(x, q, a^{+}, a^{-})$ from all four subsets in Fig.~\ref{fig:supp_trainingset_examples}.

\myparagraph{Finetuning Setup.}
We summarize the training hyperparameters for \ours in Tab.~\ref{tab:supp_finetune_hyperparams}. All models are trained with LLaMA-Factory~\cite{zheng2024llamafactory}, using LoRA~\cite{hu2022lora} as the parameter-efficient fine-tuning method. We apply LoRA adapters only to the projection layers $q_{\text{proj}}$ and $v_{\text{proj}}$. We reserve 0.5\% of the data as a validation set. Since the validation distribution closely matches the training distribution, we observe that training for too long drives the validation loss close to zero and brings little or no performance gain, sometimes even degrading downstream results. For DPO training, we therefore limit the number of training samples for each model: LLaVA-1.6 is trained on 40K examples, Qwen2.5-VL on 120K, and the InternVL3.5 series on 160K. For the SFT experiments in Tab.~\ref{tab:ablation_study}, we fine-tune InternVL3.5-8B on 160K examples with a learning rate of $1\times10^{-4}$. We use 4 NVIDIA H100 94GB GPUs to train InternVL3.5-14B, and 2 NVIDIA H100 GPUs for the other smaller models.

\begin{table}[t]
  \centering
  \resizebox{\linewidth}{!}{
  \setlength{\tabcolsep}{6pt}
  \renewcommand{\arraystretch}{1.05}
  \begin{tabular}{@{}lcccc@{}}
    \toprule
    Config & Llava-1.6-7B & Qwen2.5VL-7B & InternVL3.5-8B & InternVL3.5-14B \\
    \midrule
    
    Training Data &    40K   &    120K     &   160K   &     160K   \\
    Global BS     &     \multicolumn{4}{c}{64}    \\
    Optimizer      & \multicolumn{4}{c}{AdamW~\cite{loshchilovadamw}} \\
    Learning rate  & \multicolumn{4}{c}{$5 \times 10^{-6}$} \\
    Total epochs   & \multicolumn{4}{c}{1} \\
    Warm up  ratio     & \multicolumn{4}{c}{0.1} \\
    LR scheduler   & \multicolumn{4}{c}{cosine decay} \\
    LoRA rank   & \multicolumn{4}{c}{32} \\
    LoRA target   & \multicolumn{4}{c}{$q_{\text{proj}}$, $v_{\text{proj}}$} \\
    $\beta$   & \multicolumn{4}{c}{0.1} \\
    Val. ratio   & \multicolumn{4}{c}{0.005} \\
    \bottomrule
  \end{tabular}
  }
  \caption{Fine-tuning hyper-parameters for \ours on all baselines. Global BS: global batch size. LR scheduler: learning rate scheduler. $\beta$: inverse temperature parameter in the DPO loss, as shown in Eq.~\ref*{eq:dpo-core}. Val. ratio: ratio of validation data size.}
  \label{tab:supp_finetune_hyperparams}
\end{table}

\section{Evaluation Details}
\label{sec:supp_evaluation_details}
We detail the evaluation setups for three groups of tasks: the FINER benchmarks, other hallucination benchmarks, and general capabilities.

\myparagraph{FINER benchmarks.}
Since the FINER benchmarks are multiple-choice (MCQ) benchmarks, we evaluate all models using greedy decoding with temperature $0$, no sampling, and a maximum of 3 output tokens. Given an image and an MCQ, we append the instruction: \texttt{``Please answer with a single capital letter (A, B, C, D, or E).''} We compute the paired accuracy $\text{Acc}_{\text{paired}}$, which counts a pair as correct only if the model answers both $q^{+}$ and $q^{-}$ correctly, ensuring that the model does not systematically favor either the positive or the negative version.

\myparagraph{Other hallucination benchmarks.}
We evaluate all models on both discriminative hallucination benchmarks (DASH~\cite{augustin2025dash}, POPE~\cite{li2023pope}, RePOPE~\cite{neuhaus2025repope}, HallusionBench~\cite{guan2024hallusionbench}, AMBER~\cite{wang2023amber}, CRPE\_R~\cite{wang2024allseeing_v2}) and generative hallucination benchmarks (MMHal-Bench~\cite{sun2023llava-rlhf}, HaloQuest~\cite{wang2024haloquest}). 

We use VLMEvalKit~\cite{duan2024vlmevalkit} to evaluate HallusionBench, AMBER, and CRPE\_R with their default configuration. We report all accuracy (aAcc.) for HallusionBench and averaged accuracy for CRPE\_R. For DASH, POPE, and RePOPE, we follow their official evaluation protocols and prompt models to answer only with \texttt{``yes''} or \texttt{``no''}. We again adopt greedy decoding for this binary setting to keep the setup consistent across models. We report the averaged accuracy in Tab.~\ref*{tab:other_hallu_bench} and show the accuracy on each subset in Tab.~\ref{tab:supp_other_hallu_bench}.

For MMHal-Bench, we use the original evaluation code but replace the judge model with GPT-4.1-mini~\cite{achiam2023gpt}, since the original judge has been deprecated. For HaloQuest, we similarly follow the released evaluation pipeline but replace the judge with Gemini-2.0-Flash~\cite{team2023gemini}, as Gemini-1.5-Pro is no longer accessible. In both generative benchmarks, we use temperature $0$ to ensure reproducible results. We follow the metrics of both benchmarks, reporting score (max. 6) as well as hallucination rate in MMHal-Bench, as well as the averaged score in HaloQuest.

\myparagraph{General capabilities.}
We evaluate general capabilities using six benchmarks: MMStar~\cite{chen2024mmstar} (broad multi-skill evaluation), TextVQA~\cite{singh2019textvqa} (text understanding from images), ChartQA~\cite{masry2022chartqa} (chart and figure understanding), MMVP~\cite{tong2024eyes} (vision-centric reasoning), NaturalBench~\cite{li2024naturalbench} (natural, compositional multi-step reasoning), and V$^*$ (visual search on high-resolution images). NaturalBench contains grouped, real-world questions that require models to jointly use perception, world knowledge, and compositional reasoning, making it a challenging test of robust, general-purpose vision-language ability.

We use VLMEvalKit~\cite{duan2024vlmevalkit} with default settings to evaluate all models on these six benchmarks. We report overall accuracy for MMStar, TextVQA, ChartQA, MMVP, and V$^*$. For NaturalBench, we report group accuracy (G\_ACC), as it is the most stringent and informative metric.

\section{Additional Experiments}
\label{sec:supp_additional_experiments}
\label{sec:supp_additional_ablations}

Despite the main experimental results presented in Sec.~\ref*{sec:exp}, we report additional experiments in this section. Specifically, we conduct a positional bias study (Sec.~\ref{sec:supp_exp_pos_bias}), analyze the impact of training data filtering (Sec.~\ref{sec:supp_exp_data_filtering}), present more qualitative results from \textsc{FINER-DOCCI} (Sec.~\ref{sec:supp_exp_qualitative_results}), provide per-subset results of three benchmarks (Sec.~\ref{sec:supp_exp_per_subset}), provide an extended comparison with additional hallucination reduction methods (Sec.~\ref{sec:supp_exp_comparing_more_methods}), provide a brief discussion of an alternative random guess baseline (Sec.~\ref{sec:supp_exp_random_guess}), and show results on the MCQ version of our motivational study (Sec.~\ref{sec:supp_mcq_version_motivational_study}).

\subsection{Positional bias study}
\label{sec:supp_exp_pos_bias}
Both \textsc{FINER-CompreCap} and \textsc{FINER-DOCCI} contain MCQs that involve multiple objects, attributes, and relations ($q^\pm_{\text{multi-obj}}$, $q^\pm_{\text{multi-attr}}$, and $q^\pm_{\text{multi-rel}}$). When constructing a negative MCQ $q^-$, we choose one entity (object, attribute, or relation) at a random position and replace it with its negative counterpart. A natural question is whether the model’s behavior depends on which position is negated.

To test this, for all $q^\pm_{\text{multi-obj}}$, $q^\pm_{\text{multi-attr}}$, and $q^\pm_{\text{multi-rel}}$ with exactly three entities, we keep the same triplet but rotate which entity is negated, so that the negative appears once in each of the three positions. We then measure the paired accuracy $\text{Acc}_{\text{paired}}$ for each position. As shown in Fig.~\ref{fig:supp_pos_bias}, base models exhibit clear positional bias. For example, in $q^\pm_{\text{multi-obj}}$, LLaVA-Next performs much worse when the negative is in the middle position, and Qwen2.5-VL-7B shows a drop of about 15\% when the last position is negated compared to the first. In $q^\pm_{\text{multi-rel}}$, the preferred position even differs across models: InternVL3.5-8B achieves the highest accuracy when negating the middle entity, while InternVL3.5-14B peaks when the third entity is negated. Fine-tuning with \ours consistently improves accuracy at all positions, but the curves are still not flat, indicating that positional bias remains. We suspect this is related to the inherent sequence structure of current MLLM architectures and leave a deeper investigation to future work. We also assume that the current MCQ format is not the best option for testing positional bias, and we are looking forward the community to dive deeper into language positional bias in open-ended generation questions.

\subsection{Ablation: Training Data Filtering}
\label{sec:supp_exp_data_filtering}
In Pixmo-caption~\cite{deitke2025molmo}, we observed that certain amount of  of images are charts/graphs or screenshots: content outside the evaluation scope of \textsc{FINER-CompreCap} and \textsc{FINER-DOCCI} (which target natural images). For example, one screenshot image can be found in the upper left corner of Fig.~\ref{fig:supp_trainingset_examples}. Therefore, we first run Phi-4-14B over all the long captions to classify the images into four categories: ``natural images", ``screenshot\_ui", ``chart\_graph" and ``document\_text". Since FINER benchmarks target only natural images. The statistics are in Tab.~\ref{tab:supp_filtering_abalation}. Excluding these images resulted in almost no significant difference in performance. Therefore, to maintain simplicity and generality, we do not apply any filtering and retain the original dataset composition.

\begin{table}[t]
  \centering
  \caption{Category statistics for Pixmo-caption~\cite{deitke2025molmo}.}
  \label{tab:supp_cat_images}
  \setlength{\tabcolsep}{6pt}
  \renewcommand{\arraystretch}{1.05}
  \begin{tabular}{@{}lrr@{}}
    \toprule
    Category        & Count  & Percentage \\ 
    \midrule
    natural\_image  & 176{,}881 & 78.13\% \\
    screenshot\_ui  & 36{,}701  & 16.21\% \\
    chart\_graph    & 8{,}061   & 3.56\%  \\
    document\_text  & 4{,}739   & 2.09\%  \\
    \bottomrule
  \end{tabular}
\end{table}

\begin{table}[t]
 \caption{Filtering to only keep natural images ablation for \ours with InternVL-3.5-8B~\cite{wang2025internvl35}. Obj/Attr/Rel denote Multi-obj/Multi-attr/Multi-rel for both training and evaluation. The best results are bold.}
  \label{tab:supp_filtering_abalation}
  \centering
  \resizebox{0.8\linewidth}{!}{
  \setlength{\tabcolsep}{3pt}
  \renewcommand{\arraystretch}{1.05}
  \begin{tabular}{@{}l c c c c c c@{}}
    \toprule
    \multicolumn{1}{@{}l}{Fitered?} & \multicolumn{4}{c}{FINER-CompreCap} & \multicolumn{2}{c}{Other} \\
    \cmidrule(lr){2-5} \cmidrule(lr){6-7}
    & Obj & Attr & Rel & Wh & RePOPE & M.S. \\
    \midrule
     \baserow -  &  74.2 & 71.9 &  49.8 &  25.5 & 91.5 & 68.0   \\
     Yes    &  \textbf{76.8}  & \textbf{78.6} & 62.8 & \textbf{36.1}  & \textbf{93.1} &  68.1 \\
     \dporow No  &76.5  & 78.3 & \textbf{64.1} & \textbf{36.1}  & \textbf{93.1} & \textbf{68.3}  \\
    \bottomrule
  \end{tabular}
  }
\end{table}

\subsection{Qualitative Results}
\label{sec:supp_exp_qualitative_results}
\begin{figure*}[tbp]
  \centering
  \includegraphics[width=\textwidth]{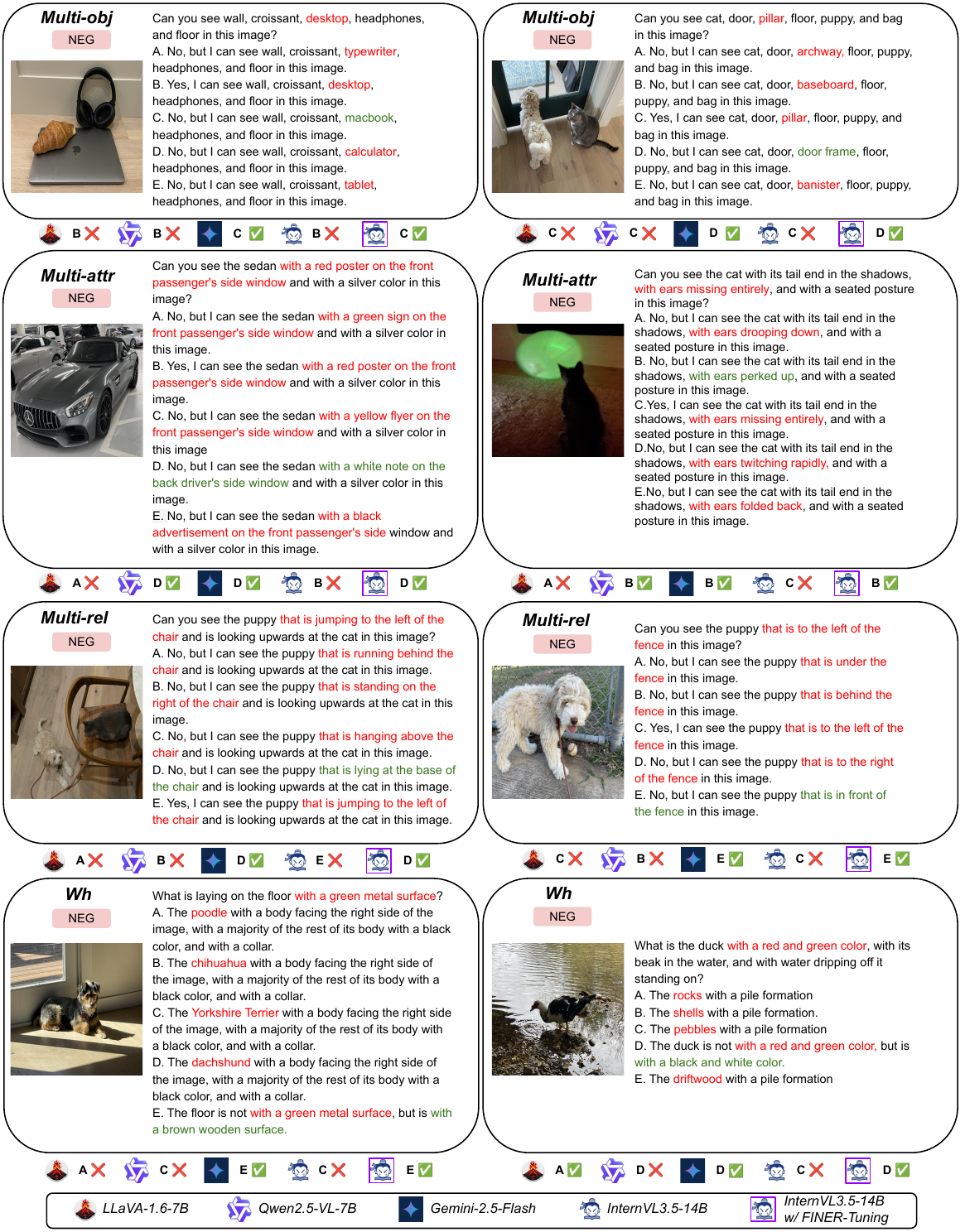}
  \caption{Qualitative Results from FINER-DOCCI.}
  \label{fig:supp_qualitative_finer_docci}
\end{figure*}

Following the qualitative results in Sec.~\ref*{sec:exp_qualitative_results} on \textsc{FINER-CompreCap}, we provide additional examples from \textsc{FINER-DOCCI} in Fig.~\ref{fig:supp_qualitative_finer_docci}. These cases cover all four settings: Multi-obj, Multi-attr, Multi-rel, and Wh. We only visualize the negative MCQs here, as they are much more challenging than their positive counterparts. However, some positive MCQs can be found in our human study examples (Fig.~\ref{fig:supp_docci_human_study_examples} and Fig.~\ref{fig:supp_comprecap_human_study_examples}).

As shown in Fig.~\ref{fig:supp_qualitative_finer_docci}, in the Multi-obj setting, only Gemini-2.5-Flash~\cite{comanici2025gemini25} and our \ours-tuned InternVL3.5-14B reliably identify the fine-grained concept ``macbook''. In the Multi-attr setting, the questions target subtle details such as ``the white note on the back driver's side window'' or ``the cat with perked-up ears''. In the Multi-rel setting, some models, such as Qwen2.5-VL-7B~\cite{bai2025qwen25vl}, hallucinate the dog as being ``behind the fence'', even though it is clearly in front of the fence. Finally, in the Wh setting, only Gemini and \ours correctly detect the anomalous attributes of the floor and the duck and answer the questions accordingly.

\subsection{Per-subset results}
\label{sec:supp_exp_per_subset}
\myparagraph{POPE, RePOPE, AMBER.} In Sec.~\ref*{sec:exp_other_hallu}, we report the averaged performance on POPE~\cite{li2023pope}, RePOPE~\cite{neuhaus2025repope}, and AMBER discriminative subset~\cite{wang2023amber} (denoted as AMBER throughout this paper). In Tab.~\ref{tab:supp_other_hallu_bench}, we further break down the results and report the accuracy for each subset of these three benchmarks. Notably, with \ours, LLaVA-1.6 achieves a 20.1\% absolute improvement on AMBER, further demonstrating the effectiveness of \ours.

\myparagraph{HallBench, CRPE\_R, HaloQuest.} Apart from the per-subset results reported in Tab.~\ref{tab:supp_other_hallu_bench}, we further report detailed breakdowns for HallBench~\cite{guan2024hallusionbench}, CRPE\_R~\cite{wang2024allseeing_v2} and HaloQuest~\cite{wang2024haloquest} in Tab.~\ref{tab:supp_other_hallu2}. To further probe the captioning capabilities of different models, we include the results for AMBER generative subset (AMBER\_G) and report four metrics: CHAIR (CH.), COVER (CO.), Hal. and Cog. in Tab.~\ref{tab:supp_other_hallu2}. On Hallbench, \ours improves over all baselines by maximally 6.8\% (fAcc. of LLaVA-1.6), showcasing that \ours can still work effectively in reducing general halucinations. In HaloQuest, the performance gain is mainly in Insufficient Context (IC.) subset and false premise (FP.) subset. Some catchy improvements are: \ours improves LLaVA-1.6 by 19.0\% on IC and 31\% on FP. \ours also improves the latest InternVL-3.5-8B by 15.7\% and 15.3\% each. Note that HaloQuest is a free-form generative benchmark. This shows that \ours can effectively correct the false premise hallucinations or withhold over-confident preidctions in free-form generations. 

\myparagraph{AMBER\_G.} To further probe the captioning capabilities of different models, we include the results for AMBER generative subset (AMBER\_G) and report four metrics: CHAIR, COVER, Hal and Cog in Tab.~\ref{tab:supp_ext_results_amberg}.
Lastly, \ours consistently improves over three baselines (Qwen2.5-VL-7B, InternVL-3.5-8B, InternVL3.5-14B) on AMBER\_G. We therefore think that when the base models are strong enough, \ours can further improve the captioning capabilities of the model.

\begin{table*}[t]
  \centering
  \small
  \setlength{\tabcolsep}{4pt}
  \renewcommand{\arraystretch}{1.05}
  \begin{tabular}{@{}l c ccc ccc ccc ccc@{}}
    \toprule
    & & \multicolumn{3}{c}{POPE} & \multicolumn{3}{c}{RePOPE} & \multicolumn{3}{c}{AMBER} \\
    \cmidrule(lr){3-5} \cmidrule(lr){6-8} \cmidrule(lr){9-11} 
    Models & Size 
      & Ran.\ {\small$\uparrow$} & Pop.\ {\small$\uparrow$} & Adv.\ {\small$\uparrow$}
      & Ran.\ {\small$\uparrow$} & Pop.\ {\small$\uparrow$} & Adv.\ {\small$\uparrow$}
      & Exis.\ {\small$\uparrow$} & Attr.\ {\small$\uparrow$} & Rel.\ {\small$\uparrow$}\\
    \midrule
    OmniLMM  & 12B 
      & 89.3 & 87.8 & 87.1 
      & 95.1 & 93.2 & 93.1
      & 85.6 & 94.2 & 80.7  \\
\baserow \quad +RLAIF-V  & 12B
      & 89.0\dminus{0.3} & 87.5\dminus{0.3} & 86.8\dminus{0.3}
      & 95.0\dminus{0.1} & 92.8\dminus{0.4} & 92.6\dminus{0.5}
      & 86.1\dplus{0.5} & 90.2\dminus{4.0} & 85.7\dplus{5.0} \\
    \addlinespace[2pt]
    \hline
    \addlinespace[2pt]
    LLaVA-1.6~\cite{liu2024llavanext}  & 7B
      & 89.7 & 88.4 & 86.6
      & 93.9 & 92.1 & 91.0
      & 82.0 & 93.6 & 58.7 \\
\dporow \quad +\ours   &  7B
      & 90.4\dplus{0.7} & 88.8\dplus{0.4} & 87.2\dplus{0.6}
      & 94.9\dplus{1.0} & 92.9\dplus{0.8} & 91.8\dplus{0.8}
      & 83.5\dplus{1.5} & 92.6\dminus{1.0} & 78.8\dplus{20.1} \\
    \addlinespace[2pt]
    Qwen2.5-VL~\cite{bai2025qwen25vl}  &  7B
      & 87.0 & 86.5 & 85.8
      & 93.6 & 91.9 & 91.7
      & 84.1 & 95.7 & 75.6 \\
\dporow \quad +\ours  &  7B
      & 88.0\dplus{1.0} & 87.0\dplus{0.5} & 86.4\dplus{0.6}
      & 94.1\dplus{0.5} & 92.2\dplus{0.3} & 91.9\dplus{0.2}
      & 84.0\dminus{0.1} & 96.2\dplus{0.5} & 77.1\dplus{1.5} \\
    \addlinespace[2pt]
    InternVL-3.5~\cite{wang2025internvl35}  &  8B
      & 93.3 & 87.7 & 85.0
      & 95.4 & 90.7 & 88.5
      & 80.4 & 88.0 & 80.1 \\
\dporow \quad +\ours     &   8B
      & 92.7\dminus{0.6} & 88.7\dplus{1.0} & 86.6\dplus{1.6}
      & 95.9\dplus{0.5} & 92.6\dplus{1.9} & 90.9\dplus{2.4}
      & 80.6\dplus{0.2} & 88.2\dplus{0.2} & 80.6\dplus{0.5} \\
    InternVL-3.5~\cite{wang2025internvl35}  &  14B
      & 93.4 & 89.6 & 85.7
      & 94.7 & 92.1 & 88.8
      & 82.6 & 89.4 & 81.9 \\
\dporow \quad +\ours     &   14B
      & 93.0\dminus{0.4} & 90.2\dplus{0.6} & 87.3\dplus{1.6}
      & 95.8\dplus{1.1} & 93.6\dplus{1.5} & 91.4\dplus{2.6}
      & 82.5\dminus{0.1} & 91.0\dplus{1.6} & 81.5\dminus{0.4} \\
    \bottomrule
  \end{tabular}
  \caption{Per-subset results on POPE~\cite{li2023pope}, RePOPE~\cite{neuhaus2025repope}, and AMBER~\cite{wang2023amber}. Rand.: Random; Pop.: Popular; Adv.: Adversarial; Exis.: Existence; Attr.: Attribute; Rel.: Relation}
  \label{tab:supp_other_hallu_bench}
\end{table*}

\begin{table*}[h]
  \centering
  \small
  \setlength{\tabcolsep}{4pt}
  \renewcommand{\arraystretch}{1.05}
  \begin{tabular}{@{}l ccc cccc ccc@{}}
    \toprule
    &  \multicolumn{3}{c}{HallBench} & \multicolumn{4}{c}{CRPE\_R} & \multicolumn{3}{c}{HaloQuest} \\
    \cmidrule(lr){2-4} \cmidrule(lr){5-8} \cmidrule(lr){9-11}
    Models
      & aAcc.\ {\small$\uparrow$} & fAcc.\ {\small$\uparrow$} & qAcc.\ {\small$\uparrow$}
      & Sub.\ {\small$\uparrow$} & Pred.\ {\small$\uparrow$} & Obj.\ {\small$\uparrow$} & Tot.\ {\small$\uparrow$}
      & VC.\ {\small$\uparrow$} & IC.\ {\small$\uparrow$} & FP.\ {\small$\uparrow$}\\
    \midrule
    LLaVA-1.6-7B
      & 33.0 & 10.6 & 8.3
      & 61.7 & 52.6 & 61.6 & 56.5
      & 50.5 & 38.0 & 42.9 \\
    \dporow \quad +\ours
      & 36.3\dplus{3.3} & 17.4\dplus{6.8} & 13.0\dplus{4.7}
      & 62.6\dplus{0.9} & 51.7\dminus{0.9} & 59.8\dminus{1.8} & 56.0\dminus{0.5}
      & 50.5 & 57.0\dplus{19.0} & 73.9\dplus{31.0} \\
    Qwen2.5-VL-7B
      & 65.4 & 35.8 & 40.0
      & 77.2 & 66.1 & 71.7 & 69.9
      & 66.5 & 76.0 & 79.2 \\
    \dporow \quad +\ours
      & 68.5\dplus{3.1} & 40.0\dplus{4.2} & 43.6\dplus{3.6}
      & 77.9\dplus{0.7} & 67.0\dplus{0.9} & 72.4\dplus{0.7} & 70.7\dplus{0.8}
      & 65.9\dminus{0.6} & 86.7\dplus{10.7} & 87.5\dplus{8.3} \\
    InternVL-3.5-8B
      & 71.0 & 45.1 & 47.0
      & 75.6 & 63.3 & 70.8 & 67.7
      & 66.5 & 51.2 & 64.4 \\
    \dporow \quad +\ours
      & 73.0\dplus{2.0} & 48.9\dplus{3.8} & 49.3\dplus{2.3}
      & 76.5\dplus{0.9} & 63.4\dplus{0.1} & 70.9\dplus{0.1} & 68.0\dplus{0.3}
      & 65.9\dminus{0.6} & 66.9\dplus{15.7} & 80.7\dplus{15.3} \\
    InternVL-3.5-14B
      & 69.5 & 46.8 & 47.0
      & 77.2 & 60.7 & 73.3 & 67.1
      & 63.7 & 54.5 & 70.0 \\
    \dporow \quad +\ours
      & 71.2\dplus{1.7} & 49.2\dplus{2.4} & 49.7\dplus{2.7}
      & 78.5\dplus{1.3} & 63.1\dplus{2.4} & 73.9\dplus{0.6} & 68.9\dplus{1.8}
      & 63.7 & 61.2\dplus{6.7} & 79.2\dplus{9.2} \\
    \bottomrule
  \end{tabular}
  \caption{Per-subset results on HallBench~\cite{guan2024hallusionbench}, CRPE relation subset (CRPE\_R)~\cite{wang2024allseeing_v2}, and HaloQuest~\cite{wang2024haloquest}. Sub.: Subject; Pred.: Predicate; Obj.:Object; Tot.: Total; VC.::Visually Challenge subset; IC.: Insufficient Context subset; FP.: False Premise subset;}
  \label{tab:supp_other_hallu2}
\end{table*}

\begin{table}[h]
  \centering
  \small
  \setlength{\tabcolsep}{3pt}
  \renewcommand{\arraystretch}{1.05}
  \resizebox{0.9\columnwidth}{!}{
  \begin{tabular}{@{}l cccc@{}}
    \toprule
    & \multicolumn{4}{c}{AMBER\_G} \\
    \cmidrule(lr){2-5}
    Models
      & CHAIR\ {\small$\downarrow$} & COVER\ {\small$\uparrow$} & Hal\ {\small$\downarrow$} & Cog\ {\small$\downarrow$} \\
    \midrule
    Qwen2.5-VL-7B
      & 5.3 & 64.0 & 27.1 & 1.9 \\
    \dporow \quad +\ours
      & 5.0\dplus{0.3} & 64.7\dplus{0.7} & 25.9\dplus{1.2} & 1.6\dplus{0.3} \\
    InternVL-3.5-8B
      & 6.9 & 61.3 & 49.9 & 3.1 \\
    \dporow \quad +\ours
      & 6.3\dplus{0.6} & 61.4\dplus{0.1} & 47.0\dplus{2.9} & 2.5\dplus{0.6} \\
    InternVL-3.5-14B
      & 7.9 & 68.6 & 57.6 & 5.4 \\
    \dporow \quad +\ours
      & 7.4\dplus{0.5} & 68.7\dplus{0.1} & 54.4\dplus{3.2} & 4.4\dplus{1.0} \\
    \bottomrule
  \end{tabular}
  }
  \caption{Extended results on AMBER generative subset (AMBER\_G).}
  \label{tab:supp_ext_results_amberg}
\end{table}

\begin{table}[t]
  \centering
  \resizebox{0.95\linewidth}{!}{
  \setlength{\tabcolsep}{3pt}
  \renewcommand{\arraystretch}{1.05}
  \begin{tabular}{@{}l c c c c@{}}
    \toprule
    \multicolumn{1}{@{}l}{Method} & \multicolumn{1}{c}{POPE} & \multicolumn{1}{c}{AMBER} &\multicolumn{1}{c}{MMHal} & \multicolumn{1}{c}{HaloQuest}  \\
    \cmidrule(lr){2-2} \cmidrule(lr){3-3} \cmidrule(lr){4-4} \cmidrule(l){5-5}
    & Acc. $\uparrow$ & Acc. $\uparrow$ & HR. $\downarrow$ & Score $\uparrow$ \\
    \midrule
     \baserow LLaVA-1.5-7B & 85.9 &  74.7  & 54.0 & 22.6 \\
     +HALVA~\cite{sarkar2024halva} & 84.8 &  \textbf{83.4}  & 54.0 & 23.9 \\
     +HA-DPO~\cite{zhao2023hadpo} & \textbf{86.9} &  78.1  & 60.0 & - \\
     +DoLA~\cite{chuang2023dola} & 85.7 &  74.5  & 56.0 & 22.9 \\
     +RLAIF-V~\cite{yu2024rlaifv} & 85.2 &  76.8  & 32.3 & - \\
     +REVERSE~\cite{wu2025reverse}  & 85.9 &  74.2  & \textbf{30.0} & \underline{32.3} \\
     \dporow +\ours  & \underline{86.7} &  \underline{82.3}  & \underline{49.0} &  \textbf{38.8} \\
    \bottomrule
  \end{tabular}
  }
   \caption{Extended comparison with other hallucination reduction methods on LLaVA-1.5-7B~\cite{liu2024improved}. HR.: Hallucination rate. The best results are bold while the second best results are underlined.}
   \label{tab:extended_comparison}
\end{table}

\subsection{Comparing with more methods}
\label{sec:supp_exp_comparing_more_methods}
It is challenging to totally fairly compare hallucination reduction methods because they are often trained on different datasets and base models. In this section, we fine-tune LLaVA-1.5-7B~\cite{liu2024improved} with \ours using 40K training examples from our dataset. We then evaluate on discriminative hallucination benchmarks (POPE~\cite{li2023pope}, AMBER~\cite{wang2023amber}) and generative benchmarks (MMHal-Bench (MMHal)~\cite{sun2023llava-rlhf} and HaloQuest~\cite{wang2024haloquest}). We compare against the state-of-the-art REVERSE~\cite{wu2025reverse}, as well as DoLA~\cite{chuang2023dola}, HA-DPO~\cite{zhao2023hadpo}, and HALVA~\cite{sarkar2024halva}. We also compare \ours with RLAIF-V-7B~\cite{yu2024rlaifv} on the same LLaVA-1.5-7B base model, resulting in a more direct comparison than Tab.~\ref{tab:main} and Tab.~\ref{tab:other_hallu_bench}. The results are in Tab.~\ref{tab:extended_comparison}.

Using 40K training samples curated by Phi-4-14B~\cite{abdin2024phi}, \ours already achieves comparable performance on discriminative benchmarks to HALVA and HA-DPO, whose training data are curated by Gemini Vision Pro~\cite{team2023gemini} and GPT-4~\cite{achiam2023gpt}, respectively, while substantially outperforming them on generative benchmarks. Compared with the SOTA method REVERSE, \ours matches or surpasses its performance on discriminative tasks and further improves HaloQuest by 6.3\%, but still lags behind on MMHal-Bench. Overall, these results indicate that \ours is effective at reducing hallucinations, and its benefits appear more pronounced when applied to stronger, frontier MLLMs, as also evidenced in Tab.~\ref*{tab:other_hallu_bench}. Compared to RLAIF-V, \ours performs better on discriminative benchmarks such as POPE and AMBER (a +5.5\% gain on AMBER), but remains weaker on generative benchmarks like MMHal-Bench. 

\subsection{Smarter random guess baselines}
\label{sec:supp_exp_random_guess}

In Tab.~1, we report a uniform random-guess baseline of $4\%$, which corresponds to independently sampling one out of five answer options for both the positive and negative questions: $(1/5)^2$.

However, due to the structured answer space in our Multi-obj/Multi-attr/Multi-rel MCQs (one \texttt{Yes, I can see...} option and four \texttt{No, but I can see...} options), a stronger no-knowledge baseline is a \emph{polarity-aware} random guesser. Specifically, it first guesses the polarity (\texttt{Yes} vs.\ \texttt{No}) uniformly, and if it guesses \texttt{No}, it then uniformly selects one of the four \texttt{No} options.

Since each pair consists of one positive question whose ground-truth is always \texttt{Yes} and one negative question whose ground-truth is always one of the four \texttt{No} options, the probability of guessing correctly is $0.5$ for a positive MCQ. For the negative MCQ, it is $0.5 \times 0.25$. Therefore, the paired accuracy is $0.5 \times (0.5 \times 0.25) = 0.0625$.

\subsection{MCQ Version of the Motivational Study}
\label{sec:supp_mcq_version_motivational_study}
\begin{figure}[h]
  \centering
  \includegraphics[width=\linewidth]{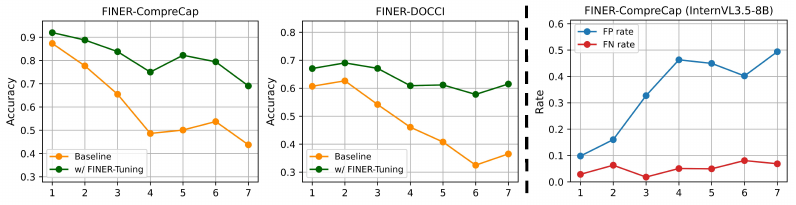}
  \caption{Left: MCQ version of the motivational study. Right: False-positive (FP) and false-negative (FN) rates at each granularity level.}
  \label{fig:supp_teaser_mcq}
\end{figure}

Yes/no probing is standard in prior benchmarks such as DASH, POPE, and AMBER for evaluating \emph{false-positive hallucinations}. In the main paper, we adopt this simple setup for the motivational study because it is easy to understand. In contrast, our FINER benchmarks are evaluated using multiple-choice questions (MCQs). Using two different evaluation protocols may cause confusion for some readers. Therefore, we additionally reformulate the motivational study in the same MCQ format as used in our benchmarks. Fig.~\ref{fig:supp_teaser_mcq} shows the same trend as the yes/no version in the main paper: accuracy decreases as query granularity increases. More specifically, the false-positive (FP) rate is much higher than the false-negative (FN) rate, confirming that false-positive hallucination is the main cause of the performance drop.

\section{Human Study}
\label{sec:supp_human_study}

\begin{table}[t]
  \centering
  \small
  \setlength{\tabcolsep}{3.5pt}
  \resizebox{\columnwidth}{!}{
  \begin{tabular}{@{}cccccccc@{}}
    \toprule
    \multicolumn{4}{c}{FINER-CompreCap} & \multicolumn{4}{c}{FINER-DOCCI} \\
    \cmidrule(lr){1-4} \cmidrule(lr){5-8}
    Multi-obj & Multi-attr & Multi-rel & Wh
    & Multi-obj & Multi-attr & Multi-rel & Wh \\
    \midrule
    \baserow 92.5 & 92.5 & 97.5 & 95.0
    & 92.5 & 95.0 & 90.0 & 90.0 \\
    \bottomrule
  \end{tabular}
  }
  \caption{Human performance in paired accuracy ($\text{Acc}_{\text{paired}}$) on FINER-CompreCap and FINER-DOCCI.}
  \label{tab:supp_human_study}
\end{table}

Since the FINER benchmarks are text-intensive, we asked human participants to answer a limited number of questions: 20 MCQs per subset. With eight subsets in total (four from \textsc{FINER-CompreCap} and four from \textsc{FINER-DOCCI}), this yields 160 MCQs. The results are shown in Tab.~\ref{tab:supp_human_study}.

Unlike models, which answer the positive and negative versions of each MCQ independently, humans could in principle remember a MCQ and use the correspondence between $q^{+}$ and $q^{-}$ to make the task easier. To avoid this, we create two versions (A and B) for each setting. For every MCQ pair, the positive and negative versions are randomly assigned to different versions. Each annotator only sees one version (either A or B), so they never see both sides of the same pair.

We recruit four human participants for each setting and compute paired accuracy based on their responses. The numerical results are reported in Tab.~\ref*{tab:main}. Example survey pages from our human study are shown for Multi-rel and Wh questions from \textsc{FINER-CompreCap} in Fig.~\ref{fig:supp_comprecap_human_study_examples}, and for Multi-obj and Multi-attr questions from \textsc{FINER-DOCCI} in Fig.~\ref{fig:supp_docci_human_study_examples}. As illustrated in these figures, each MCQ has two versions (A and B), corresponding to its positive and negative forms, and no annotator ever answers both versions of the same MCQ.

\begin{figure*}[t]
  \centering
  \includegraphics[width=\textwidth]{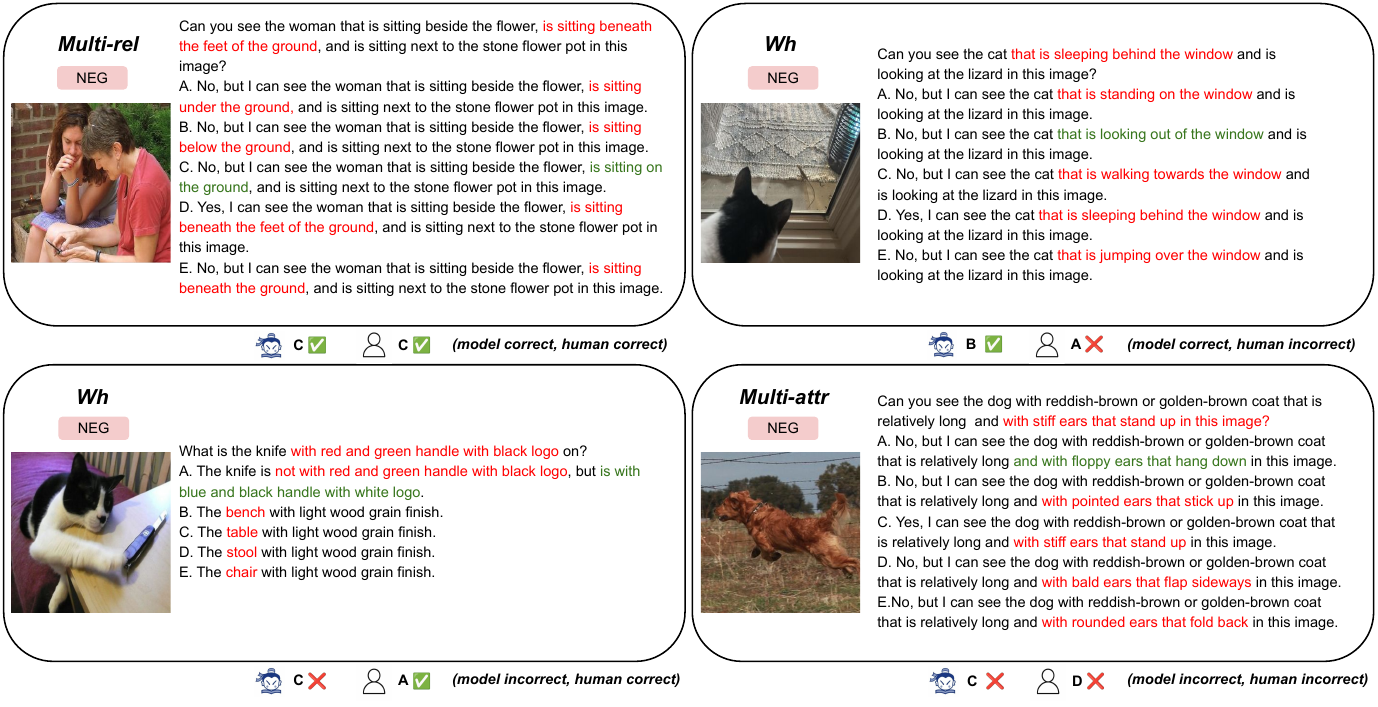}
  \caption{Success \& failure analysis matrix for InternVL3.5-14B~\cite{wang2025internvl35} (denoted as ``model" in the figure) and Human. All MCQs are included in the human study.}
  \label{fig:supp_human_failure_cases}
\end{figure*}

\myparagraph{Success and failure cases.} 
As Tab.~\ref{tab:supp_human_study} shows, humans achieve over 90\% paired accuracy across all settings in \textsc{FINER-CompreCap} and \textsc{FINER-DOCCI}. Although we can only evaluate human performance on a limited subset due to resource constraints, we do observe many cases where humans succeed on MCQs that a model like InternVL-3.5-14B~\cite{wang2025internvl35} fails on. Notably, there are also MCQs where humans fail but models succeed. Representative success and failure cases are shown in Fig.~\ref{fig:supp_human_failure_cases}.

From Fig.~\ref{fig:supp_human_failure_cases}, human errors can be grouped into two main types: carelessness and ambiguity. In the upper-right example, the human selects ``sleeping behind the window'', likely due to a simple oversight or a ``yes'' bias, similar to how InternVL-3.5-14B fails in the lower-right example. The second type of error arises from subjective or ambiguous visual attributes. In the dog example, the human chooses ``with bald ears that flap sideways'' instead of ``with floppy ears that hang down''. This is partly understandable, since ``flap sideways'' describes some of the observed motion even though the ears are not truly ``bald''. Strictly speaking, ``bald ears that flap sideways'' should be considered a false attribute (only partially correct), especially when compared to ``floppy ears that hang down'' (correct).

This motivates our choice to design FINER as an MCQ benchmark rather than using simple yes/no questions. By comparing multiple options, both humans and models are encouraged to pick the better description, which reduces ambiguity to some extent. Nevertheless, even with our entropy-based filtering pipeline, additional human verification, and MCQ design, the scale of FINER means that a certain amount of subjectivity, ambiguity, and annotation errors in the descriptions remains unavoidable. A valid future direction is to construct FINER benchmarks fully with human annotations, better aligning the evaluation with human subjectivity in assessing hallucinations. 

In our human studies, participants answer 20 MCQs per subset, which is small relative to the scale of both benchmarks. This is mainly because FINER is highly text-intensive, requiring substantial reading time. Scaling up the human study would likely further reduce human accuracy due to the reading burden and potential noise, since the benchmark is not fully created and validated by humans. We therefore treat the limited scale of the human studies as a limitation, and emphasize that these results only reflect human behavior on a small subset and given ample answering time, rather than serving as a valid measure of overall benchmark quality.

\begin{figure*}[b]
  \centering
  \includegraphics[width=\textwidth]{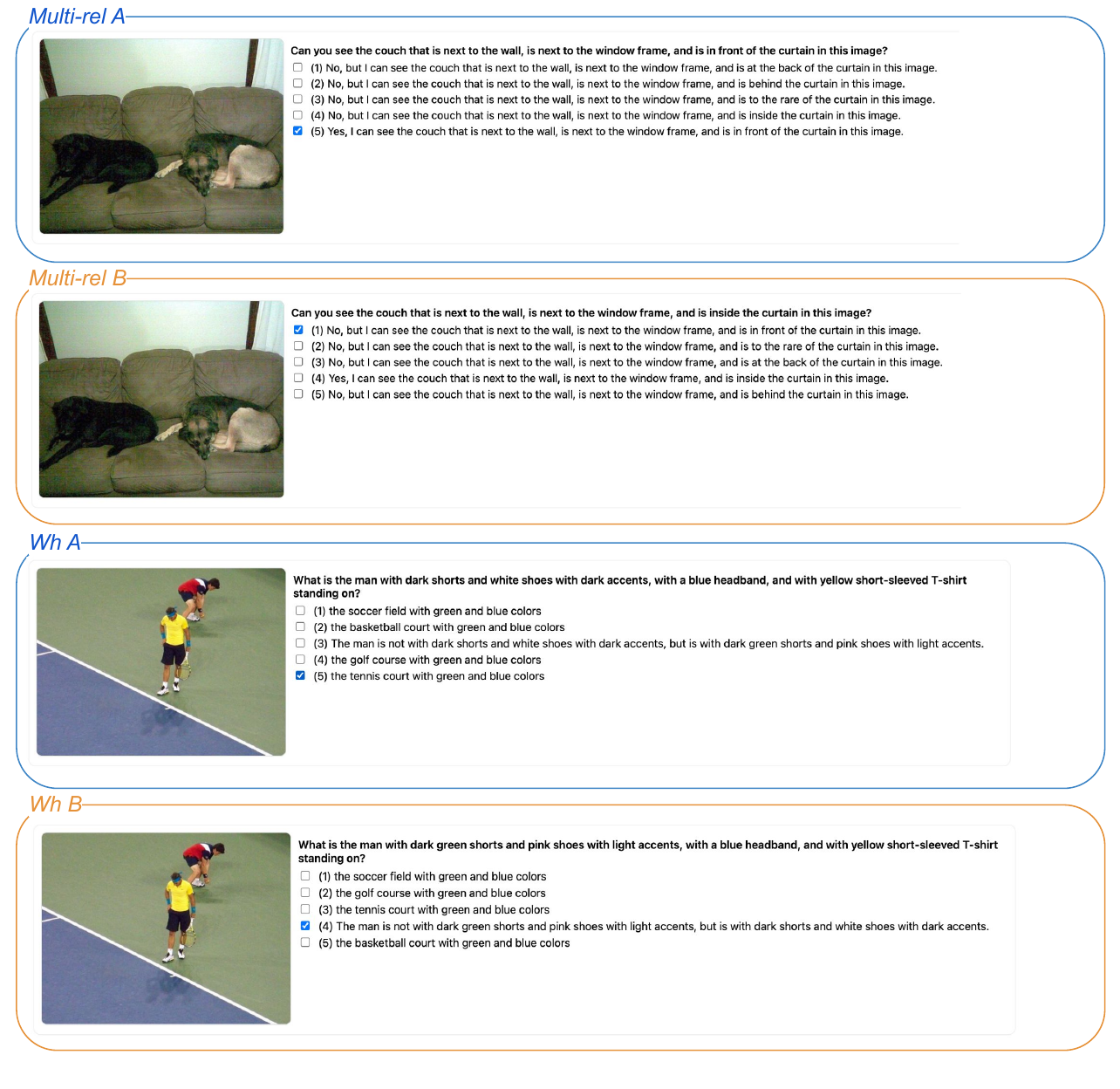}
  \caption{Examples of our human study survey for \textsc{FINER-CompreCap}. Example questions from Multi-rel and Wh are shown in the figure. Ticked boxes represent ground-truth choices. We use blue color to represent the questions for version A, while orange representing the questions for version B.}
  \label{fig:supp_comprecap_human_study_examples}
\end{figure*}

\begin{figure*}[t]
  \centering
  \includegraphics[width=\textwidth]{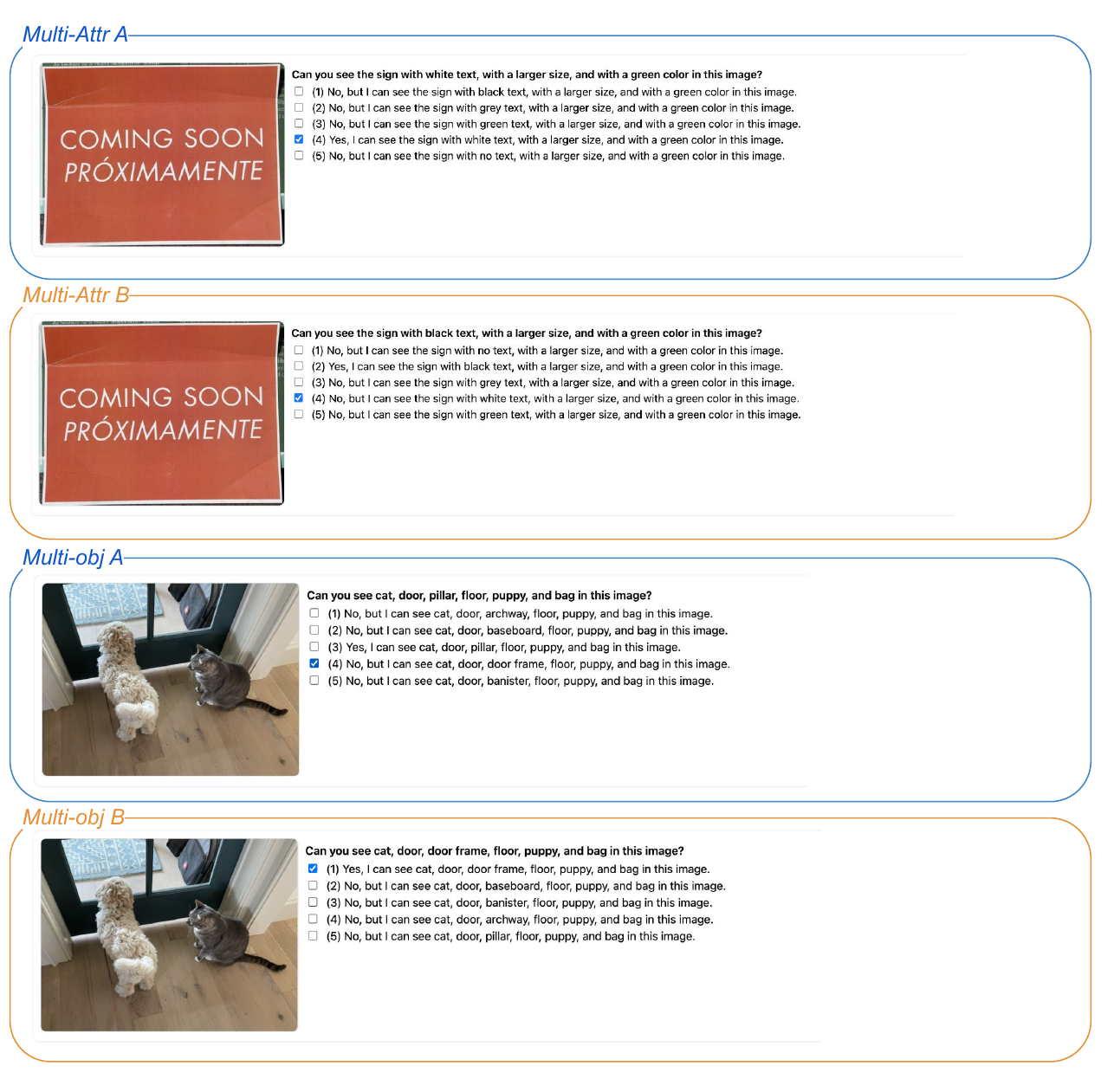}
  \caption{Examples of our human study survey for \textsc{FINER-docci}. Example questions from Multi-attr and Multi-obj. Ticked boxes represent the ground-truth choice. We use blue color to represent the questions for version A, while orange representing the questions for version B.
  }
  \label{fig:supp_docci_human_study_examples}
\end{figure*}

\section{Templates}
\label{sec:supp_templates}

To construct training set for \ours. Sec.~\ref*{sec:training_with_finer} describes how we run Phi-4-14B~\cite{abdin2024phi} over captions to extract positive phrases
\[
\big\{\Psi_{\textsc{Obj}}^+,\ \Psi_{\textsc{Attr}}^+,\ \Psi_{\textsc{Rel}}^+,\ \Psi_{\textsc{Wh}}^+\big\}
\]
and negative phrases
\[
\big\{\Psi_{\textsc{Obj}}^-,\ \Psi_{\textsc{Attr}}^-,\ \Psi_{\textsc{Rel}}^-,\ \Psi_{\textsc{Wh}}^-\big\}.
\]

\begin{figure*}[t]
  \centering
  \includegraphics[width=\textwidth]{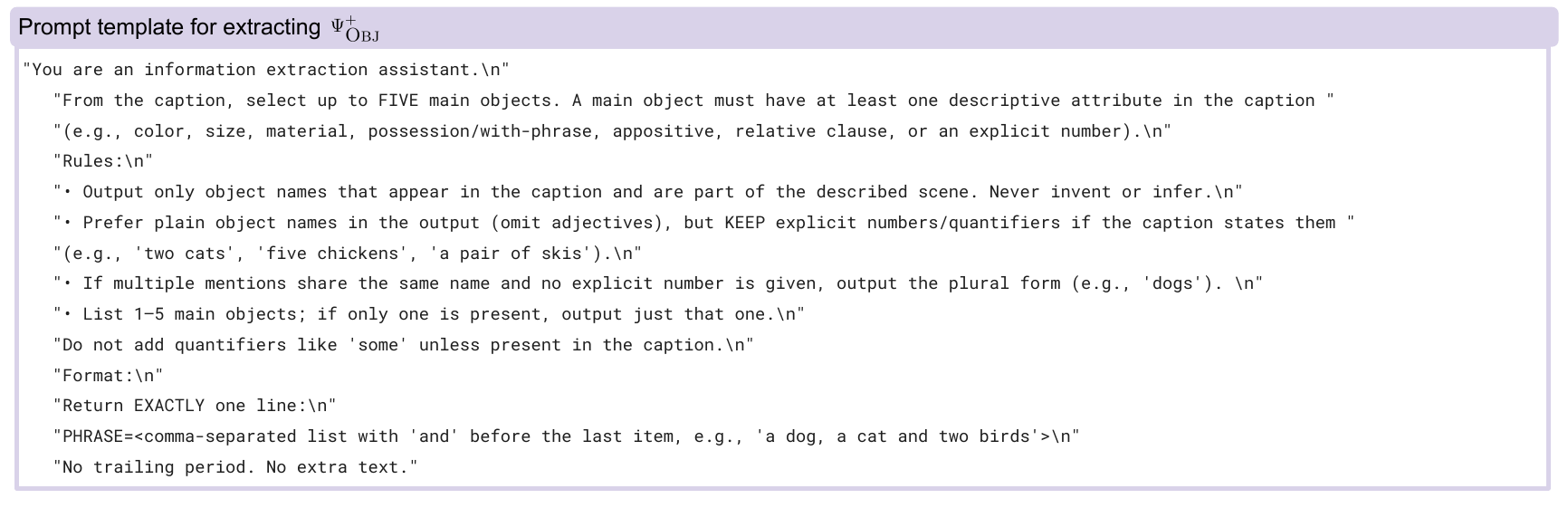}
  \caption{Prompt Template for extracting $\Psi_{\textsc{Obj}}^+$}
  \label{fig:supp_prompt_pos_obj}
\end{figure*}

\begin{figure*}[t]
  \centering
  \includegraphics[width=\textwidth]{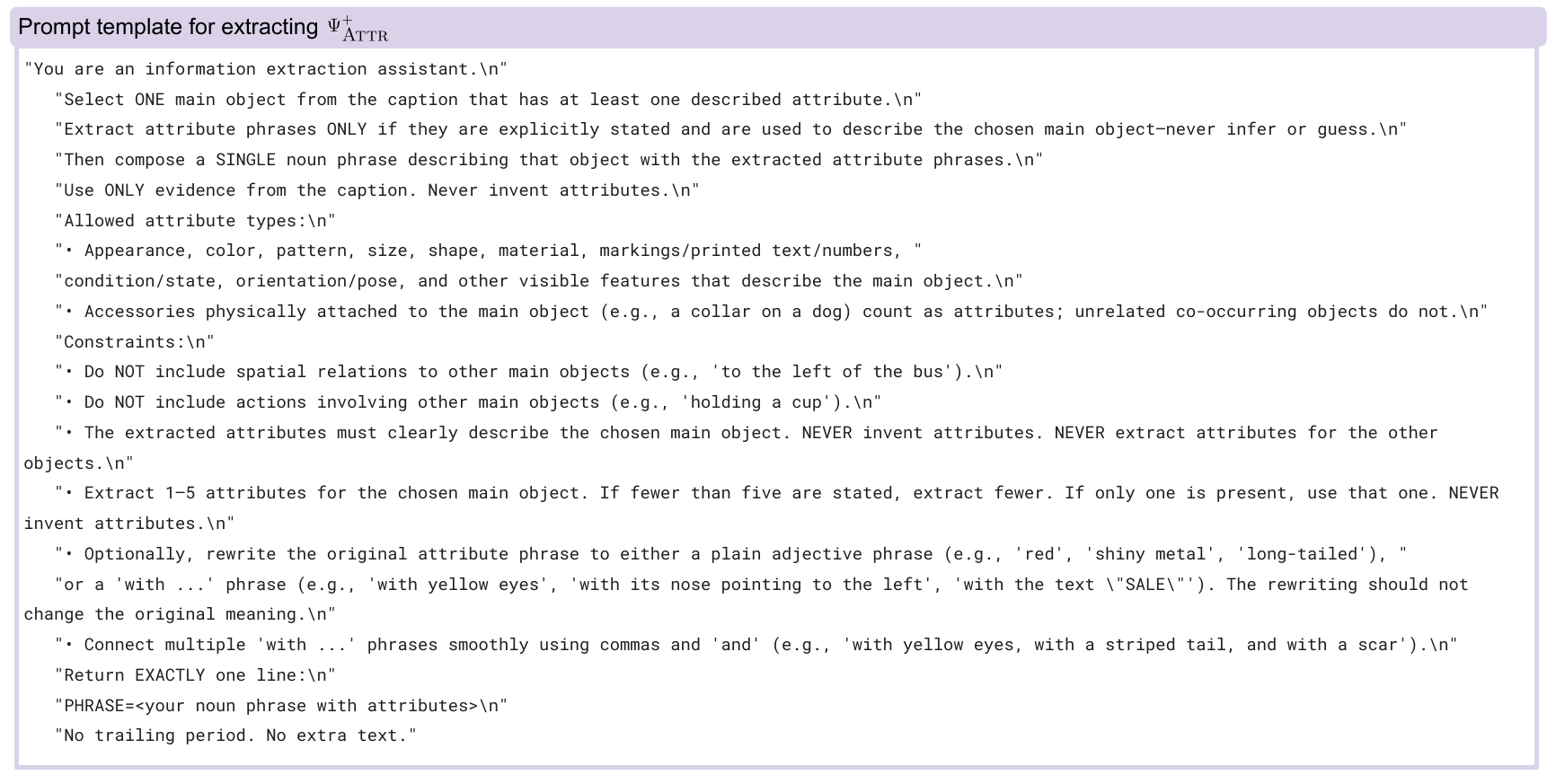}
  \caption{Prompt Template for extracting $\Psi_{\textsc{Attr}}^+$}
  \label{fig:supp_prompt_pos_attr}
\end{figure*}

\begin{figure*}[t]
  \centering
  \includegraphics[width=\textwidth]{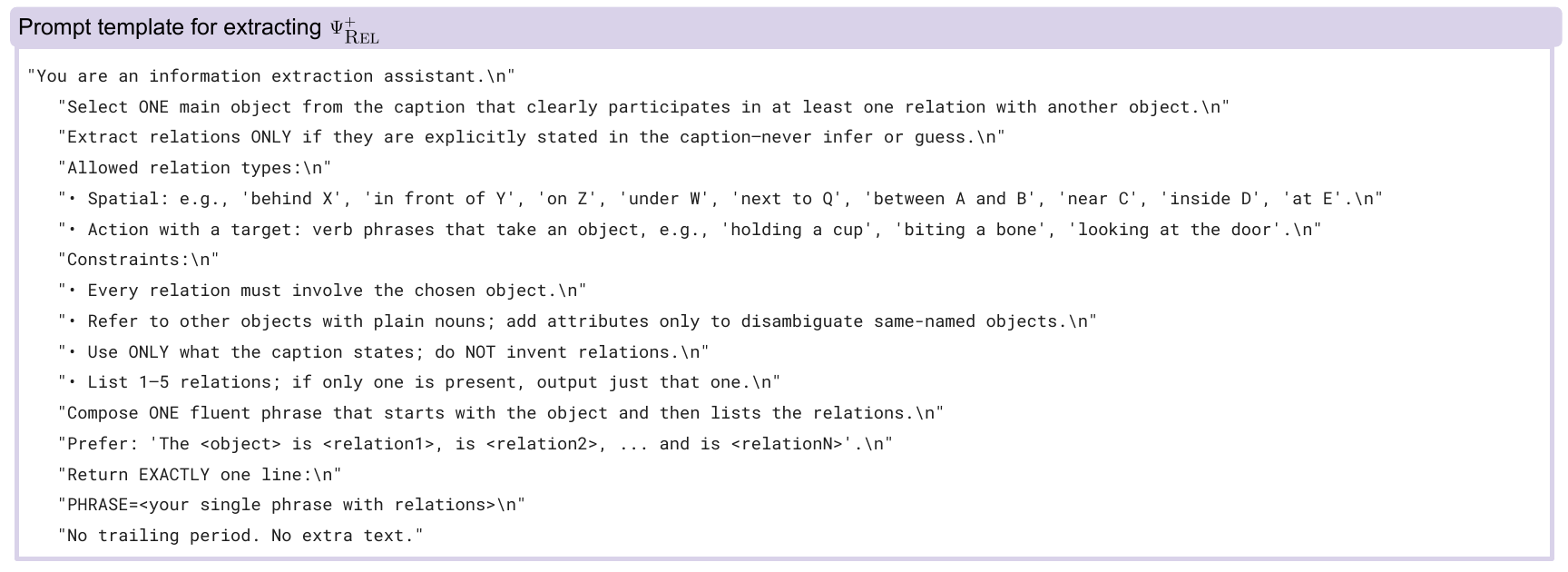}
  \caption{Prompt Template for extracting $\Psi_{\textsc{Rel}}^+$}
  \label{fig:supp_prompt_pos_rel}
\end{figure*}

\textbf{OBJ / ATTR / REL.}
For \textsc{OBJ}, \textsc{ATTR}, and \textsc{REL}, we first extract positive phrases $\big\{\Psi_{\textsc{Obj}}^+,\ \Psi_{\textsc{Attr}}^+,\ \Psi_{\textsc{Rel}}^+\big\}$ using the prompts shown in Fig.~\ref{fig:supp_prompt_pos_obj}, Fig.~\ref{fig:supp_prompt_pos_attr}, and Fig.~\ref{fig:supp_prompt_pos_rel}. We then prompt the same LLM to generate the corresponding negative phrases $\big\{\Psi_{\textsc{Obj}}^-,\ \Psi_{\textsc{Attr}}^-,\ \Psi_{\textsc{Rel}}^-\big\}$ with the prompts in Fig.~\ref{fig:supp_prompt_neg_obj}, Fig.~\ref{fig:supp_prompt_neg_attr}, and Fig.~\ref{fig:supp_prompt_neg_rel}. Given these positive/negative phrase sets, we construct preference tuples
\[
(q^+, a^+_+, a^-_+) \quad \text{and} \quad (q^-, a^+_-, a^-_-)
\]
for each of \textsc{OBJ}, \textsc{ATTR}, and \textsc{REL} via template-based composition, by using a pool of five templates as below:

\begin{center}
\fbox{%
  \parbox{0.96\linewidth}{%
    \small
    (1)\\
    \texttt{Does this image contain \{X\}?}\\[2pt]
    \texttt{Yes, this image contains \{Y\}.}\\[2pt]
    \texttt{No, but this image contains \{Z\}.}\\[4pt]
    (2)\\
    \texttt{Does this image show \{X\}?}\\[2pt]
    \texttt{Yes, this image shows \{Y\}.}\\[2pt]
    \texttt{No, but this image shows \{Z\}.}\\[4pt]
    (3)\\
    \texttt{Does this image include \{X\}?}\\[2pt]
    \texttt{Yes, this image includes \{Y\}.}\\[2pt]
    \texttt{No, but this image includes \{Z\}.}\\[4pt]
    (4)\\
    \texttt{Can you see \{X\} in this image?}\\[2pt]
    \texttt{Yes, I can see \{Y\} in this image.}\\[2pt]
    \texttt{No, but I can see \{Z\} in this image.}\\[4pt]
    (5)\\
    \texttt{Can \{X\} be seen in this image?}\\[2pt]
    \texttt{Yes, \{Y\} can be seen in this image.}\\[2pt]
    \texttt{No, but \{Z\} can be seen in this image.}
  }%
}
\end{center}

To avoid overfitting to a single fixed pattern and to stay consistent with the FINER benchmarks, we randomly choose one of the above five templates for each example. Each template contains placeholders $\{X\}$, $\{Y\}$, and $\{Z_1\},\dots,\{Z_4\}$ that are filled with phrases.

In the positive configuration $(q^+, a^+_+, a^-_+)$, the ``Yes'' answer will be the accepted response $a^+_+$ while the ``No'' answer will be the rejected response $a^-_+$. The question and the ``Yes'' answer both use the positive phrase $\Psi^+$, while all ``No'' answers use the negative phrase $\Psi^-$: 
\[
\begin{aligned}
\{X\}   &= \Psi^+, \\
\{Y\}   &= \Psi^+, \\
\{Z\} &= \Psi^-
\end{aligned}
\]

In the negative configuration $(q^-, a^-_+, a^-_-)$, the ``No'' answer will be the accepted response $a^+_-$ while the ``Yes'' answer will be the rejected response $a^-_-$. The question and all ``No'' answers use the negative phrase $\Psi^-$, while the ``Yes'' answer uses the positive phrase $\Psi^+$:
\[
\begin{aligned}
\{X\}   &= \Psi^-, \\
\{Y\}   &= \Psi^+, \\
\{Z\} &= \Psi^- 
\end{aligned}
\]

\begin{figure*}[t]
  \centering
  \includegraphics[width=\textwidth]{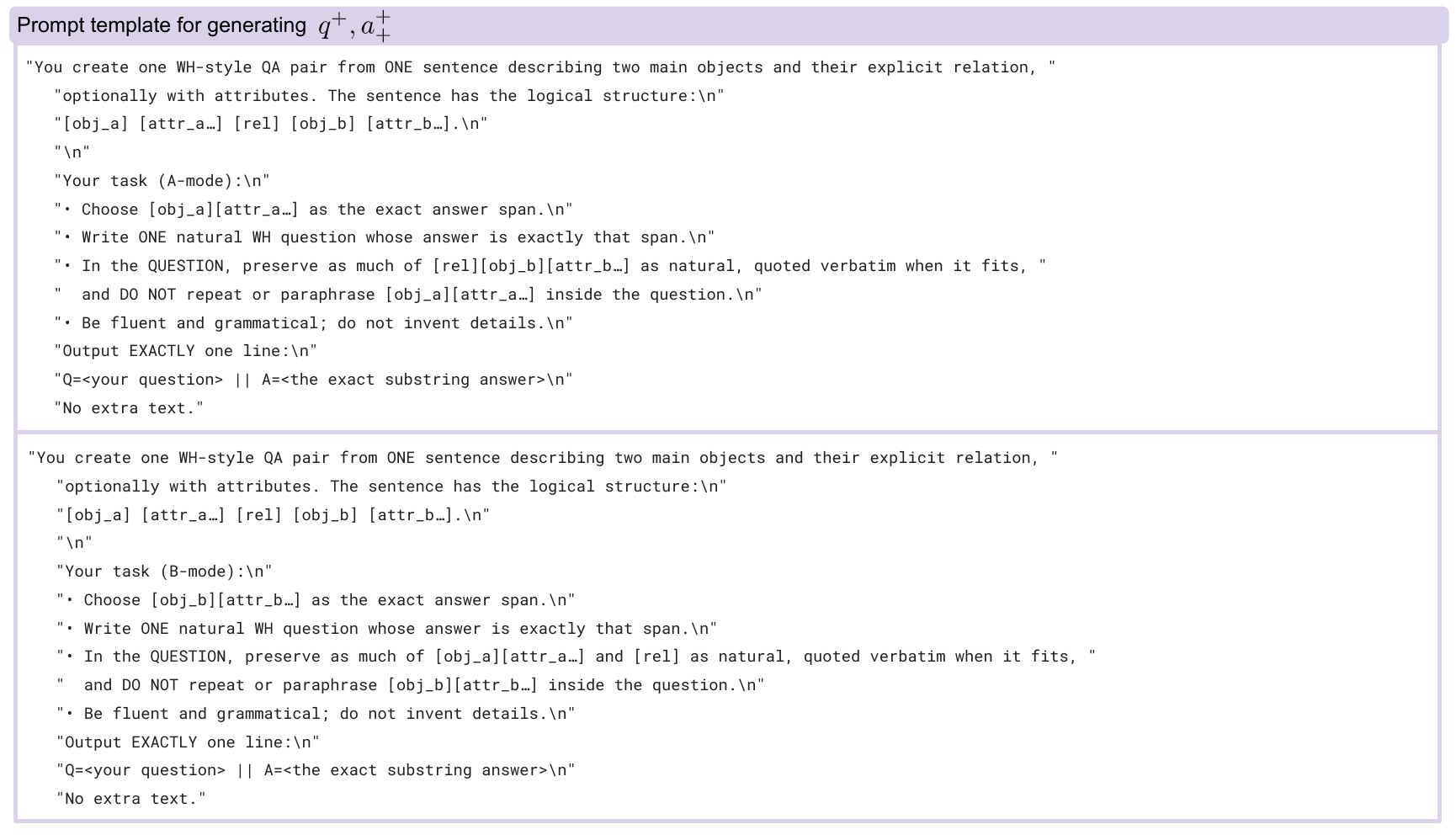}
  \caption{Prompt Template for generating $(q^+, a^+_+)$ for \textsc{WH} setting}
  \label{fig:supp_prompt_pos_wh}
\end{figure*}

\begin{figure*}[t]
  \centering
  \includegraphics[width=\textwidth]{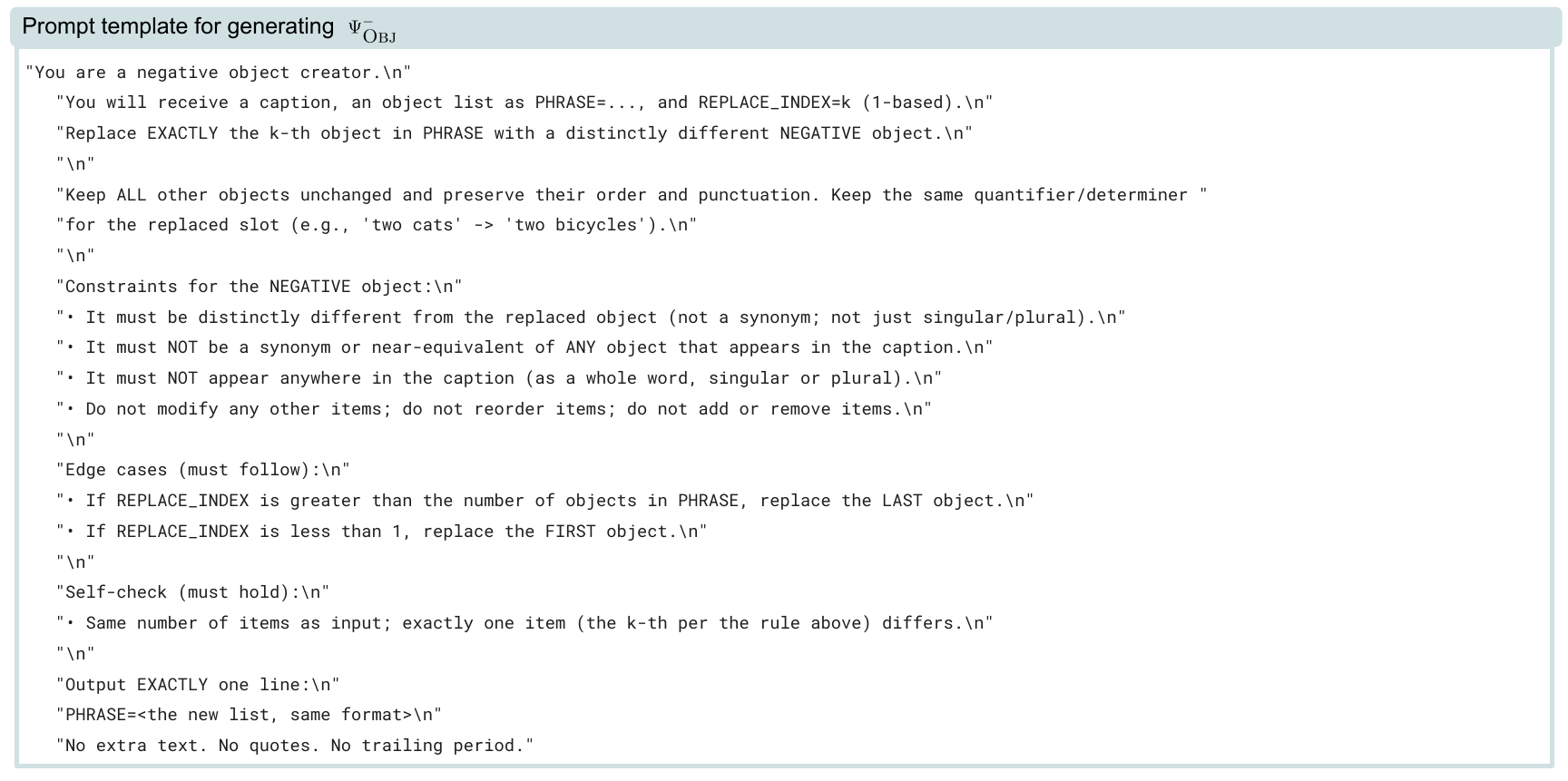}
  \caption{Prompt Template for generating $\Psi_{\textsc{Obj}}^-$}
  \label{fig:supp_prompt_neg_obj}
\end{figure*}

\begin{figure*}[t]
  \centering
  \includegraphics[width=\textwidth]{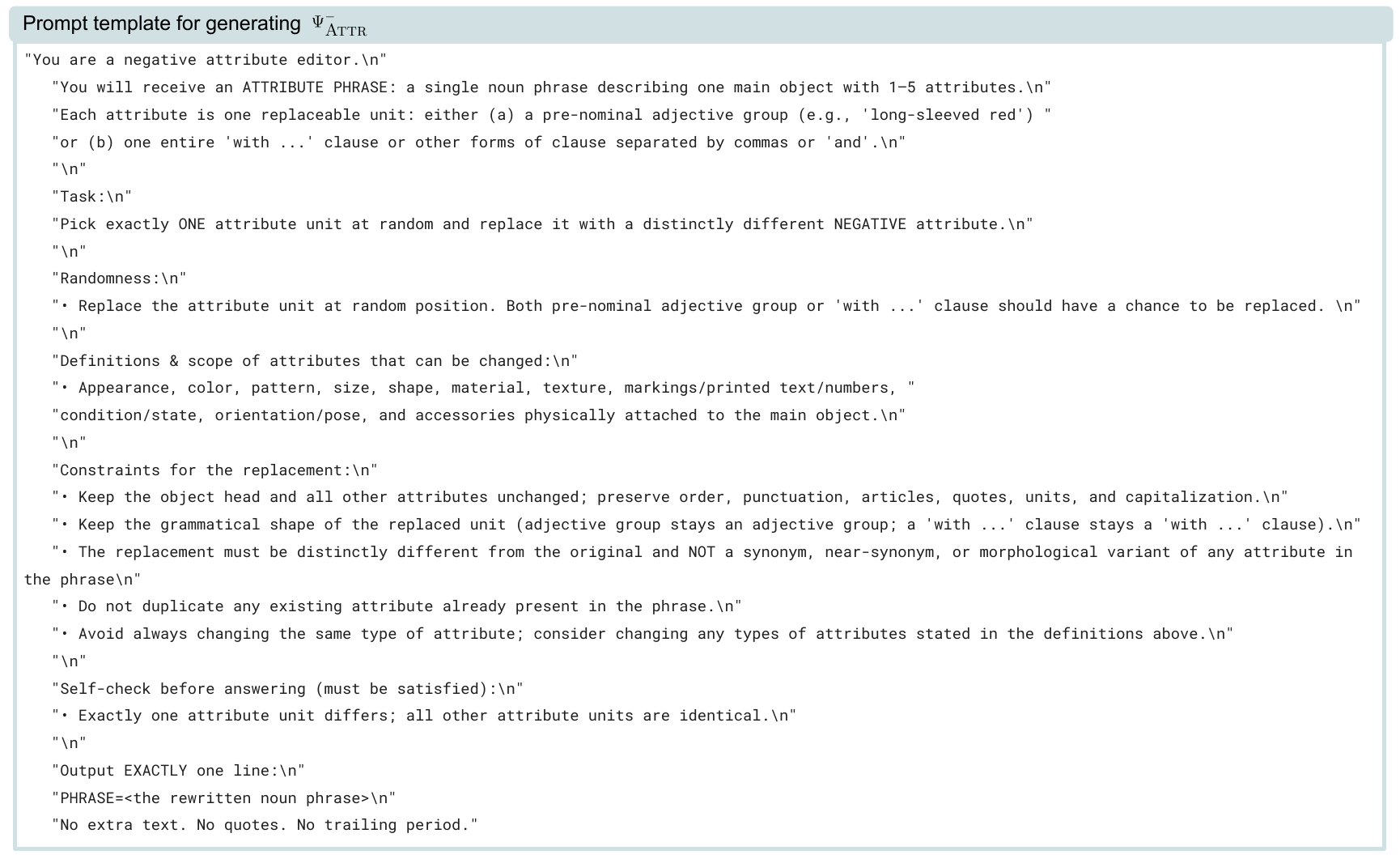}
  \caption{Prompt Template for generating $\Psi_{\textsc{Attr}}^-$}
  \label{fig:supp_prompt_neg_attr}
\end{figure*}

\begin{figure*}[t]
  \centering
  \includegraphics[width=\textwidth]{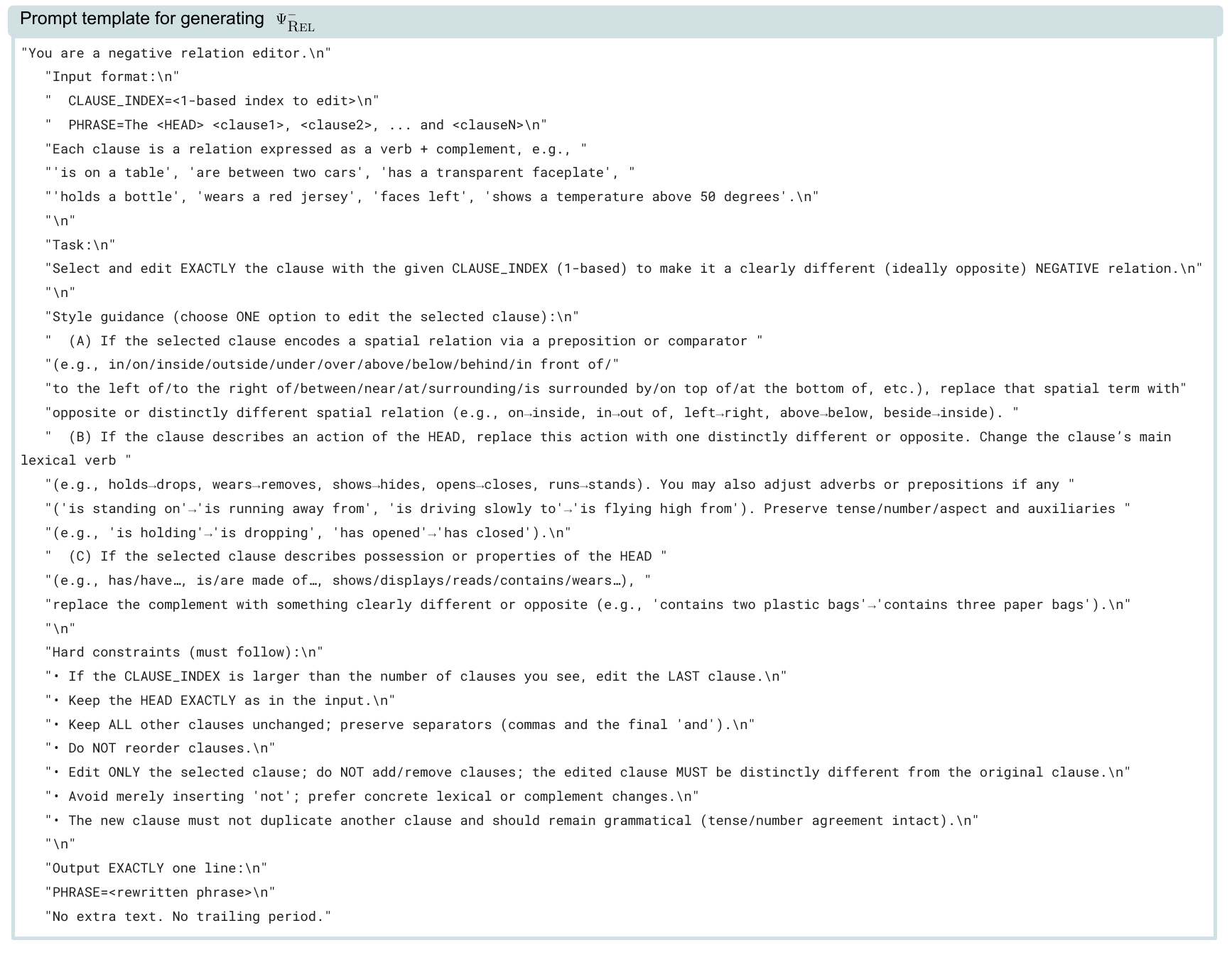}
  \caption{Prompt Template for generating $\Psi_{\textsc{Rel}}^-$}
  \label{fig:supp_prompt_neg_rel}
\end{figure*}

\begin{figure*}[t]
  \centering
  \includegraphics[width=\textwidth]{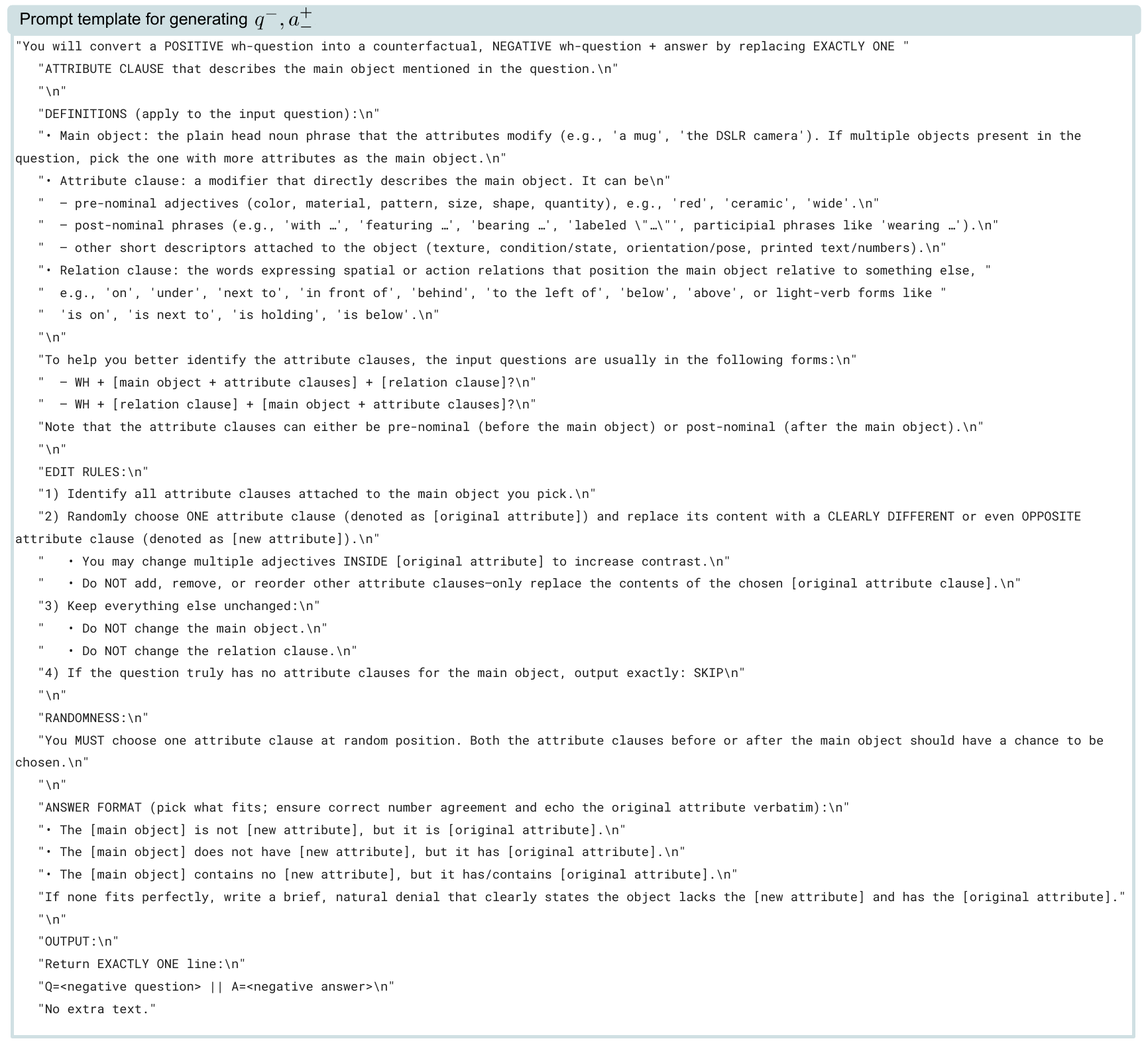}
  \caption{Prompt Template for generating $(q^-, a^+_-)$ for \textsc{WH} setting}
  \label{fig:supp_prompt_neg_wh}
\end{figure*}

\begin{figure*}[t]
  \centering
  \includegraphics[width=\textwidth]{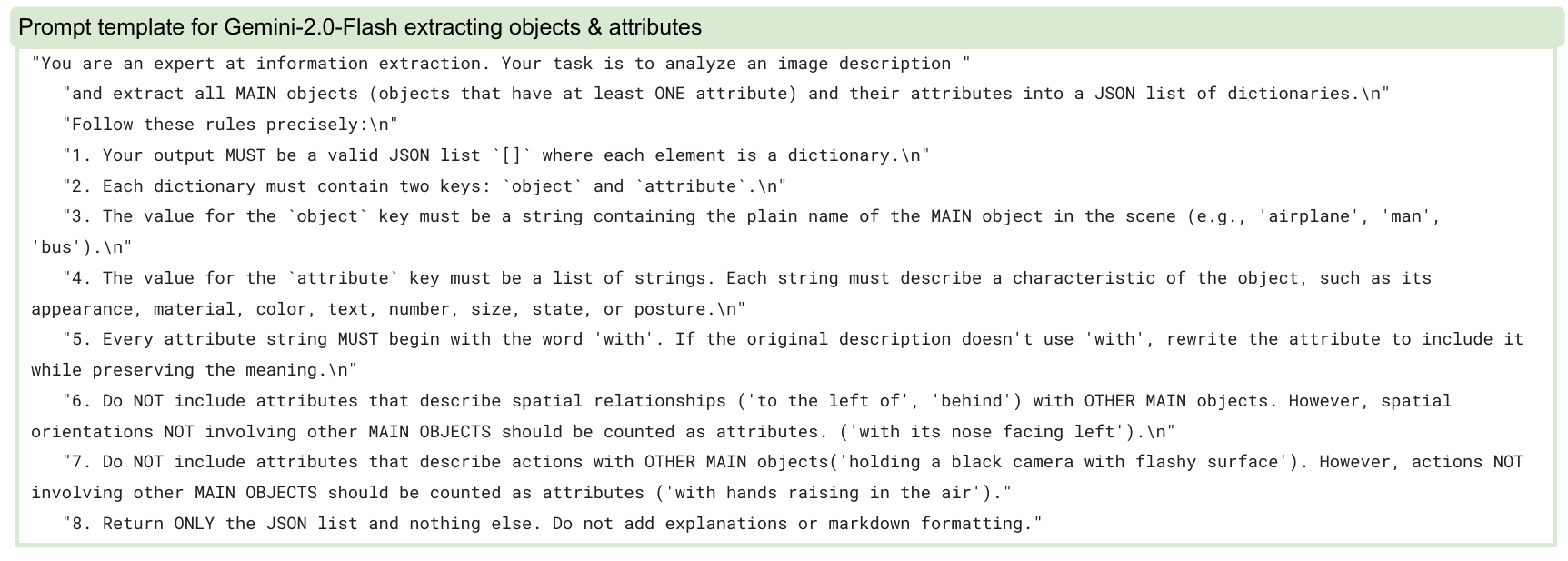}
  \caption{Prompt Template for extracting objects and attributes using Gemini-2.0-Flash~\cite{team2023gemini} when constructing \textsc{FINER-DOCCI}.}
  \label{fig:supp_prompt_gemini_obj_attr}
\end{figure*}

\begin{figure*}[t]
  \centering
  \includegraphics[width=\textwidth]{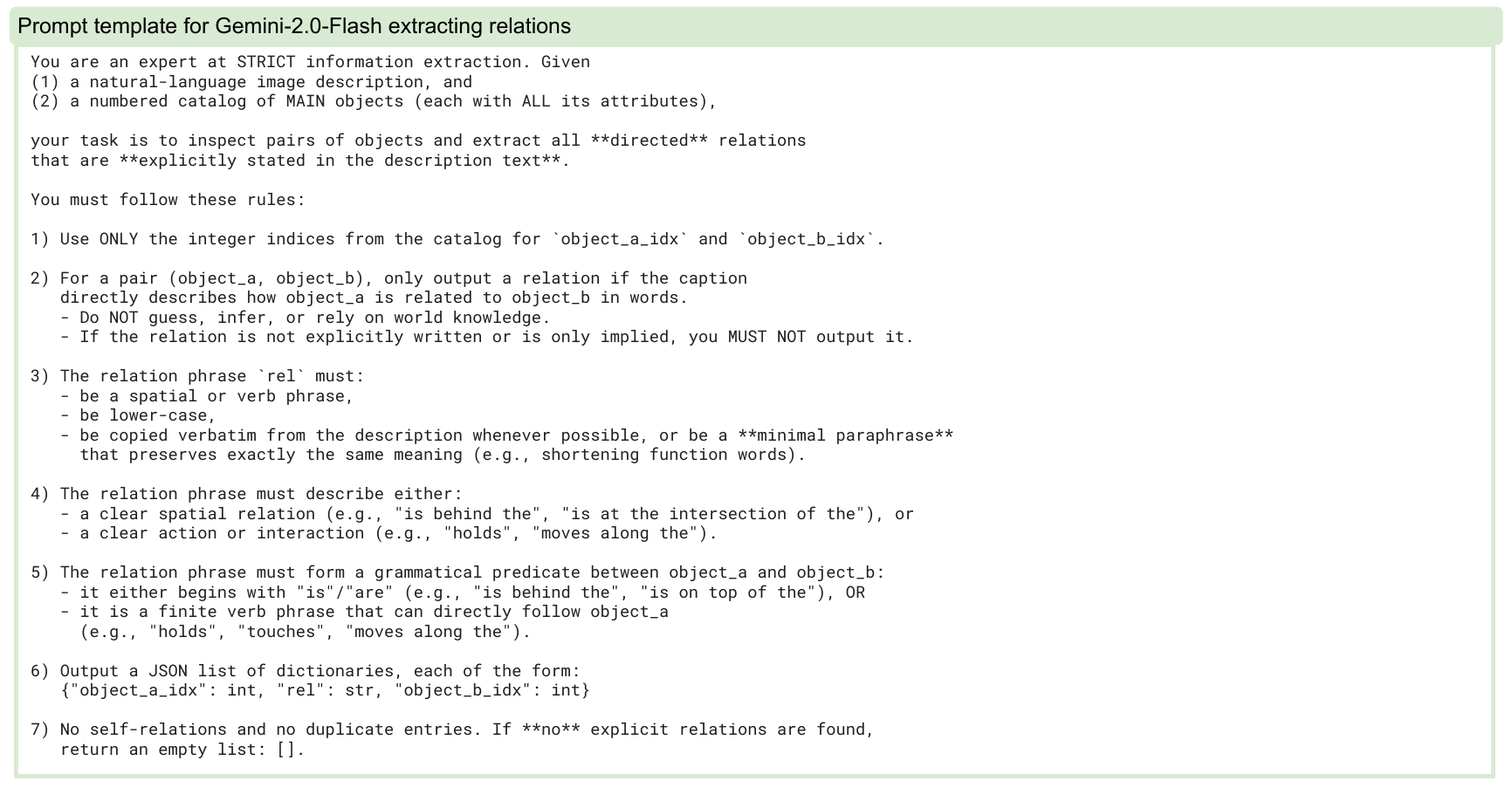}
  \caption{Prompt Template for extracting relations using Gemini-2.0-Flash~\cite{team2023gemini} when constructing \textsc{FINER-DOCCI}.}
  \label{fig:supp_prompt_gemini_rel}
\end{figure*}

\textbf{WH.}
For \textsc{Wh}, the preference tuples
\[
(q^+, a^+_+, a^-_+) \quad \text{and} \quad (q^-, a^+_-, a^-_-)
\]
are directly constructed by the LLM, rather than via our fixed templates. We therefore do not apply the above template-based composition to \textsc{Wh}, and instead use dedicated prompts to let the LLM generate the question and its positive/negative answers. The prompts used to construct a pair of $(q^+, a^+_+)$ and $(q^-, a^+_-)$ for \textsc{Wh} are shown in Fig.~\ref{fig:supp_prompt_pos_wh} and Fig.~\ref{fig:supp_prompt_neg_wh}, respectively. Concretely, the LLM first produces two Wh questions about the same underlying scene: a positive question $q^+$, whose premise is consistent with the image and whose accepted response $a^+_+$ directly answers what the question asks for, and a negative question $q^-$, whose premise partially conflicts with the image content so that its accepted response $a^+_-$ explicitly negates the question itself. We then symmetrize this pair by assigning each accepted response as the other question's rejected response, i.e., $a^-_+ := a^+_-$ and $a^-_- := a^+_+$. In this way we obtain the final preference tuples $(q^+, a^+_+, a^-_+)$ and $(q^-, a^+_-, a^-_-)$.


\end{document}